\RequirePackage{amsmath}
\documentclass[twocolumn]{svjour3}          
\smartqed  
\usepackage{url}
\usepackage{times}
\usepackage{epsfig}
\usepackage{amssymb}
\usepackage{pifont}
\usepackage{multirow}
\usepackage{graphicx}
\usepackage{makecell}
\usepackage{xcolor}
\usepackage{hyphenat}
\usepackage[colorinlistoftodos]{todonotes}

\graphicspath{{figures/},{figures/qual_results/},{figures/tu-berlin-confusing/}}
\DeclareGraphicsExtensions{.pdf,.png}

\newcommand{\viz}{\textit{viz.}}
\newcommand{\ie}{\textit{i.e.}}
\newcommand{\eg}{\textit{e.g.}}
\newcommand{\etal}{\textit{et al.}}
\newcommand{\fig}[1]{Fig. \ref{#1}}
\newcommand{\eq}[1]{eqn. (\ref{#1})}

\newcommand{\sect}[1]{Section \ref{#1}}
\newcommand{\tab}[1]{Table \ref{#1}}

\newcommand{\cmark}{\Huge\textcolor{green}{\ding{51}}}%
\newcommand{\xmark}{\Huge\textcolor{red}{\ding{55}}}%
\newcommand{\myparagraph}[1]{\vspace{4pt}\noindent{\bf #1}}

\begin{document}

\title{Semantically Tied Paired Cycle Consistency for Any-Shot Sketch-based Image Retrieval}



\author{Anjan Dutta         \and
        Zeynep Akata 
}


\institute{Anjan Dutta \at
              Department of Computer Science, University of Exeter, Innovation Centre, Streatham Campus, Exeter, EX4 4RN, UK \\
              \email{a.dutta@exeter.ac.uk}           
           \and
           Zeynep Akata \at
              Cluster of Excellence Machine Learning, University of T\"ubingen, T\"{u}bingen AI Center, 72076, T\"{u}bingen, Germany \\
              \email{zeynep.akata@uni-tuebingen.de}
}

\date{Received: date / Accepted: date}

\maketitle

\begin{abstract}
Low-shot sketch-based image retrieval is an emerging task in computer vision, allowing to retrieve natural images relevant to hand-drawn sketch queries that are rarely seen during the training phase. Related prior works either require aligned sketch-image pairs that are costly to obtain or inefficient memory fusion layer for mapping the visual information to a semantic space. In this paper, we address any-shot, \ie~zero-shot and few-shot, sketch-based image retrieval (SBIR) tasks, where we introduce the few-shot setting for SBIR. For solving these tasks, we propose a semantically aligned paired cycle-consistent generative adversarial network (SEM-PCYC) for any-shot SBIR, where each branch of the generative adversarial network maps the visual information from sketch and image to a common semantic space via adversarial training. Each of these branches maintains cycle consistency that only requires supervision at the category level, and avoids the need of aligned sketch-image pairs. A classification criteria on the generators' outputs ensures the visual to semantic space mapping to be class-specific. Furthermore, we propose to combine textual and hierarchical side information via an auto-encoder that selects discriminating side information within a same end-to-end model. Our results demonstrate a significant boost in any-shot SBIR performance over the state-of-the-art on the extended version of the challenging Sketchy, TU-Berlin and QuickDraw datasets.
\end{abstract}%

\section{Introduction}
\label{sec:intro}
Matching natural images with free-hand sketches, \ie~\emph{sketch-based image retrieval} (SBIR)~\cite{Yu2015,Yu2016,Liu2017DSH,Pang2017FGSBIR,Song2017SpatSemAttn,Shen2018ZSIH,Zhang2018GDH,Chen2018DeepSB3DSR,Yelamarthi2018ZSBIR,Dutta2019SEM-PCYC,Dey2019doodle2search} has received a lot of attention. Since sketches can effectively express shape, pose and some fine-grained details of the target images, SBIR serves a favorable scenario complementary to the conventional text-image cross-modal retrieval or the classical content based image retrieval protocol. This may be because in some situations it is difficult to provide a textual description or a suitable image of the desired query, whereas, an user can easily draw a sketch of the desired object on a touch screen.

\begin{figure*}[t]
\centering
\includegraphics[width=0.8\textwidth]{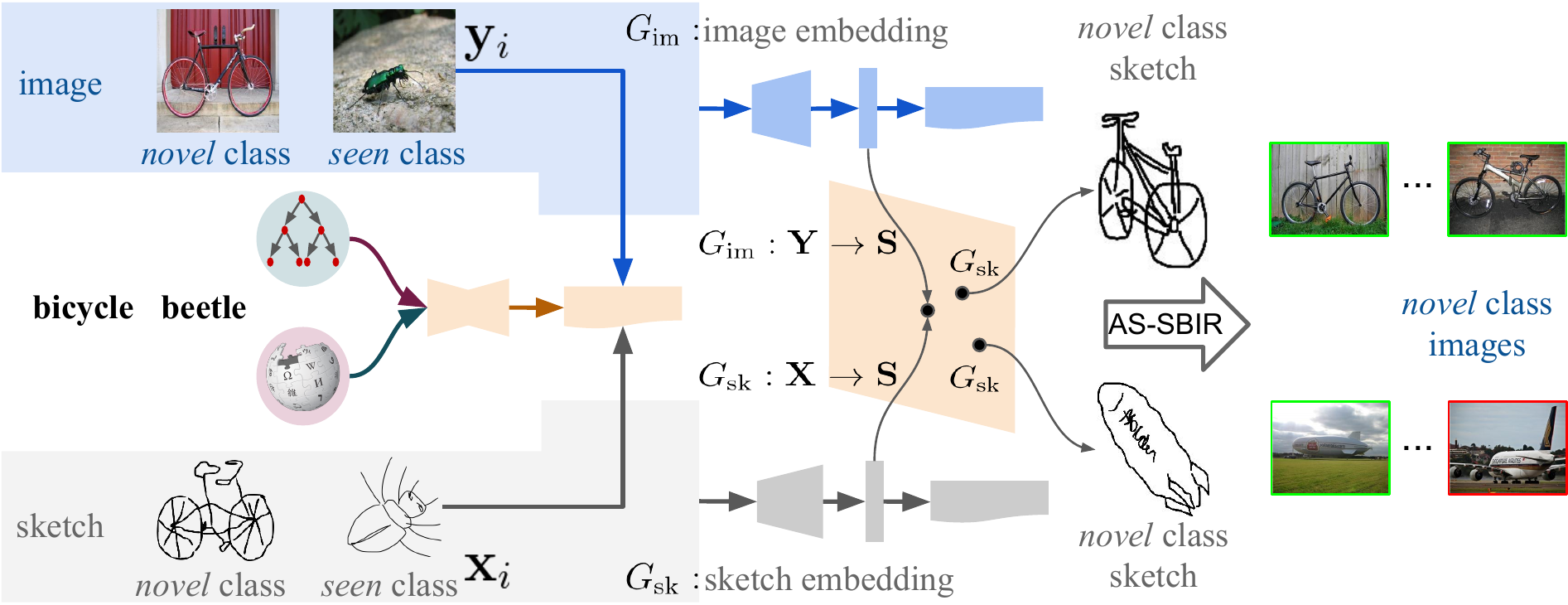}
\caption{Our SEM-PCYC model learns to map visual information from \emph{seen}-class sketches and images to a semantic space through an adversarial training procedure in zero-shot SBIR setting. Furthermore, our mdoel is flexible enough to use a few examples from \emph{novel} classes to fine-tune the model, where the \emph{novel} classes contain a few labeled samples in few-shot SBIR (FS-SBIR) setting. During the testing phase the learned mappings are used to generate embeddings of the \emph{novel} classes. We refer to the combination of zero- and few-shot SBIR as any-shot SBIR (AS-SBIR).}
\label{fig:sem-pcyc-teaser}
\end{figure*}

As the visual information from all classes gets explored by the system during training, with overlapping training and test classes, existing SBIR methods perform well~\cite{Zhang2018GDH}. Since for practical applications there is no guarantee that the training data would include all possible queries, a more realistic setting is \emph{low-shot} or \emph{any-shot} SBIR (AS-SBIR)~\cite{Shen2018ZSIH,Yelamarthi2018ZSBIR,Dutta2019SEM-PCYC,Dey2019doodle2search}, which combines zero- and few-shot learning~\cite{Lampert2014ZSL,Vinyals2016MatchingNet,Xian2018ZSLGBU,Ravi2017FSL} and SBIR as a single task, where the aim is an accurate class prediction and a competent retrieval performance. However, this is an extremely challenging task, as it simultaneously deals with domain gap, intra-class variability and limited or no knowledge on \emph{novel} classes. Additionally, fine-grained SBIR~\cite{Pang2017FGSBIR,Pang2019GenFGSBIR} is an alternative sketch-based image retrieval task, allowing to search for specific object images, which has already received remarkable attention in the computer vision community. However, it has never been explored in low shot setting, which is an extremely challenging and at the same time of high practical relevance.

One of the major shortcomings of the prior work on any-shot SBIR is that a natural image is retrieved after learning a mapping from an input sketch to an output image using a training set of labelled \emph{aligned} pairs~\cite{Yelamarthi2018ZSBIR}. The supervision of the pair correspondence is to enhance the correlation of multi-modal data (here, sketch and image) so that learning can be guided by semantics. However, for many realistic scenarios, paired (aligned) training data is either unavailable or obtaining it is very expensive. Furthermore, often a joint representation of two or more modalities is learned by using a memory fusion layer~\cite{Shen2018ZSIH}, such as, tensor fusion~\cite{Hu2017TFN}, bilinear pooling~\cite{Yu2017MFBP} etc. These fusion layers are often expensive in terms of memory~\cite{Yu2017MFBP}, and extracting useful information from this high dimensional space could result in information loss~\cite{Yu2018MHBN}.

To alleviate these shortcomings, we propose a semantically aligned paired cycle consistent generative adversarial network (SEM-PCYC) model for any-shot SBIR task, where each branch either maps the sketch or image features to a common semantic space via an adversarial training. These two branches dealing with two different modalities (sketch and image) constitute an essential component for solving SBIR task. The cycle consistency constraint on each branch guarantees that the mapping of sketch or image modality to a common semantic space and their translation back to the original modality, avoiding the necessity of aligned sketch-image pairs. Imposing a classification loss on the semantically aligned outputs from the sketch and image space enforces the generated features in the semantic space to be discriminative which is very crucial for effective any-shot SBIR. Furthermore, inspired by the previous works on label embedding~\cite{Akata2015OutputEmbedding}, we propose to combine side information from text-based and hierarchical models via a feature selection auto-encoder~\cite{Wang2017FSAE} which selects discriminating side information based on intra and inter class covariance.


This paper extends our CVPR 2019 conference paper \cite{Dutta2019SEM-PCYC}, with the following additional contributions: (1) We propose to apply the SEM-PCYC model for any-shot SBIR task, \ie~addition to zero-shot paradigm, we introduce few-shot setting for SBIR and combine it with generalized setting, which has been experimentally proven to be effective for difficult or confusing classes. (2) We adapt the recent zero-shot SBIR models and ours to fine-grained SBIR in the generalized low-shot setting and provide an extensive benchmark including quantitative and qualitative evaluations. (3) We evaluate our model on one recent dataset, i.e. QuickDraw, in addition to extending our experiments to new settings with Sketchy and TU-Berlin. We show that our proposed model consistently improves the state-of-the-art results of any-shot SBIR on all the three datasets.

\section{Related Work}
\label{sec:related}
As our work belongs at the verge of sketch-based image retrieval and any-shot learning task, we briefly review the relevant literature from these fields.

\myparagraph{Sketch Based Image Retrieval (SBIR).}
Attempts for solving SBIR task mostly focus on bridging the domain gap between sketch and image, which can roughly be grouped in \emph{hand-crafted} and \emph{cross-domain deep learning-based} methods~\cite{Liu2017DSH}. Hand-crafted methods mostly work by extracting the edge map from natural image and then matching them with sketch using a Bag-of-Words model on top of some specifically designed SBIR features, \viz, gradient field HOG \cite{Hu2013}, histogram of oriented edges \cite{Saavedra2014SoftComp}, learned key shapes \cite{Saavedra2015LKS} etc. However, the difficulty of reducing domain gap remained unresolved as it is extremely challenging to match edge maps with unaligned hand drawn sketch. This domain shift issue is further addressed by neural network models where domain transferable features from sketch to image are learned in an end-to-end manner. Majority of such models use variant of siamese networks~\cite{Qi2016SBIRSiamese,Sangkloy2016,Yu2016,Song2017FineGrained} that are suitable for cross-modal retrieval. These frameworks either use generic ranking losses, \viz, contrastive loss~\cite{Chopra2005}, triplet ranking loss~\cite{Sangkloy2016} or more sophisticated HOLEF based loss~\cite{Song2017SpatSemAttn}) for the same. Further to these discriminative losses, Pang~\etal~\cite{Pang2017FGSBIR} introduced a discriminative-generative hybrid model for preserving all the domain invariant information useful for reducing the domain gap between sketch and image. Alternatively, \cite{Liu2017DSH,Zhang2018GDH} focus on learning cross-modal hash code for category level SBIR within an end-to-end deep model. 

In addition to the above coarse-grained SBIR models, fine-grained sketch-based image retrieval (FG-SBIR) has gained popularity recently \cite{Li2014FGSBIRDeformPart,Song2017FineGrained,Song2017SpatSemAttn,Pang2017FGSBIR}. In this more realistic setting, a FG-SBIR model allows to search a specific object or image. First, models tackled this task using deformable part model and graph matching~\cite{Li2014FGSBIRDeformPart}. Recently, different ranking frameworks and corresponding losses, such as, siamese~\cite{Pang2017FGSBIR}, triplet~\cite{Sangkloy2016}, quadruplet~\cite{Song2017FineGrained} networks were used for the same. \cite{Song2017SpatSemAttn} proposed attention model for FG-SBIR task, \cite{Zhang2018GDH} improving retrieval efficiency using a hashing scheme. 

\myparagraph{Zero-Shot Learning (ZSL) and Few-Shot Learning (FSL).}
Zero-shot learning in computer vision refers to recognizing objects whose instances are not seen during the training phase; a comprehensive and detailed survey on ZSL is available in~\cite{Xian2018ZSLGBU}. Early works on ZSL~\cite{Lampert2014ZSL,Jayaraman2014ZSR,Changpinyo2016ZSL,Al-Halah2016ZSL} make use of attributes within a two-stage approach to infer the label of an image that belong to the \emph{unseen} classes. However, the recent works~\cite{Frome2013Devise,Romera-Paredes2015ESA,Akata2015OutputEmbedding,Akata2016LabelEmbedding,Kodirov2017SAE} directly learn a mapping from image feature space to a semantic space. Many other ZSL approaches learn non-linear multi-modal embedding~\cite{Socher2013ZSLCrossModalT,Akata2016LabelEmbedding,Xian2016ZSLLatentEmbedding,Changpinyo2017ZSL,Zhang2017ZSLDeepEmbedding}, where most of the methods focus to learn a non-linear mapping from the image space to the semantic space. Mapping both image and semantic features into another common intermediate space is another direction that ZSL approaches adapt~\cite{Zhang2015ZSLSemSim,Fu2015ZSOR,Zhang2016ZSLJointLatentSim,Akata2016ZSLSS,Long2017ZSL}. Although, most of the deep neural network models in this domain are trained using a discriminative loss function, a few generative models also exist~\cite{Wang2018ZSL,Xian2018ZSL,Chen2018ZSVR} that are used as a data augmentation mechanism. In ZSL, some form of side information is required, so that the knowledge learned from \emph{seen} classes gets transferred to \emph{unseen} classes. One popular form of side information is attributes~\cite{Lampert2014ZSL} that, however, require costly expert annotation. Thus, there has been a large group of studies~\cite{Mensink2014COSTA,Akata2015OutputEmbedding,Xian2016ZSLLatentEmbedding,Reed2016LDR,Qiao2016LiM,Ding2017} which utilize other auxiliary information, such as, text-based~\cite{Mikolov2013a} or hierarchical model~\cite{Miller1995WN} for label embedding. 

On the other hand, few-shot learning (FSL) refers to the task of recognizing images or detecting objects with a model trained on very few samples~\cite{Xian2019VAEGAND2,Schonfeld2018GenZFSL}. Directly training a given model with small amount of training samples could have the risk of over fitting. Hence a general step to overcome this hurdle is to initially train the model on classes with sufficient examples, and then generalize it to classes with fewer examples without learning any new parameters. This setup already attracted a lot of attention within the computer vision community. One of the first attempts~\cite{Koch2015SiameseOneShot} is a siamese convolutional network model for computing similarity between pair of images, and then the learned similarity was used to solve the one-shot problem by k-nearest neighbors classification. On the other hand, matching network model~\cite{Vinyals2016MatchingNet} uses cosine distance to predict image label based on support sets and apply the episodic training strategy that mimics few-shot learning. An extension, i.e. prototypical network~\cite{Snell2017PrototypNet}, used Euclidean distance instead of cosine distance and built a prototype representation of each class for the few-shot learning scenario. As an orthogonal direction~\cite{Ravi2017FSL} introduced meta-learning framework for FSL, which updates weights of a classifier for a given episode. Model agnostic meta-learner~\cite{Finn2017MAML} learns better weight initialization capable to generalize in FSL scenario with fewer gradient descent steps. There also exist few low shot methods that learn a generator from the base class data to generate novel class features for data augmentation~\cite{Girshick2015,Wang2018LowShotImaginaryData}. Alternatively, GNN~\cite{Kipf2016} was also proposed as a framework for few-shot learning task~\cite{Garcia2018}.

\myparagraph{Our work.} 
The prior work on zero-shot sketch-based image retrieval (ZS-SBIR)~\cite{Shen2018ZSIH}, proposed a generative cross-modal hashing scheme using a graph convolution network for aligning the sketch and image in the semantic space. Inspired by them, \cite{Yelamarthi2018ZSBIR} proposed two similar autoencoder-based generative models for zero-shot SBIR, where they have used the aligned pairs of sketch and image for learning the semantics between them. In this work, we propose a paired cycle consistent generative model where each branch either maps sketch or image features to a common semantic space via adversarial training, which we found to be effective for reducing the domain gap between sketch and image. The cycle consistency constraint on each branch allows supervision only at category level, and avoids the need of aligned sketch-image pairs. Furthermore, we address zero-shot and few-shot cross-modal (sketch to image) retrieval, for that, we effectively combine different side information within an end-to-end framework, and map visual information to the semantic space through an adversarial training. Finally, we unify low-shot learning models and generalize them to fine-grained SBIR scenario.

%

\section{Semantically Aligned Paired Cycle Consistent GAN (SEM-PCYC)}
\label{sec:method}
\begin{figure*}[t]
\centering
\includegraphics[width=\textwidth]{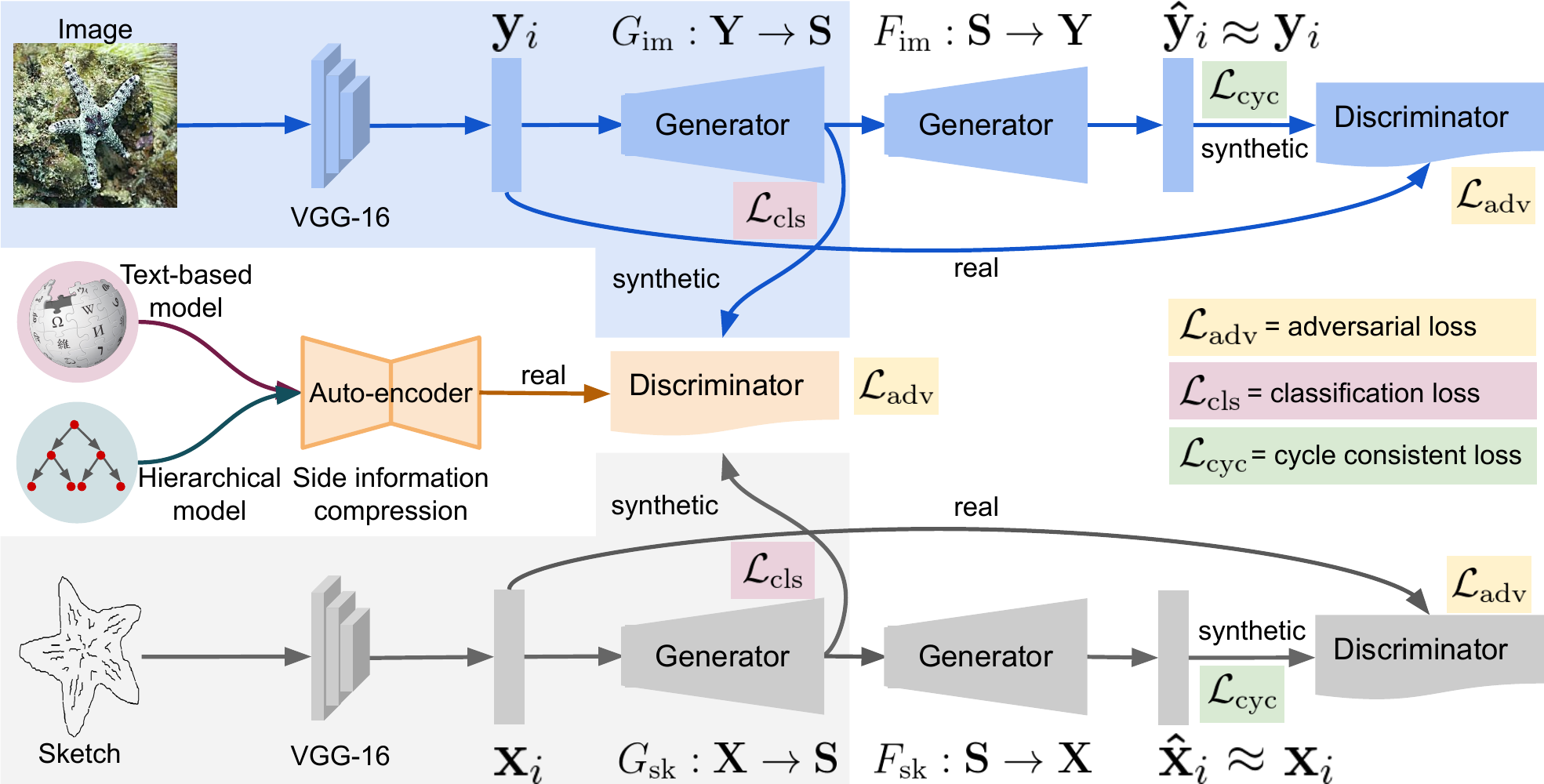}
\caption{Our SEM-PCYC Model. The sketch (in light gray) and image cycle consistent networks (in light blue) respectively map the sketch and image to the semantic space and then the original input space. An auto-encoder (light orange) combines the semantic information based on text and hierarchical model, and produces a compressed semantic representation which acts as a true example to the discriminator. During the test phase only the learned sketch (light gray polygonal region) and image (light blue polygonal region) encoders to the semantic space are used for generating embeddings on the \emph{novel} classes for any-shot, \ie~zero- and few-shot SBIR. (best viewed in color)}
\label{fig:sem-pcyc}
\end{figure*}

Our Semantically Aligned Paired Cycle Consistent GAN (SEM-PCYC) model uses the sketch and image data from the \emph{seen} categories for training the underlying model. It then encodes and matches the sketch and image categories that remain novel during the training phase. The overall pipeline of our end-to-end deep architecture is shown in~\fig{fig:sem-pcyc}.

We define $\mathcal{D}^s=\{\mathbf{X}^s, \mathbf{Y}^s\}$ to be a collection of sketch-image data from the training categories $\mathcal{C}^s$, which contains sketch images $\mathbf{X}^s=\{\mathbf{x}_i^s\}_{i=1}^N$ as well as natural images $\mathbf{Y}^s=\{\mathbf{y}_i^s\}_{i=1}^N$, where $N$ is the total number of sketch and image pairs that are not necessarily aligned. Without loss of generality, a sketch and an image have the same index $i$, and share the same category label. The set $\mathbf{S}^s=\{\mathbf{s}_i^s\}_{i=1}^{N}$ indicates the side information necessary for transferring knowledge from seen to the novel classes (a.k.a unseen classes in zero-shot learning literature). In our setting, we also use an auxiliary training set $\mathcal{D}^a=\{\mathbf{X}^a, \mathbf{Y}^a\}$ from the unseen classes $\mathcal{C}^u$ which is disjoint from $\mathcal{C}^s$, where the number of samples per class is fixed to $k$. 

Our aim is to learn two deep functions $G_\text{sk}(\cdot)$ and $G_\text{im}(\cdot)$ respectively for sketch and image for mapping them to a common semantic space where the learned knowledge is applied to the novel classes. Now, given a second set $\mathcal{D}^u=\{\mathbf{X}^u, \mathbf{Y}^u\}$ from the test categories $\mathcal{C}^u$, the proposed deep networks $G_\text{sk}:\mathbb{R}^d\rightarrow\mathbb{R}^M$, $G_\text{im}:\mathbb{R}^d\rightarrow\mathbb{R}^M$ ($d$ is the dimension of the original data and $M$ is the targeted dimension of the common representation) map the sketch and natural image to a common semantic space where the retrieval is performed. Depending on $k$, \ie~the number of samples considered per class as an auxiliary set, the scenario is called $k$-shot. In the classical zero-shot sketch-based image retrieval setting, the test categories belong to $\mathcal{C}^u$, in other words, at test time the assumption is that every image will come from a previously unseen class. This is not realistic as the true generalization performance of the classifier can only be measured with how well it generalizes to unseen classes without forgetting the classes it has seen. Hence, in the generalized zero-shot sketch based image retrieval scenario the search space contains both $\mathcal{C}^u$ and $\mathcal{C}^s$. In other words, at test time an image may come either from a previously seen or an unseen class. As this setting is significantly more challenging, the accuracy decreases for all the methods considered.

\subsection{Paired Cycle Consistent Generative Model}
\label{ssec:cyc-con-gen-model}

To achieve the flexibility to handle sketch and image individually, \ie~even without aligned sketch-image pairs, during training $G_\text{sk}$ and $G_\text{im}$, we propose a cycle consistent generative model whose each branch is semantically aligned with a common discriminator. The cycle consistency constraint on each branch of the model ensures the mapping of sketch or image modality to a common semantic space, and their translation back to the original modality, which only requires supervision at the category level. Imposing a classification loss on the output of $G_\text{sk}$ and $G_\text{im}$ allows generating highly discriminative features.

Our main goal is to learn two mappings $G_\text{sk}$ and $G_\text{im}$ that can respectively translate the unaligned sketch and natural image to a common semantic space. Zhu~\etal~\cite{Zhu2017CycleGAN} pointed out about the existence of underlying intrinsic relationship between modalities and domains, for example, sketch or image of same object category have the same semantic meaning, and possess that relationship. Even though, we lack visual supervision as we do not have access to aligned pairs, we can exploit semantic supervision at category levels. We train a mapping $G_\text{sk}:\mathbf{X}\rightarrow\mathbf{S}$
so that $\hat{\mathbf{s}}_i=G_\text{sk}(\mathbf{x}_i)$, where $\mathbf{s}_i\in\mathbf{S}$ is the corresponding side information and is indistinguishable from $\hat{\mathbf{s}}_i$ via an adversarial training that classifies $\hat{\mathbf{s}}_i$ different from $\mathbf{s}_i$. The optimal $G_\text{sk}$ thereby translates the modality $\mathbf{X}$ into a modality $\hat{\mathbf{S}}$ which is identically distributed to $\mathbf{S}$. Similarly, another function $G_\text{im}:\mathbf{Y}\rightarrow\mathbf{S}$ can be trained via the same discriminator such that $\hat{\mathbf{s}}_i=G_\text{im}(\mathbf{y}_i)$.

\myparagraph{Adversarial Loss.}
As shown in~\fig{fig:sem-pcyc}, for mapping the sketch and image representation to a common semantic space, we introduce four generators $G_\text{sk}:\mathbf{X}\rightarrow\mathbf{S}$, $G_\text{im}:\mathbf{Y}\rightarrow\mathbf{S}$, $F_\text{sk}:\mathbf{S}\rightarrow\mathbf{X}$ and $F_\text{im}:\mathbf{S}\rightarrow\mathbf{Y}$. In addition, we bring in three adversarial discriminators: $D_\text{se}(\cdot)$, $D_\text{sk}(\cdot)$ and $D_\text{im}(\cdot)$, where $D_\text{se}$ discriminates among original side information $\{\mathbf{s}\}$, sketch transformed to side information $\{G_\text{sk}(\mathbf{x})\}$ and image transformed to side information $\{G_\text{im}(\mathbf{y})\}$; likewise $D_\text{sk}$ discriminates between original sketch representation $\{\mathbf{x}\}$ and side information transformed to sketch representation $\{F_\text{sk}(\mathbf{s})\}$; in a similar way $D_\text{im}$ distinguishes between $\{\mathbf{y}\}$ and $\{F_\text{im}(\mathbf{s})\}$. For the generators $G_\text{sk}$, $G_\text{im}$ and their common discriminator $D_\text{se}$, the objective is:
\begin{align}
&\mathcal{L}_\text{adv}(G_\text{sk}, G_\text{im}, D_\text{se}, \mathbf{x}, \mathbf{y}, \mathbf{s})=2\times\mathbb{E}\left[\log D_\text{se}(\mathbf{s})\right]\nonumber\\
&+\mathbb{E}\left[\log(1- D_\text{se}(G_\text{sk}(\mathbf{x})))\right]+\mathbb{E}\left[\log(1- D_\text{se}(G_\text{im}(\mathbf{y})))\right]
\label{eqn:adv_sem}
\end{align}
where $G_\text{sk}$ and $G_\text{im}$ generate side information similar to the ones in $\mathbf{S}$ while $D_\text{se}$ distinguishes between the generated and original side information. Here, $G_\text{sk}$ and $G_\text{im}$ minimize the objective against an opponent $D_\text{se}$ that tries to maximize it, namely 

$$\min_{G_\text{sk}, G_\text{im}}\max_{D_\text{se}}\mathcal{L}_\text{adv}(G_\text{sk}, G_\text{im}, D_\text{se}, \mathbf{x}, \mathbf{y}, \mathbf{s})$$ 

In a similar way, for the generator $F_\text{sk}$ and its discriminator $D_\text{sk}$, the objective is:

\begin{align}
\mathcal{L}_\text{adv}(F_\text{sk}, D_\text{sk}, \mathbf{x}, \mathbf{s}) = & \mathbb{E}\left[\log D_\text{sk}(\mathbf{x})\right]\nonumber\\ & +\mathbb{E}\left[\log(1- D_\text{sk}(F_\text{sk}(\mathbf{s})))\right] \nonumber
\label{eqn:adv_sk}
\end{align}

$F_\text{sk}$ minimizes the objective and its adversary $D_\text{sk}$ intends to maximize it, namely 

$$\min_{F_\text{sk}}\max_{D_\text{sk}}\mathcal{L}_\text{adv}(F_\text{sk}, D_\text{sk}, \mathbf{x}, \mathbf{s})$$ 
Similarly, another adversarial loss is introduced for the mapping $F_\text{im}$ and its discriminator $D_\text{im}$,~\ie $$\min_{F_\text{im}}\max_{D_\text{im}}\mathcal{L}_\text{adv}(F_\text{im}, D_\text{im}, \mathbf{y}, \mathbf{s})$$

\myparagraph{Cycle Consistency Loss.}
The adversarial mechanism effectively reduces the domain or modality gap, however, it is not guaranteed that an input $\mathbf{x}_i$ and an output $\mathbf{s}_i$ are matched well. To this end, we impose cycle consistency~\cite{Zhu2017CycleGAN}. When we map the feature of a sketch of an object to the corresponding semantic space, and then further translate it back from the semantic space to the sketch feature space, we should reach back to the original sketch feature. This cycle consistency loss also assists in learning mappings across domains where paired or aligned examples are not available. Specifically, if we have a function $G_\text{sk}:\mathbf{X}\rightarrow\mathbf{S}$ and another mapping $F_\text{sk}:\mathbf{S}\rightarrow\mathbf{X}$, then both $G_\text{sk}$ and $F_\text{sk}$ are reverse of each other, and hence form a one-to-one correspondence or bijective mapping.
\begin{equation*}
\begin{split}
\mathcal{L}_\text{cyc}(G_\text{sk}, F_\text{sk})= \mathbb{E}&\left[\Vert F_\text{sk}(G_\text{sk}(\mathbf{x}))-\mathbf{x} \Vert_{1}\right]\\ & +\mathbb{E}\left[\Vert G_\text{sk}(F_\text{sk}(\mathbf{s}))-\mathbf{s} \Vert_{1}\right]
\end{split}
\end{equation*}
where $\mathbf{s}$ is the semantic features of the class $c$ which is the category label of $\mathbf{x}$. Similarly, a cycle consistency loss is imposed for the mappings $G_\text{im}:\mathbf{Y}\rightarrow\mathbf{S}$ and $F_\text{im}:\mathbf{S}\rightarrow\mathbf{Y}$: $\mathcal{L}_\text{cyc}(G_\text{im}, F_\text{im})$. These consistent loss functions also behave as a regularizer to the adversarial training to assure that the learned function maps a specific input $\mathbf{x}_i$ to a desired output $\mathbf{s}_i$.

\myparagraph{Classification Loss.}
On the other hand, adversarial training and cycle-consistency constraints do not explicitly ensure whether the generated features by the mappings $G_\text{sk}$ and $G_\text{im}$ are class discriminative, \ie~a requirement for the zero-shot sketch-based image retrieval task. We conjecture that this issue can be alleviated by introducing a discriminative classifier pre-trained on the input data. At this end we  minimize a classification loss over the generated features.
\begin{align*}
\mathcal{L}_\text{cls}(G_\text{sk})=-\mathbb{E}_{\mathbf{x}\sim\mathbf{X}}\left[\log P(c|G_\text{sk}(\mathbf{x});\theta) \right]
\end{align*}
where $c$ is the category label of $\mathbf{x}$, $P(c|G_\text{sk}(\mathbf{x});\theta)$ denotes the probability of $G_\text{sk}(\mathbf{x}$) being predicted with its true class label $c$. The conditional probability is computed by a linear softmax classifier parameterized by $\theta$. Similarly, a classification loss $\mathcal{L}_\text{cls}(G_\text{im})$ is also imposed on the generator $G_\text{im}$.

\subsection{Selection of Side Information}
\label{ssec:autoenc}
Learning a compatibility or a matching function between multiple modalities in zero-shot scenario~\cite{Shen2018ZSIH,Dey2019doodle2search,Liu2019SKP} requires structure in the class embedding space where the image features are mapped to. Attributes provide one such a structured class embedding space~\cite{Lampert2014ZSL}, however obtaining attributes requires costly human annotation. On the other hand, side information can also be learned at a much lower cost from large-scale text corpora such as Wikipedia. Similarly, output embeddings built from hierarchical organization of classes such as WordNet can also provide structure in the output space and substitute the attributes. Motivated by attribute selection for zero-shot learning~\cite{Guo2018ZSL}, indicating that a subset of discriminative attributes are more effective than the whole set of attributes for ZSL, we incorporate a joint learning framework integrating an auto-encoder to select side information. Let $\mathbf{s}\in\mathbb{R}^{k}$ be the side information with $k$ as the original dimension. The loss function is: 
\begin{align}
\mathcal{L}_\text{aenc}(f,g)=\Vert \mathbf{s}-g(f(\mathbf{s}))\Vert_{F}+\lambda \Vert W_1 \Vert_{2,1}
\label{eqn:aenc_loss}
\end{align}
where $f(\mathbf{s})=\sigma(W_1\mathbf{s}+b_1)$, $g(f(\mathbf{s}))=\sigma(W_2f(\mathbf{s})+b_2)$, with $W_1\in\mathbb{R}^{k\times m}$, $W_2\in\mathbb{R}^{m\times k}$ and $b_1$, $b_2$ respectively as the weights and biases for the function $f$ and $g$. Additionally, $\Vert.\Vert_{F}$ denotes the Frobenius norm defined as the square root of the sum of the absolute squares of its elements and $\Vert.\Vert_{2,1}$ indicates $\ell_{2,1}$ norm~\cite{Nie2010}. Selecting side information reduces the dimensionality of embeddings, which further improves retrieval time. Therefore, the training objective of our model:
\begin{align}
&\mathcal{L}(G_\text{sk}, G_\text{im}, F_\text{sk}, F_\text{im}, D_\text{se}, D_\text{sk}, D_\text{im}, f, g, \mathbf{x}, \mathbf{y}, \mathbf{s})\nonumber\\
&=\lambda_\text{adv}^\text{se}\mathcal{L}_\text{adv}(G_\text{sk}, G_\text{im}, D_\text{se}, \mathbf{x}, \mathbf{y}, \mathbf{s})\nonumber\\ &+\lambda_\text{adv}^\text{sk}\mathcal{L}_\text{adv}(F_\text{sk}, D_\text{sk}, \mathbf{x}, \mathbf{s})+\lambda_\text{adv}^\text{im}\mathcal{L}_\text{adv}(F_\text{im}, D_\text{im}, \mathbf{y}, \mathbf{s})\nonumber\\ &+\lambda_\text{cyc}^\text{sk}\mathcal{L}_\text{cyc}(G_\text{sk}, F_\text{sk})\nonumber+\lambda_\text{cyc}^\text{im}\mathcal{L}_\text{cyc}(G_\text{im}, F_\text{im})\nonumber\\ &+\lambda_\text{cls}^\text{sk}\mathcal{L}_\text{cls}(G_\text{sk})+\lambda_\text{cls}^\text{im}\mathcal{L}_\text{cls}(G_\text{im})+\lambda_\text{aenc}\mathcal{L}_\text{aenc}(f,g)
\label{eqn:comb-loss}
\end{align}
where different $\lambda$s are the weights on respective loss terms. For obtaining the initial side information, we combine a text-based and a hierarchical model, which are complementary and robust~\cite{Akata2015OutputEmbedding}. Below, we provide a description of our text-based and hierarchical models for side information.

\myparagraph{Text-based Model.}
We use three different text-based side information. (1) Word2Vec \cite{Mikolov2013} is a two layered neural network that are trained to reconstruct linguistic contexts of words. During training, it takes a large corpus of text and creates a vector space of several hundred dimensions, with each unique word being assigned to a corresponding vector in that space. The model can be trained with a hierarchical softmax with either skip-gram or continuous bag-of-words formulation for target prediction. (2) GloVe~\cite{Pennington2014GloVe} considers global word-word co-occurrence statistics that frequently appear in a corpus. Intuitively, co-occurrence statistics encode important semantic information. The objective is to learn word vectors such that their dot product equals to the probability of their co-occurrence. (3) FastText \cite{Joulin2016FastText} extends the Word2Vec model, where instead of learning vector for words directly, FastText represents each word as n-gram of characters and then trains a skip-gram model to learn the embeddings. FastText works well with rare words, even if a word was not seen during training, it can be broken down into n-grams to get its embeddings, which is a huge advantage of this model.

\myparagraph{Hierarchical Model.}
Semantic distance (or similarity) between words can also be approximated by their distance (or similarity) in a large ontology such as WordNet\footnote{\url{https://wordnet.princeton.edu}} with $\approx 100,000$ words in English. One can measure the similarity ($\mathcal{S}_\text{WN}$ in \eq{eqn:hier_emb}) between words represented as nodes in the ontology using techniques, such as \emph{path similarity}, e.g. counting the number of hops required to reach from one node to the other, and Jiang-Conrath~\cite{Jiang1997SemSim}. For a set $\mathbb{S}$ of nodes in a dictionary $\mathbb{D}$ that consists of a set of classes, similarities between every class $c$ and all the other nodes considered in the same order in $\mathbb{S}$ to determine the entries of the class embedding vector~\cite{Akata2015OutputEmbedding} of $c$ ($\mathbf{s}_\text{hier}(c)$ in \eq{eqn:hier_emb}):
\begin{align}
  \mathbf{s}_\text{hier}(c) = [\mathcal{S}_\text{WN}(c,c_1), \ldots, \mathcal{S}_\text{WN}(c,c_{|\mathbb{S}|})]
  \label{eqn:hier_emb}
\end{align}
Note that, $\mathbb{S}$ considers all the nodes on the path from each node in $\mathbb{D}$ to its highest level ancestor. The WordNet hierarchy contains most of the classes of the Sketchy~\cite{Sangkloy2016}, Tu-Berlin~\cite{Eitz2012TUBerlin} and QuickDraw~\cite{Dey2019doodle2search} datasets. Few exceptions are: \emph{jack-o-lantern} which we replaced with \emph{lantern} that appears higher in the hierarchy, similarly \emph{human skeleton} with \emph{skeleton}, and \emph{octopus} with \emph{octopods} etc. $|\mathbb{S}|$, i.e. the number of nodes, for Sketchy, TU-Berlin and QuickDraw datasets are respectively $354$, $664$ and $344$.

\section{Experiments}
\label{sec:expt}

In this section, we detail our datasets, implementation protocol and present our results on (generalized) zero-shot, (generalized) few-shot and fine-grained settings. 

\myparagraph{Datasets.} We experimentally validate our model on three popular SBIR datasets, namely Sketchy (Extended), TU-Berlin (Extended) and QuickDraw (Extended). For brevity, we refer to these extended datasets as Sketchy, TU-Berlin and QuickDraw respectively.

The Sketchy Dataset~\cite{Sangkloy2016} is a large collection of sketch-photo pairs. The dataset originally consists of images from $125$ different classes, with $100$ photos each. The $75,471$ sketch images of the objects that appear in these $12,500$ images are collected via crowd sourcing. This dataset also contains a fine grained correspondence (alignment) between particular photos and sketches as well as various data augmentations for deep learning based methods. Liu~\etal~\cite{Liu2017DSH} extended the dataset by adding $60,502$ photos yielding in total $73,002$ images. We randomly pick $25$ classes as the \emph{novel} test set, and the data from remaining $100$ \emph{training} classes.

The original TU-Berlin Dataset~\cite{Eitz2012TUBerlin} contains $250$ categories with a total of $20,000$ sketches extended by~\cite{Liu2017DSH} with $204,489$ natural images corresponding to the sketch classes. $30$ classes of sketches and images are randomly chosen to respectively from the query set and the retrieval gallery. The remaining $220$ classes are utilized for training. We follow Shen~\etal~\cite{Shen2018ZSIH} and select classes with at least $400$ images to form a test set.

The QuickDraw (Extended), a large-scale dataset proposed recently in~\cite{Dey2019doodle2search}, contains the sketch-image pairs of $110$ classes consisting of $203,885$ images and $330,111$ sketches, \ie approximately $1854$ images/class and $3000$ sketches/class. The main difference of this dataset from the previous ones is in the abstractness of the sketches which are collected from the \emph{Quick, Draw!}\footnote{\url{https://quickdraw.withgoogle.com}} online game. The increased abstractness in the drawings has eventually enlarged the sketch-image domain gap, and hence increased the challenge of SBIR task.

\myparagraph{Implementation details.}
We implemented the SEM-PCYC model using PyTorch~\cite{Paszke2017PyTorch} deep learning toolbox\footnote{Our code and models are available at: \url{https://github.com/AnjanDutta/sem-pcyc-ijcv}} on a single TITAN Xp or TITAN V graphics card. Unless otherwise mentioned, we extract features from sketch and image from the VGG-$16$~\cite{Simonyan2014} network model pre-trained on ImageNet~\cite{Deng2009ImageNet} (before the last pooling layer). In \sect{ssec:gzs-sbir}, we compare the VGG-16 features with SE-ResNet-50 features for zero-shot SBIR task, which is only restricted to that experimentation. Since in this work, we deal with single object retrieval and an object usually spans only on certain regions of a sketch or image, we apply an attention mechanism inspired by Song~\etal~\cite{Song2017SpatSemAttn} without the shortcut connection for extracting only the informative regions from sketch and image. The attended $512$d representation is obtained by a pooling operation guided by the attention model and fully connected (fc) layer. This entire model is fine tuned on our training set ($100$ classes for Sketchy, $220$ classes for TU-Berlin and $80$ classes for QuickDraw). Both the generators $G_\text{sk}$ and $G_\text{im}$ are built with a fc layer followed by a ReLU non-linearity that accept $512$d vector and output $M$d representation, whereas, the generators $F_\text{sk}$ and $F_\text{im}$ take $M$d features and produce $512$d vector. Accordingly, all discriminators are designed to take the output of respective generators and produce a single dimensional output. The auto-encoder is designed by stacking two non-linear fc layers respectively as encoder and decoder for obtaining a compressed and encoded representation of dimension $M$. We experimentally set $\lambda_\text{adv}^\text{se}=1.0$, $\lambda_\text{adv}^\text{sk}=0.5$, $\lambda_\text{adv}^\text{im}=0.5$, $\lambda_\text{cyc}^\text{sk}=1.0$, $\lambda_\text{cyc}^\text{im}=1.0$, $\lambda_\text{cls}^\text{sk}=1.0$, $\lambda_\text{cls}^\text{im}=1.0$, $\lambda_\text{aenc}=0.01$ to give the optimum performance of our model.

While constructing the hierarchy for the class embedding, we only consider the \emph{training} classes belong to that dataset. In this way, the WordNet hierarchy or the knowledge graph for the Sketchy, TU-Berlin and QuickDraw datasets respectively contain $354$ and $664$ nodes. Although our method does not produce binary hash code as a final representation for matching sketch and image, for the sake of comparison with some related works, such as, ZSH~\cite{Yang2016ZSH}, ZSIH~\cite{Shen2018ZSIH}, GDH~\cite{Zhang2018GDH}, that produce hash codes, we have used the iterative quantization (ITQ)~\cite{Gong2013ITQ} algorithm to obtain the binary codes for sketch and image. We have used final representation of sketches and images from the train set to learn the optimized rotation which later used on our final representation for obtaining the binary codes.

\subsection{(Generalized) Zero-Shot Sketch-based Image Retrieval}
\label{ssec:gzs-sbir}

Apart from the two prior Zero-Shot SBIR works closest to ours, \ie~ZSIH~\cite{Shen2018ZSIH} and ZS-SBIR~\cite{Yelamarthi2018ZSBIR}, we adopt fourteen ZSL and SBIR models to the zero-shot SBIR task. 
Note that in this setting, the training classes are indicated as ``seen'' and novel classes as ``unseen'' since none of the sketches of these classes are visible to the model during training.

The SBIR methods that we evaluate are SaN~\cite{Yu2015}, 3D Shape~\cite{Wang2015a}, Siamese CNN \cite{Qi2016SBIRSiamese}, GN Triplet~\cite{Sangkloy2016}, DSH~\cite{Liu2017DSH} and GDH~\cite{Zhang2018GDH}. A softmax baseline is also added, which is based on computing the $4096$d VGG-$16$~\cite{Simonyan2014} feature vector pre-trained on the \emph{seen} classes for nearest neighbour search. The ZSL methods that we evaluate are: CMT~\cite{Socher2013ZSLCrossModalT}, DeViSE \cite{Frome2013Devise}, SSE~\cite{Zhang2015ZSLSemSim}, JLSE~\cite{Zhang2016ZSLJointLatentSim}, ZSH~\cite{Yang2016ZSH}, SAE~\cite{Kodirov2017SAE} and FRWGAN \cite{Felix2018FRWGAN}.
We use the same \emph{seen}-\emph{unseen} splits of categories for all the experiments for a fair comparison. We compute the mean average precision (mAP@all) and precision considering top $100$ (Precision@100)~\cite{Su2015PerfEvalIR,Shen2018ZSIH} retrievals for the performance evaluation and comparison.

{
 \setlength{\tabcolsep}{2pt}
 \renewcommand{\arraystretch}{1.2} 
\begin{table*}[!t]
\centering
\resizebox{\textwidth}{!}{
\begin{tabular}{l l cccc|cccc|cccc}
\hline
  &  & \multicolumn{4}{c|}{\textbf{Sketchy (Extended)}} & \multicolumn{4}{c|}{\textbf{TU-Berlin (Extended)}} & \multicolumn{4}{c}{\textbf{QuickDraw (Extended)}} \\
 & \textbf{Method} & \textbf{mAP} & \textbf{Prec.} & \textbf{Feat.} & \textbf{Retr.} & \textbf{mAP} & \textbf{Prec.} & \textbf{Feat.} & \textbf{Retr.} & \textbf{mAP} & \textbf{Prec.} & \textbf{Feat.} & \textbf{Retr.} \\
  &  & \textbf{@all} & \textbf{@100} & \textbf{dim} & \textbf{Time (s)} & \textbf{@all} & \textbf{@100} & \textbf{dim} & \textbf{Time (s)} & \textbf{@all} & \textbf{@100} & \textbf{dim} & \textbf{Time (s)} \\
\hline
\multirow{7}{*}{SBIR} & Softmax Baseline & $0.114$ & $0.172$ & $4096$ & $3.5\times10^{-1}$ & $0.089$ & $0.143$ & $4096$ & $4.3\times10^{-1}$ & $0.058$ & $0.095$ & $4096$ & $4.6\times10^{-1}$ \\
  & Siamese CNN~\cite{Qi2016SBIRSiamese} & $0.132$ & $0.175$ & $64$ & $5.7\times10^{-3}$ & $0.109$ & $0.141$ & $64$ & $5.9\times10^{-3}$ & $0.074$ & $0.112$ & $64$ & $5.8\times10^{-3}$ \\
  & SaN~\cite{Yu2016a} & $0.115$ & $0.125$ & $512$ & $4.8\times10^{-2}$ & $0.089$ & $0.108$ & $512$ & $5.5\times10^{-2}$ & $0.060$ & $0.093$ & $512$ & $5.9\times10^{-2}$ \\
  & GN Triplet~\cite{Sangkloy2016} & $0.204$ & $0.296$ & $1024$ & $9.1\times10^{-2}$ & $0.175$ & $0.253$ & $1024$ & $1.9\times10^{-1}$ & $0.118$ & $0.142$ & $1024$ & $2.3\times10^{-1}$ \\
  & 3D Shape~\cite{Wang2015} & $0.067$ & $0.078$ & $64$ & $7.8\times10^{-3}$ & $0.054$ & $0.067$ & $64$ & $7.2\times10^{-3}$ & $0.036$ & $0.081$ & $64$ & $8.1\times10^{-1}$ \\
  & DSH (binary)~\cite{Liu2017DSH} & $0.171$ & $0.231$ & $64$ & $6.1\times10^{-5}$ & $0.129$ & $0.189$ & $64$ & $7.2\times10^{-5}$ & $0.087$ & $0.127$ & $64$ & $7.6\times10^{-5}$ \\
  & GDH (binary)~\cite{Zhang2018GDH} & $0.187$ & $0.259$ & $64$ & $7.8\times10^{-5}$ & $0.135$ & $0.212$ & $64$ & $9.6\times10^{-5}$ & $0.095$ & $0.146$ & $64$ & $1.1\times10^{-4}$ \\
\hline
\multirow{7}{*}{ZSL} & CMT~\cite{Socher2013ZSLCrossModalT} & $0.087$ & $0.102$ & $300$ & $2.8\times10^{-2}$ & $0.062$ & $0.078$ & $300$ & $3.3\times10^{-2}$ & $0.036$ & $0.062$ & $300$ & $3.6\times10^{-2}$ \\
  & DeViSE~\cite{Frome2013Devise} & $0.067$ & $0.077$ & $300$ & $3.6\times10^{-2}$ & $0.059$ & $0.071$ & $300$ & $3.2\times10^{-2}$ & $0.034$ & $0.073$ & $300$ & $3.4\times10^{-2}$ \\
  & SSE~\cite{Zhang2015BitScalable} & $0.116$ & $0.161$ & $100$ & $1.3\times10^{-2}$ & $0.089$ & $0.121$ & $220$ & $1.7\times10^{-2}$ & $0.051$ & $0.093$ & $80$ & $1.8\times10^{-2}$ \\
  & JLSE~\cite{Zhang2016ZSLJointLatentSim} 	& $0.131$ & $0.185$ & $100$ & $1.5\times10^{-2}$ & $0.109$ & $0.155$ & $220$ & $1.4\times10^{-2}$ & $0.063$ & $0.084$ & $80$ & $1.5\times10^{-2}$ \\
  & SAE~\cite{Kodirov2017SAE} & $0.216$ & $0.293$ & $300$ & $2.9\times10^{-2}$ & $0.167$ & $0.221$ & $300$ & $3.2\times10^{-2}$ & $0.096$ & $0.112$ & $300$ & $3.3\times10^{-2}$ \\
  & FRWGAN~\cite{Felix2018FRWGAN} & $0.127$ & $0.169$ & $512$ & $3.2\times10^{-2}$ & $0.110$ & $0.157$ & $512$ & $3.9\times10^{-2}$ & $0.064$ & $0.093$ & $512$ & $4.2\times10^{-2}$ \\
  & ZSH (binary)~\cite{Yang2016} & $0.159$ & $0.214$ & $64$  & $5.9\times10^{-5}$ & $0.141$ & $0.177$ & $64$ & $7.6\times10^{-5}$ & $0.081$ & $0.118$ & $64$ & $7.8\times10^{-5}$ \\
\hline 
\multirow{4}{*}{\makecell{Zero-Shot \\SBIR}} & ZSIH (binary)~\cite{Shen2018ZSIH} & $0.258$ & $0.342$ & $64$ & $6.7\times10^{-5}$ & $0.223$ & $0.294$ & $64$ & $7.7\times10^{-5}$ & $0.131$ & $0.188$ & $64$ & $7.9\times10^{-5}$ \\
  & ZS-SBIR~\cite{Yelamarthi2018ZSBIR} & $0.196$ & $0.284$ & $1024$ & $9.6\times10^{-2}$ & $0.005$ & $0.001$ & $1024$ & $1.2\times10^{-1}$ & $0.006$ & $0.001$ & $1024$ & $1.6\times10^{-1}$ \\
  & \textbf{SEM-PCYC} & $\mathbf{0.349}$ & $\mathbf{0.463}$ & $64$ & $1.7\times 10^{-3}$ & $\mathbf{0.297}$ & $\mathbf{0.426}$ & $64$ & $1.9\times 10^{-3}$ & $\mathbf{0.177}$ & $\mathbf{0.255}$ & $64$ & $2.1\times 10^{-3}$ \\
  & \textbf{SEM-PCYC (binary)} & $\mathbf{0.344}$ & $\mathbf{0.399}$ & $64$ & $9.5 \times 10^{-5}$ & $\mathbf{0.293}$ & $\mathbf{0.392}$ & $64$ & $9.3\times 10^{-4}$ & $\mathbf{0.164}$ & $\mathbf{0.243}$ & $64$ & $9.6\times 10^{-4}$ \\
\hline
\multirow{4}{*}{\makecell{Generalized\\Zero-Shot \\SBIR}} & ZSIH (binary)~\cite{Shen2018ZSIH} & $0.219$ & $0.296$ & $64$ & $6.7\times10^{-5}$ & $0.142$ & $0.218$ & $64$ & $7.7\times10^{-5}$ & $0.130$ & $0.163$ & $64$ & $8.1\times10^{-5}$ \\
 & ZS-SBIR~\cite{Yelamarthi2018ZSBIR} & $0.146$ & $0.190$ & $1024$ & $7.8\times10^{-2}$ & $0.003$ & $0.001$ & $1024$ & $6.7\times10^{-2}$ & $0.002$ & $0.001$ & $1024$ & $8.2\times10^{-2}$ \\
 & \textbf{SEM-PCYC} & $\mathbf{0.307}$ & $\mathbf{0.364}$ & $64$ & $1.7\times 10^{-3}$ & $\mathbf{0.192}$ & $\mathbf{0.298}$ & $64$ & $2.0\times 10^{-3}$ & $\mathbf{0.140}$ & $\mathbf{0.221}$ & $64$ & $2.1\times 10^{-4}$ \\
 & \textbf{SEM-PCYC (binary)} & $\mathbf{0.260}$ & $\mathbf{0.317}$ & $64$ & $9.4 \times 10^{-5}$ & $\mathbf{0.174}$ & $\mathbf{0.267}$ & $64$ & $9.3\times 10^{-4}$ & $\mathbf{0.135}$ & $\mathbf{0.216}$ & $64$ & $9.4\times 10^{-4}$ \\
\hline
\end{tabular}
}
\caption{(Generalized) Zero-Shot Sketch-based Image Retrieval and (Generalized) Fine-Grained Sketch-based Image Retrieval performance comparison with existing SBIR, ZSL, zero-shot SBIR and generalized zero-shot SBIR methods. Note: SBIR and ZSL methods are adapted to the Zero-Shot SBIR task, same \emph{seen} and \emph{unseen} classes are used for a fair comparison. }
\label{tab:results_zs-sbir}
\end{table*}
}

\begin{figure*}[!t]
\resizebox{\textwidth}{!}{
\begin{tabular}{cccc}
\includegraphics[width=5cm,height=5cm]{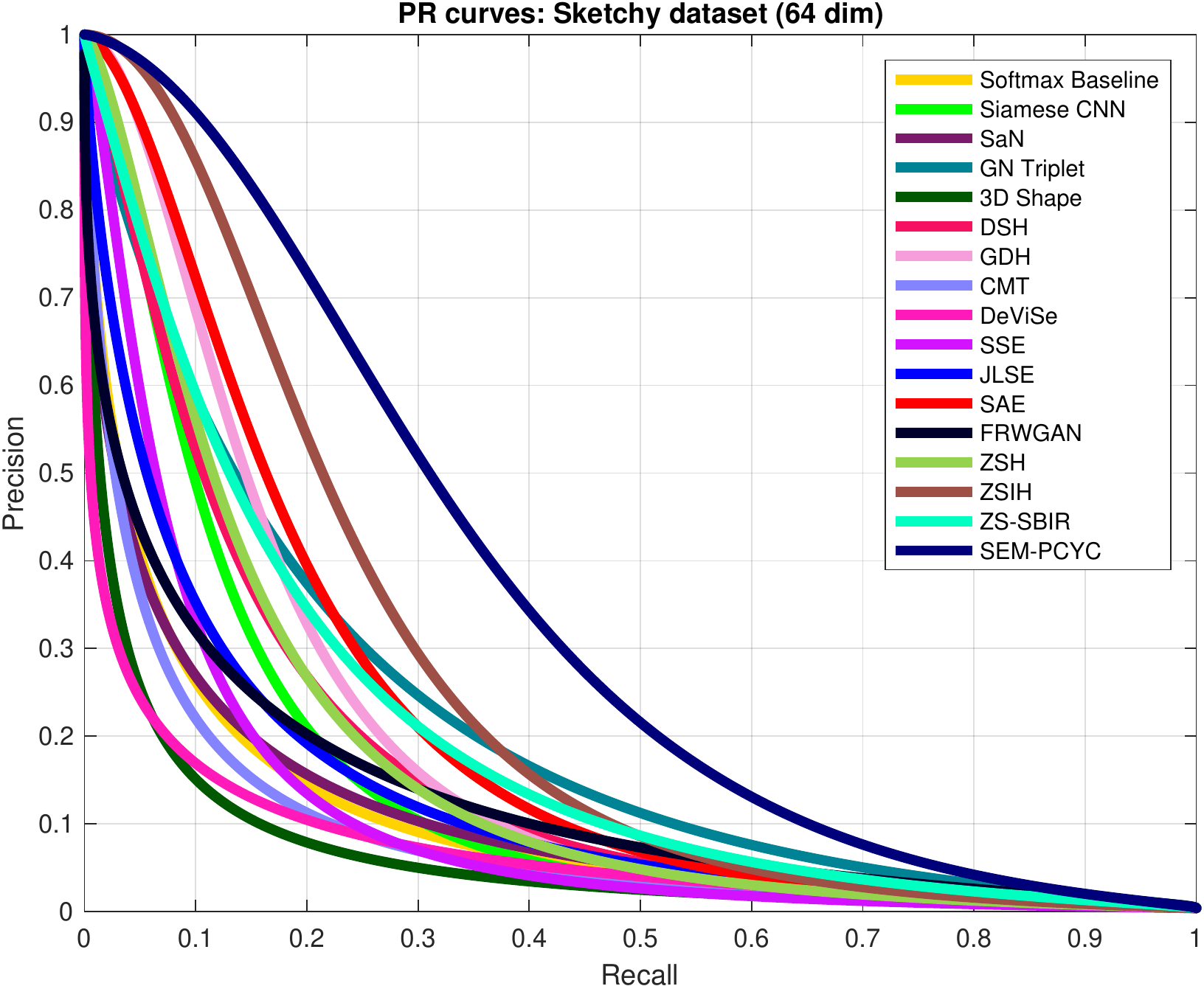} & \includegraphics[width=5cm,height=5cm]{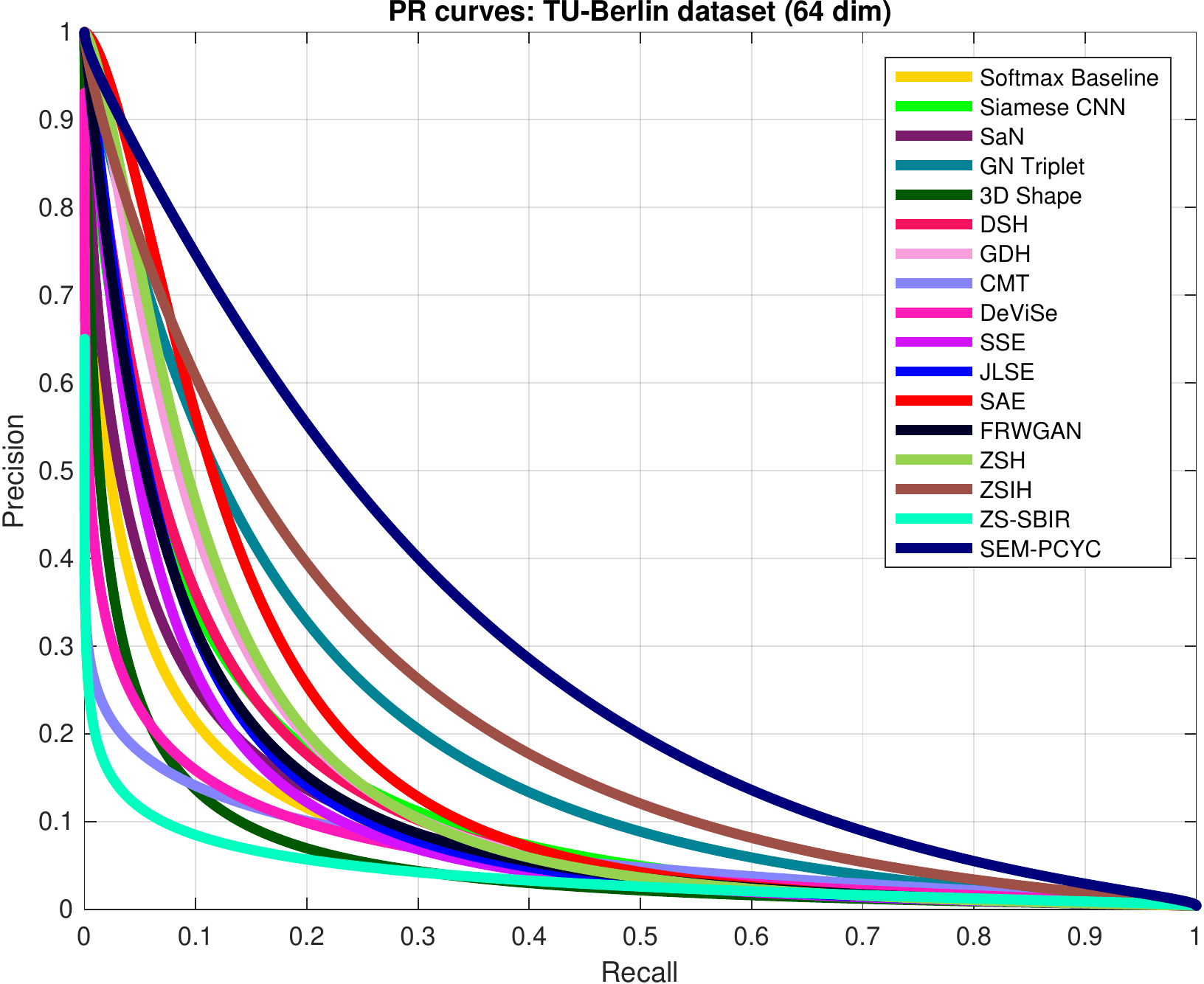} & \includegraphics[width=5cm,height=5cm]{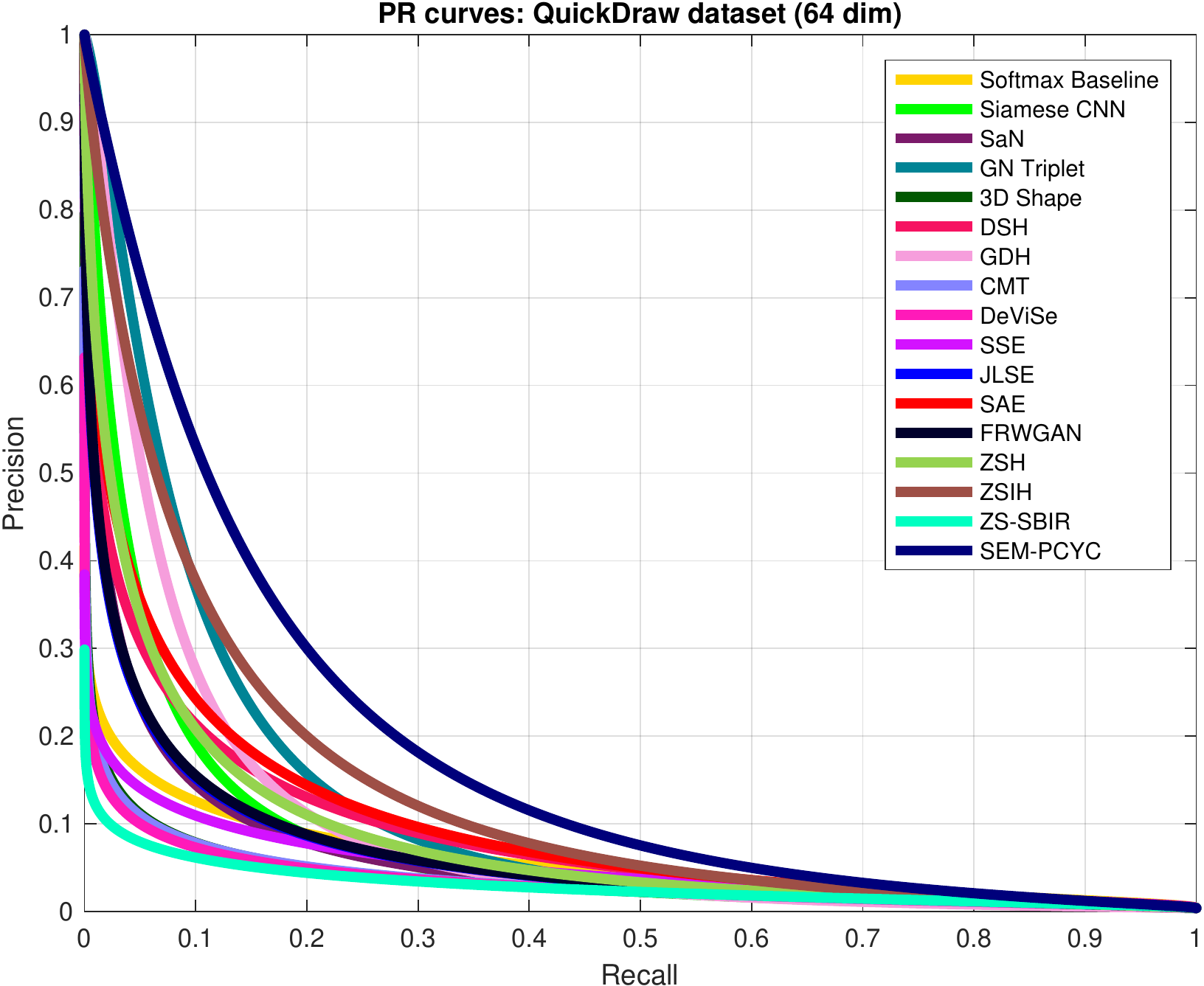} & \includegraphics[width=5cm,height=5cm]{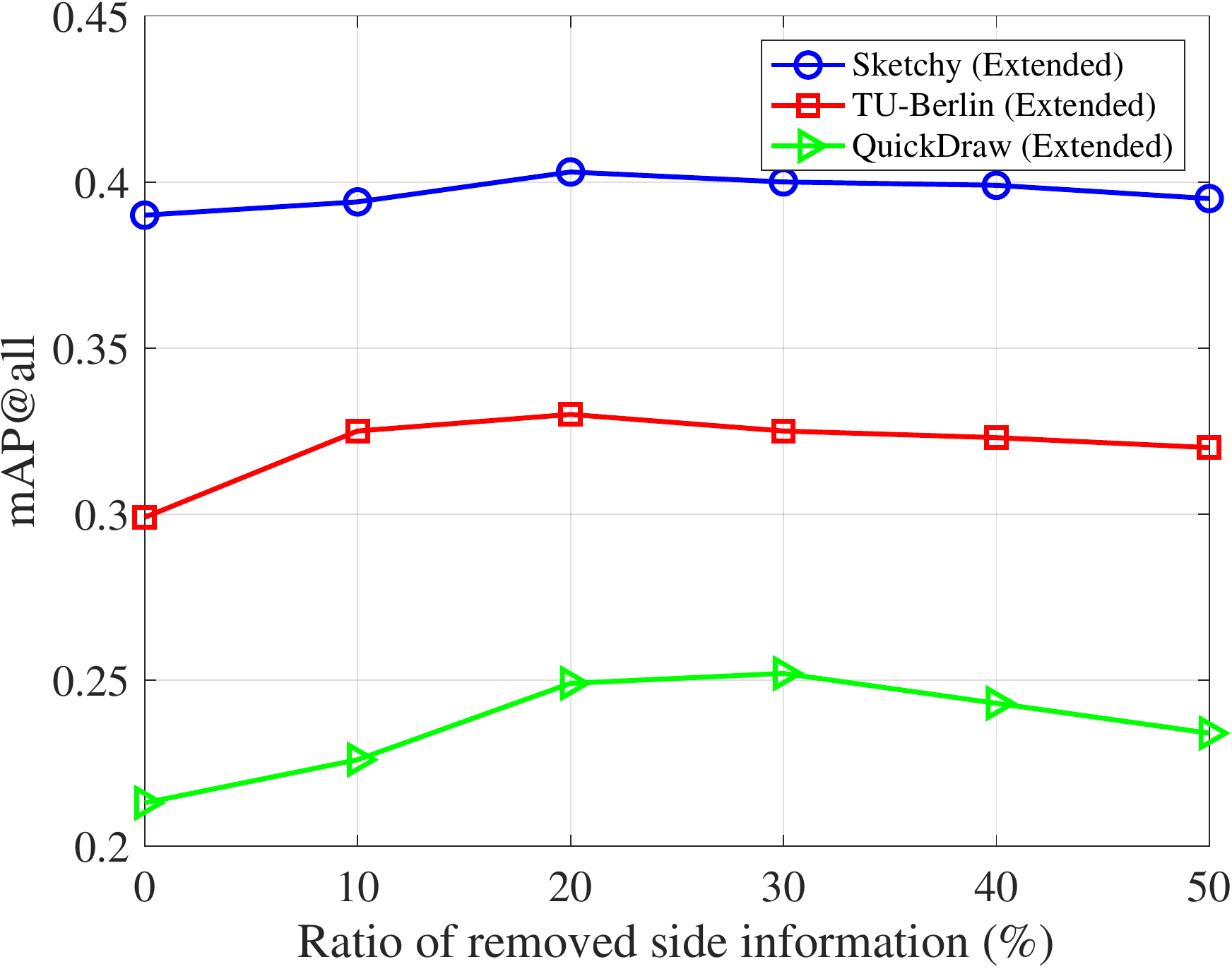} \\
(a) & (b) & (c) & (d)
\end{tabular}}
\caption{(a)-(c) PR curves of SEM-PCYC model and several SBIR, ZSL and zero-shot SBIR methods respectively on the Sketchy, TU-Berlin and QuickDraw datasets, (d) Plot showing mAP@all wrt the ratio of removed side information. (best viewed in color)}
\label{fig:plots}
\end{figure*}

\tab{tab:results_zs-sbir} shows that most of the SBIR and ZSL methods perform worse than the zero-shot SBIR methods. Among them, the ZSL methods usually suffer from the domain gap between the sketch and image modalities. The majority SBIR methods although have performed better than their ZSL counterparts, fail to generalize the learned representations to \emph{unseen} classes. However, GN Triplet~\cite{Sangkloy2016}, DSH~\cite{Liu2017DSH}, GDH~\cite{Zhang2018GDH} have shown reasonable potential to generalize information only from object with common shape.

As per the expectation, the specialized zero-shot SBIR methods have surpassed most of the ZSL and SBIR baselines as they possess both the ability of reducing the domain gap and generalizing the learned information for the \emph{unseen} classes. ZS-SBIR learns to generalize between sketch and image from the aligned sketch-image pairs, as a result it performs well on the Sketchy dataset, but not on the TU-Berlin or QuickDraw datasets, as in these datasets, aligned sketch-image pairs are not available.  
Our proposed method has excels the state-of-the-art method by $0.091$ mAP@all on the Sketchy, $0.074$ mAP@all on the TU-Berlin and $0.046$ mAP@all on the QuickDraw, which shows the effectiveness of our proposed SEM-PCYC model due to the cycle consistency between sketch, image and semantic space, as well as the compact and discriminative side information. 

In general, the main challenge in TU-Berlin dataset is the large number of visually similar and overlapping classes. On the other hand, in QuickDraw datatset there is a the large domain gap that is intentionally introduced for designing future realistic models. Also, the ambiguity in annotation, e.g. non-professional sketches, is a major challenge in this dataset. Although our results are encouraging in that they show that the cycle consistency helps zero-shot SBIR task and our model sets the new state-of-the-art in this domain, we hope that our work will encourage further research in improving these results. 

Finally, the PR-curves of SEM-PCYC and considered baselines on Sketchy, TU-Berlin and QuickDraw are respectively shown in~\fig{fig:plots}(a)-(c) which show that the precision-recall curves correspond to our SEM-PCYC model (dark blue line) are always plotted above the other methods. This indicates that our proposed model consistently exhibits the superiority on all three datasets, which clearly show the benefit of our proposal.

\myparagraph{Generalized Zero-Shot Sketch-based Image Retrieval.} 
We conducted experiments on generalized ZS-SBIR setting where search space contains both \emph{seen} and \emph{unseen} classes. This task is significantly more challenging than ZS-SBIR as \emph{seen} classes create distraction to the test queries. Our results in \tab{tab:results_zs-sbir} show that our model significantly outperforms both the existing models~\cite{Shen2018ZSIH,Yelamarthi2018ZSBIR}, due to the benefit of our cross-modal adversarial mechanism and heterogeneous side information.

\begin{figure}[!t]
\centering
\resizebox{\columnwidth}{!}{
\begin{tabular}{@{}c@{}c@{}c@{}c@{}c}
\includegraphics[width=0.25\columnwidth]{swan} &
\includegraphics[width=0.25\columnwidth]{duck} & \includegraphics[width=0.25\columnwidth]{owl} & 
\includegraphics[width=0.25\columnwidth]{penguin} & 
\includegraphics[width=0.25\columnwidth]{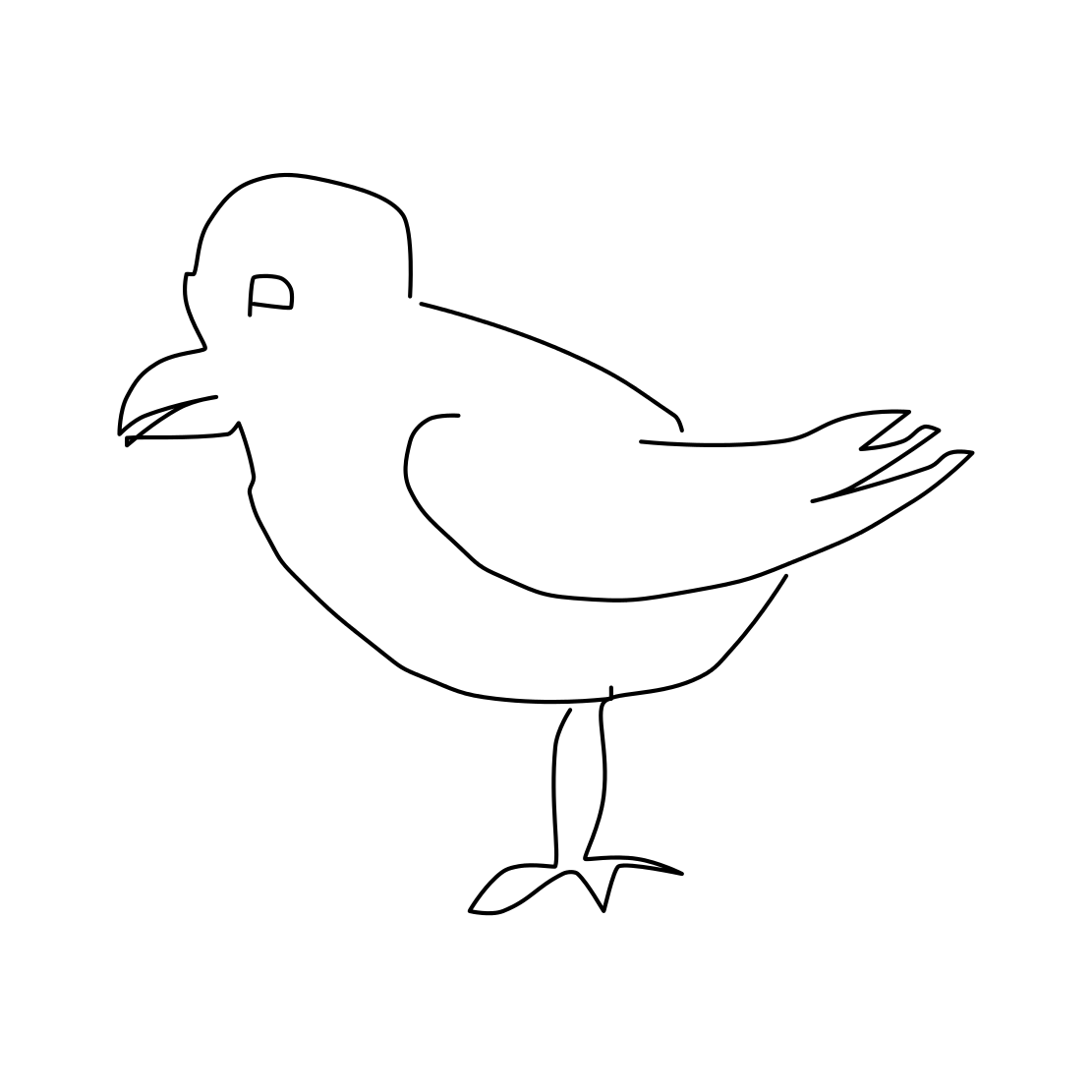} \\
\texttt{swan} & \texttt{duck} & \texttt{owl} & \texttt{penguin} & \texttt{standing bird}
\end{tabular}
}
\caption{Inter-class similarity in TU-Berlin dataset may indicate the challenge of the task.}
\label{tab:sketches_tu_berlin}
\end{figure}

\begin{figure*}[!t]
\begin{center}
\resizebox{\textwidth}{!}{
\begin{tabular}{@{}c@{}c@{}c@{}c@{}c@{}c@{}c@{}c@{}c@{}c@{}c@{}c@{}c@{}c@{}c@{}c@{}c@{}c@{}c@{}c@{}c}
\includegraphics[width=1.5cm, height=1.5cm]{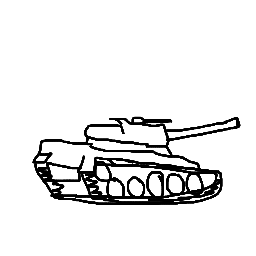} & \includegraphics[width=1.5cm, height=1.5cm]{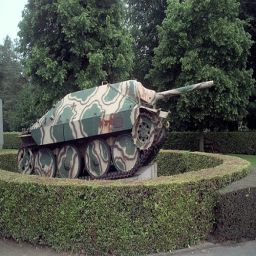} & \includegraphics[width=1.5cm, height=1.5cm]{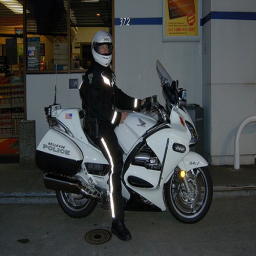} & \includegraphics[width=1.5cm, height=1.5cm]{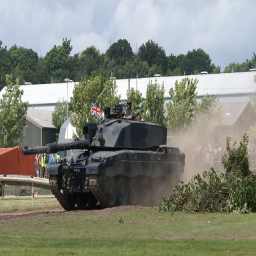} & \includegraphics[width=1.5cm, height=1.5cm]{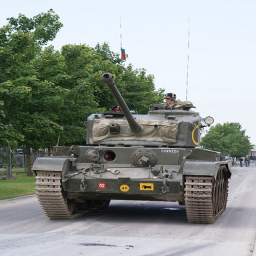} & \includegraphics[width=1.5cm, height=1.5cm]{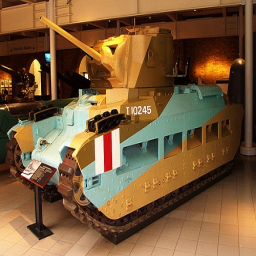} & \includegraphics[width=1.5cm, height=1.5cm]{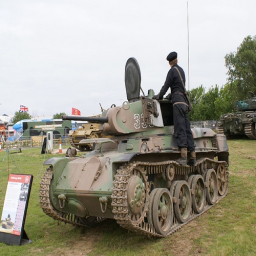} & \includegraphics[width=1.5cm, height=1.5cm]{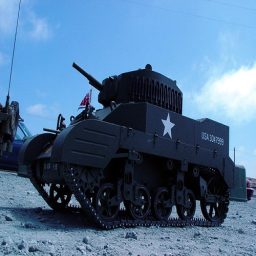} & \includegraphics[width=1.5cm, height=1.5cm]{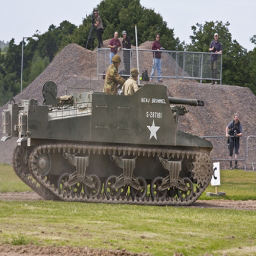} & \includegraphics[width=1.5cm, height=1.5cm]{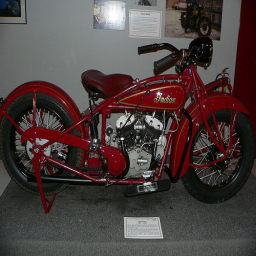} & \includegraphics[width=1.5cm, height=1.5cm]{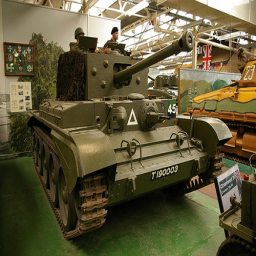} & \includegraphics[width=1.5cm, height=1.5cm]{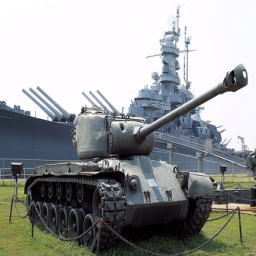} & \includegraphics[width=1.5cm, height=1.5cm]{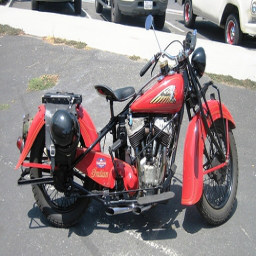} & \includegraphics[width=1.5cm, height=1.5cm]{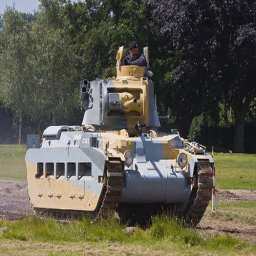} & \includegraphics[width=1.5cm, height=1.5cm]{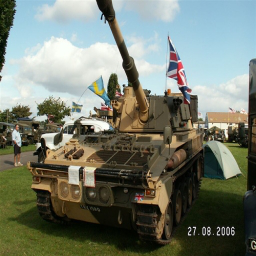} & \includegraphics[width=1.5cm, height=1.5cm]{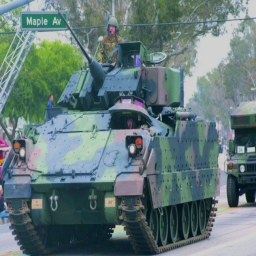} & \includegraphics[width=1.5cm, height=1.5cm]{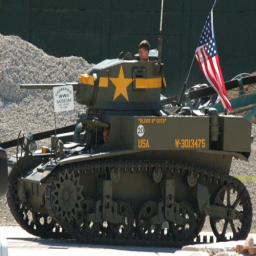} & \includegraphics[width=1.5cm, height=1.5cm]{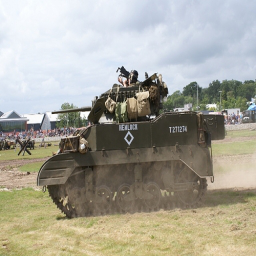} & \includegraphics[width=1.5cm, height=1.5cm]{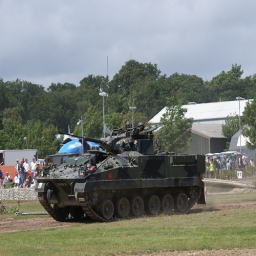} & \includegraphics[width=1.5cm, height=1.5cm]{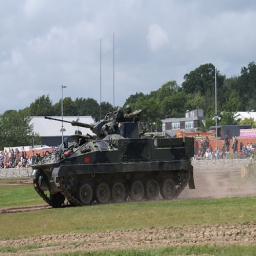} & \includegraphics[width=1.5cm, height=1.5cm]{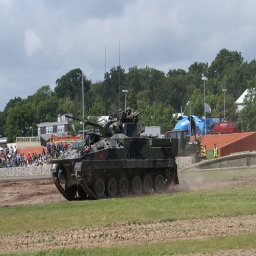} \\
 & \cmark & \xmark & \cmark & \cmark & \cmark & \cmark & \cmark & \cmark & \xmark & \cmark & \cmark & \xmark & \cmark & \cmark & \cmark & \cmark & \cmark & \cmark & \cmark & \cmark \\
\includegraphics[width=1.5cm, height=1.5cm]{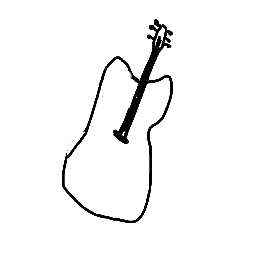} & \includegraphics[width=1.5cm, height=1.5cm]{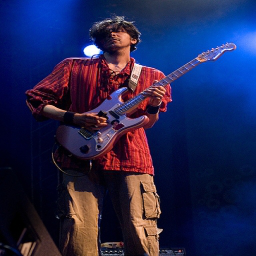} & \includegraphics[width=1.5cm, height=1.5cm]{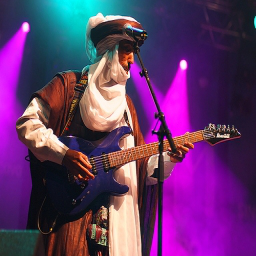} & \includegraphics[width=1.5cm, height=1.5cm]{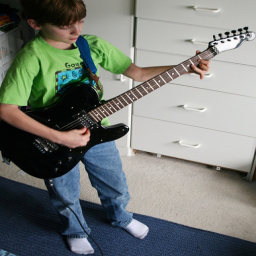} & \includegraphics[width=1.5cm, height=1.5cm]{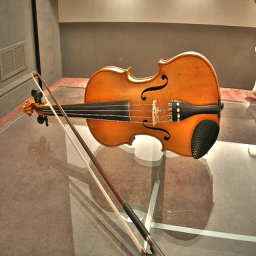} & \includegraphics[width=1.5cm, height=1.5cm]{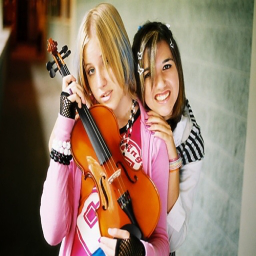} & \includegraphics[width=1.5cm, height=1.5cm]{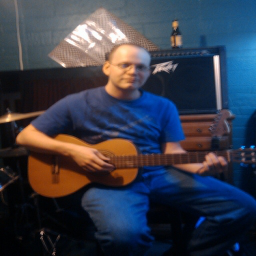} & \includegraphics[width=1.5cm, height=1.5cm]{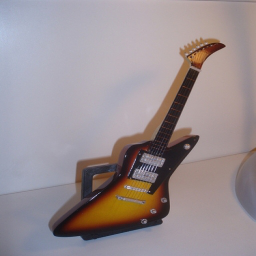} & \includegraphics[width=1.5cm, height=1.5cm]{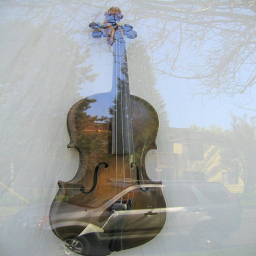} & \includegraphics[width=1.5cm, height=1.5cm]{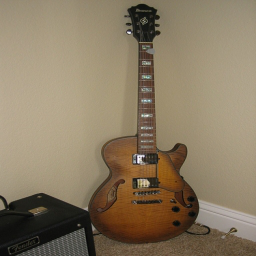} & \includegraphics[width=1.5cm, height=1.5cm]{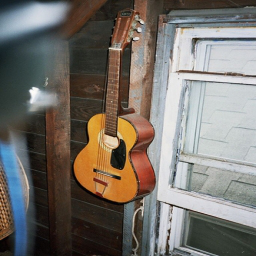} & \includegraphics[width=1.5cm, height=1.5cm]{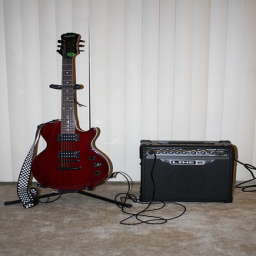} & \includegraphics[width=1.5cm, height=1.5cm]{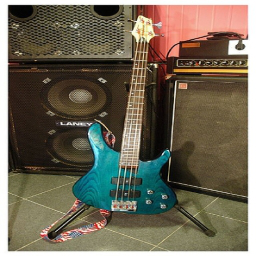} & \includegraphics[width=1.5cm, height=1.5cm]{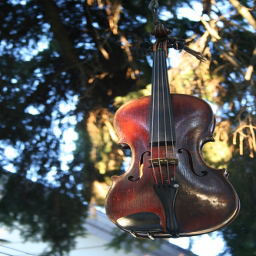} & \includegraphics[width=1.5cm, height=1.5cm]{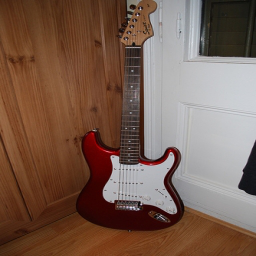} & \includegraphics[width=1.5cm, height=1.5cm]{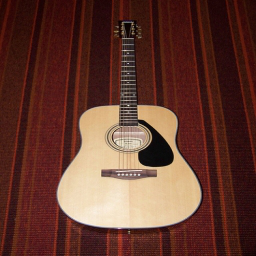} & \includegraphics[width=1.5cm, height=1.5cm]{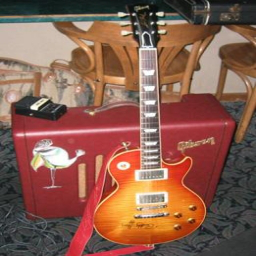} & \includegraphics[width=1.5cm, height=1.5cm]{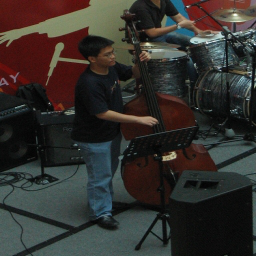} & \includegraphics[width=1.5cm, height=1.5cm]{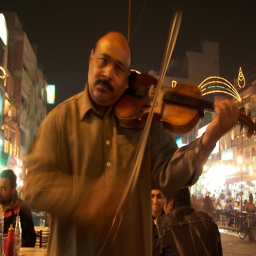} & \includegraphics[width=1.5cm, height=1.5cm]{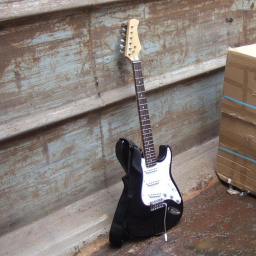} & \includegraphics[width=1.5cm, height=1.5cm]{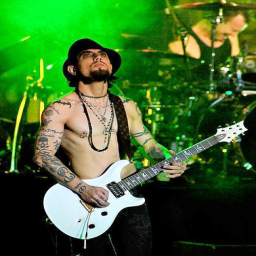} \\
 & \cmark & \cmark & \cmark & \xmark & \xmark & \cmark & \cmark & \xmark & \cmark & \cmark & \cmark & \cmark & \xmark & \cmark & \cmark & \cmark & \xmark & \xmark & \cmark & \cmark \\
\includegraphics[width=1.5cm, height=1.5cm]{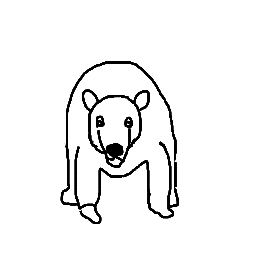} & \includegraphics[width=1.5cm, height=1.5cm]{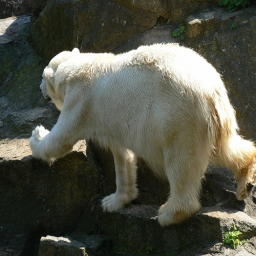} & \includegraphics[width=1.5cm, height=1.5cm]{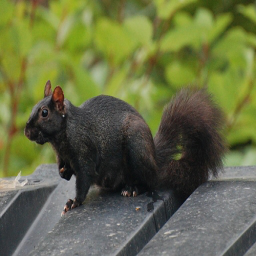} & \includegraphics[width=1.5cm, height=1.5cm]{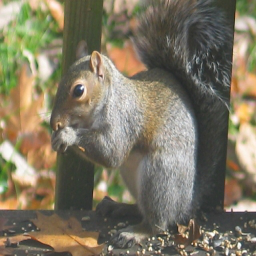} & \includegraphics[width=1.5cm, height=1.5cm]{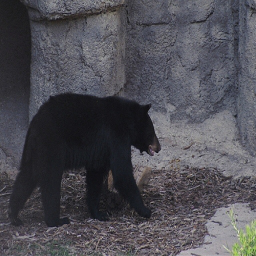} & \includegraphics[width=1.5cm, height=1.5cm]{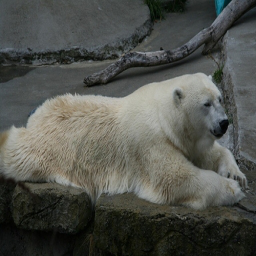} & \includegraphics[width=1.5cm, height=1.5cm]{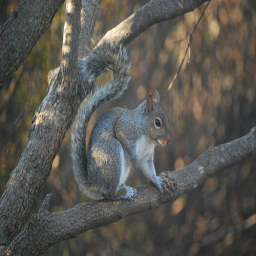} & \includegraphics[width=1.5cm, height=1.5cm]{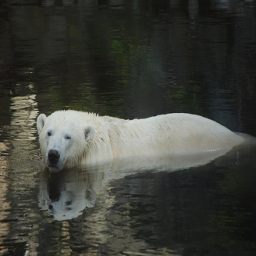} & \includegraphics[width=1.5cm, height=1.5cm]{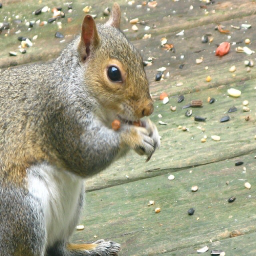} & \includegraphics[width=1.5cm, height=1.5cm]{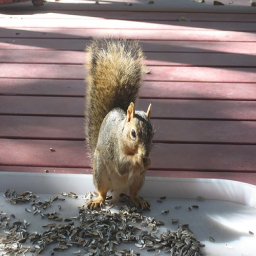} & \includegraphics[width=1.5cm, height=1.5cm]{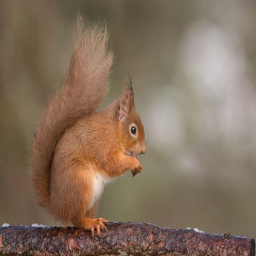} & \includegraphics[width=1.5cm, height=1.5cm]{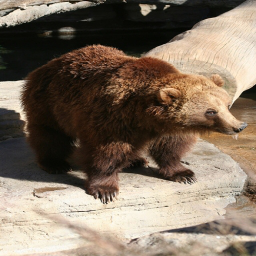} & \includegraphics[width=1.5cm, height=1.5cm]{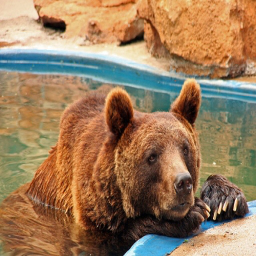} & \includegraphics[width=1.5cm, height=1.5cm]{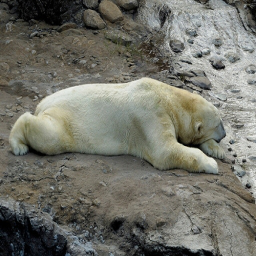} & \includegraphics[width=1.5cm, height=1.5cm]{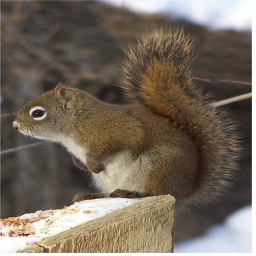} & \includegraphics[width=1.5cm, height=1.5cm]{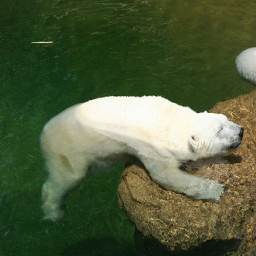} & \includegraphics[width=1.5cm, height=1.5cm]{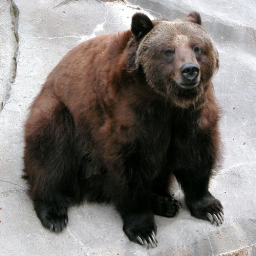} & \includegraphics[width=1.5cm, height=1.5cm]{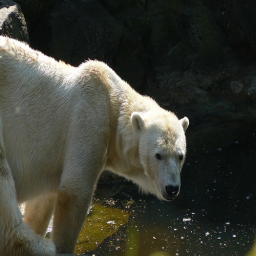} & \includegraphics[width=1.5cm, height=1.5cm]{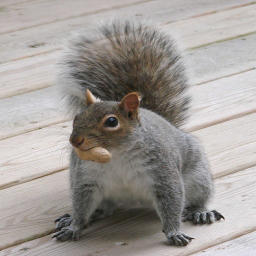} & \includegraphics[width=1.5cm, height=1.5cm]{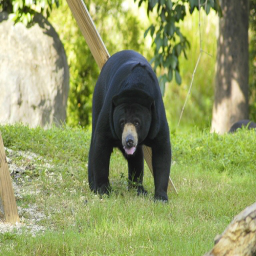} & \includegraphics[width=1.5cm, height=1.5cm]{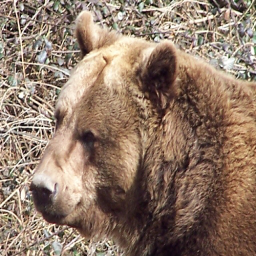} \\
 & \cmark & \xmark & \xmark & \cmark & \cmark & \xmark & \cmark & \xmark & \xmark & \xmark & \cmark & \cmark & \cmark & \xmark & \cmark & \cmark & \cmark & \xmark & \cmark & \cmark \\
\includegraphics[width=1.5cm, height=1.5cm]{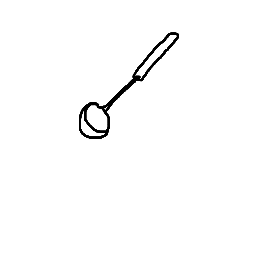} & \includegraphics[width=1.5cm, height=1.5cm]{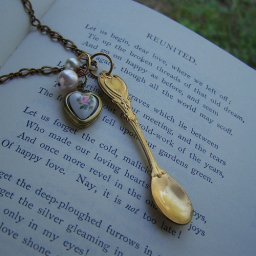} & \includegraphics[width=1.5cm, height=1.5cm]{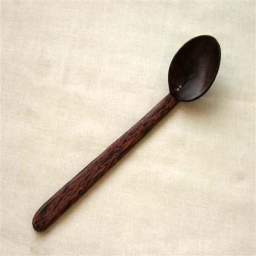} & \includegraphics[width=1.5cm, height=1.5cm]{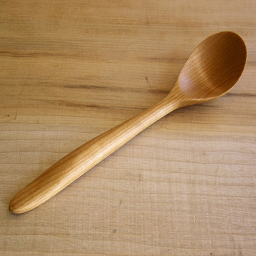} & \includegraphics[width=1.5cm, height=1.5cm]{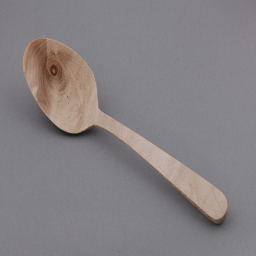} & \includegraphics[width=1.5cm, height=1.5cm]{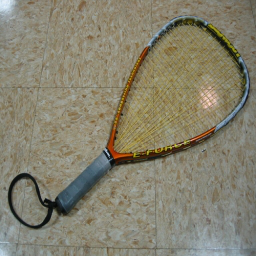} & \includegraphics[width=1.5cm, height=1.5cm]{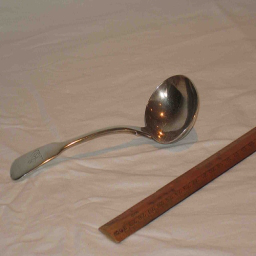} & \includegraphics[width=1.5cm, height=1.5cm]{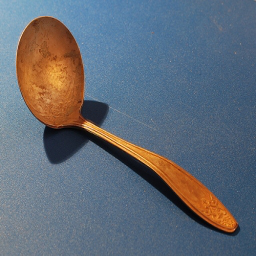} & \includegraphics[width=1.5cm, height=1.5cm]{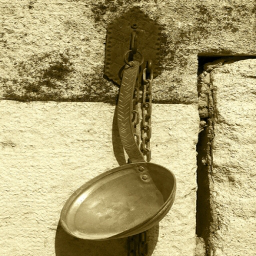} & \includegraphics[width=1.5cm, height=1.5cm]{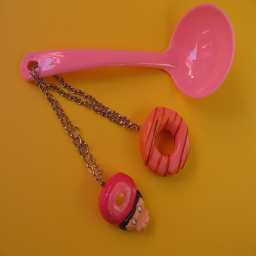} & \includegraphics[width=1.5cm, height=1.5cm]{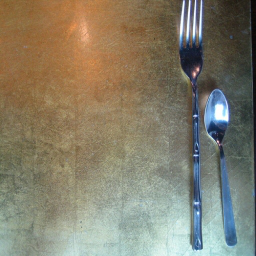} & \includegraphics[width=1.5cm, height=1.5cm]{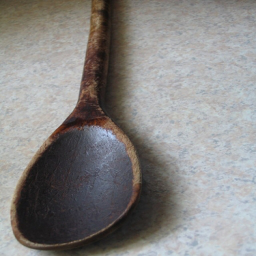} & \includegraphics[width=1.5cm, height=1.5cm]{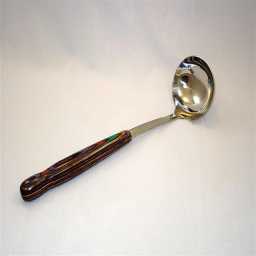} & \includegraphics[width=1.5cm, height=1.5cm]{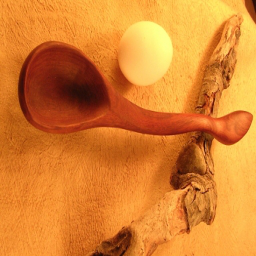} & \includegraphics[width=1.5cm, height=1.5cm]{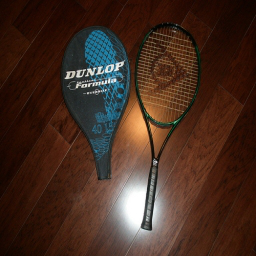} & \includegraphics[width=1.5cm, height=1.5cm]{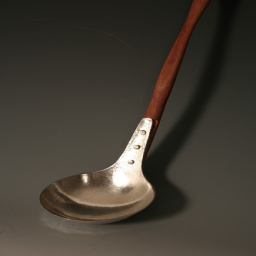} & \includegraphics[width=1.5cm, height=1.5cm]{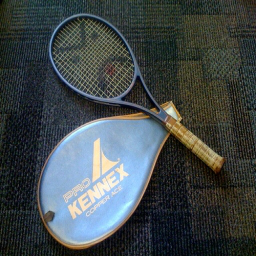} & \includegraphics[width=1.5cm, height=1.5cm]{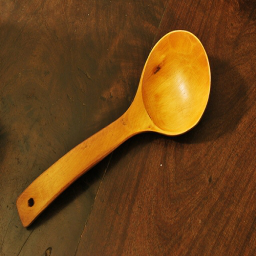} & \includegraphics[width=1.5cm, height=1.5cm]{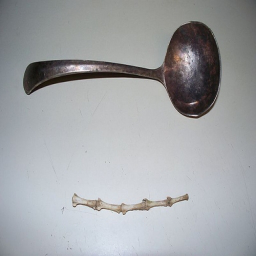} & \includegraphics[width=1.5cm, height=1.5cm]{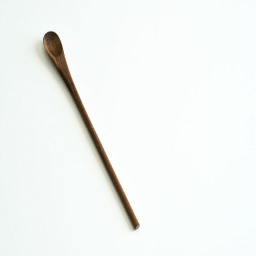} & \includegraphics[width=1.5cm, height=1.5cm]{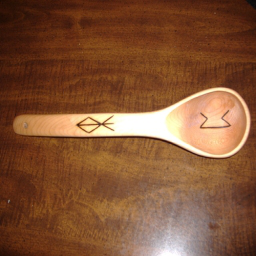} \\
 & \cmark & \cmark & \cmark & \cmark & \xmark & \cmark & \cmark & \cmark & \cmark & \cmark & \cmark & \cmark & \cmark & \xmark & \cmark & \cmark & \cmark & \cmark & \cmark & \cmark \\
\includegraphics[width=1.5cm, height=1.5cm]{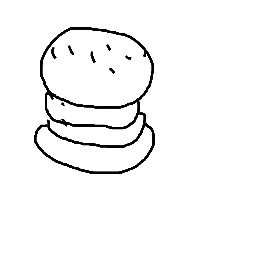} & \includegraphics[width=1.5cm, height=1.5cm]{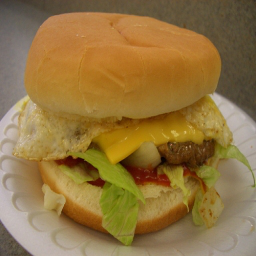} & \includegraphics[width=1.5cm, height=1.5cm]{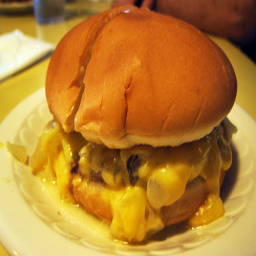} & \includegraphics[width=1.5cm, height=1.5cm]{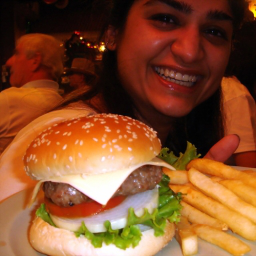} & \includegraphics[width=1.5cm, height=1.5cm]{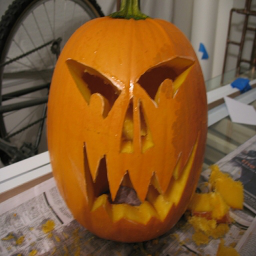} & \includegraphics[width=1.5cm, height=1.5cm]{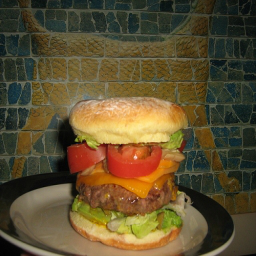} & \includegraphics[width=1.5cm, height=1.5cm]{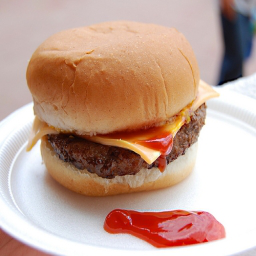} & \includegraphics[width=1.5cm, height=1.5cm]{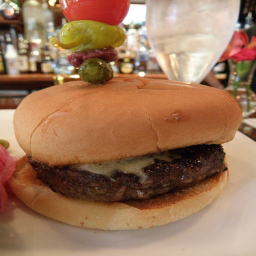} & \includegraphics[width=1.5cm, height=1.5cm]{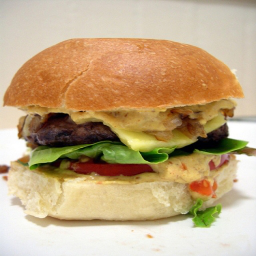} & \includegraphics[width=1.5cm, height=1.5cm]{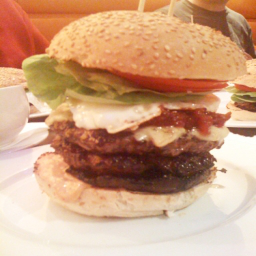} & \includegraphics[width=1.5cm, height=1.5cm]{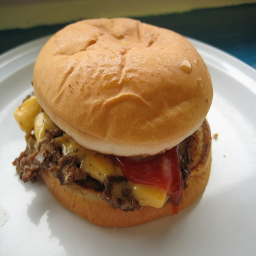} & \includegraphics[width=1.5cm, height=1.5cm]{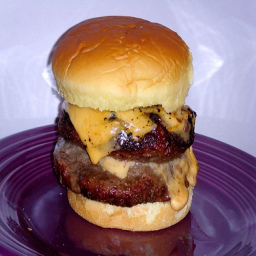} & \includegraphics[width=1.5cm, height=1.5cm]{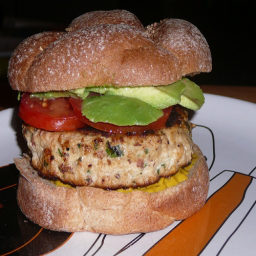} & \includegraphics[width=1.5cm, height=1.5cm]{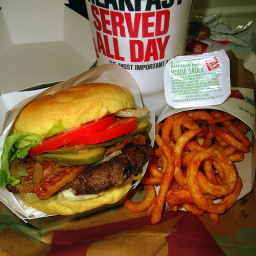} & \includegraphics[width=1.5cm, height=1.5cm]{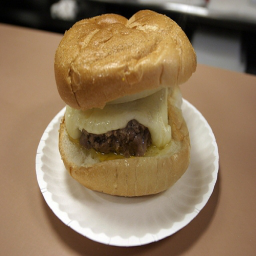} & \includegraphics[width=1.5cm, height=1.5cm]{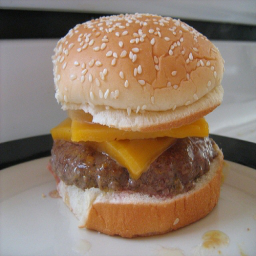} & \includegraphics[width=1.5cm, height=1.5cm]{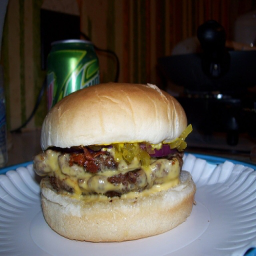} & \includegraphics[width=1.5cm, height=1.5cm]{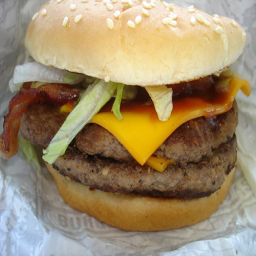} & \includegraphics[width=1.5cm, height=1.5cm]{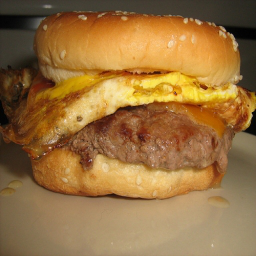} & \includegraphics[width=1.5cm, height=1.5cm]{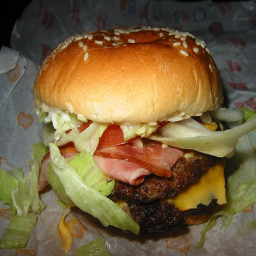} & \includegraphics[width=1.5cm, height=1.5cm]{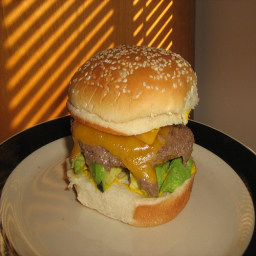} \\ 
 & \cmark & \cmark & \cmark & \xmark & \cmark & \cmark & \cmark & \cmark & \cmark & \cmark & \cmark & \cmark & \cmark & \cmark & \cmark & \cmark & \cmark & \cmark & \cmark & \cmark \\
\includegraphics[width=1.5cm, height=1.5cm]{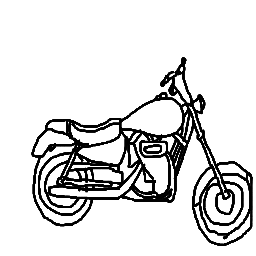} & \includegraphics[width=1.5cm, height=1.5cm]{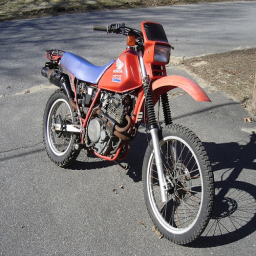} & \includegraphics[width=1.5cm, height=1.5cm]{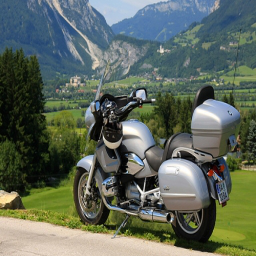} & \includegraphics[width=1.5cm, height=1.5cm]{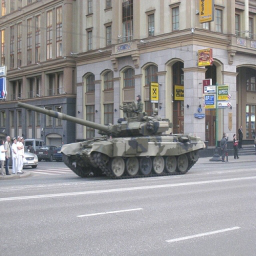} & \includegraphics[width=1.5cm, height=1.5cm]{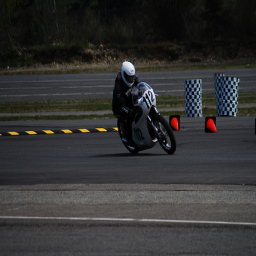} & \includegraphics[width=1.5cm, height=1.5cm]{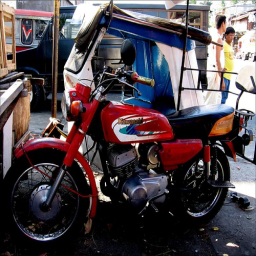} & \includegraphics[width=1.5cm, height=1.5cm]{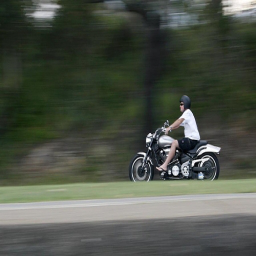} & \includegraphics[width=1.5cm, height=1.5cm]{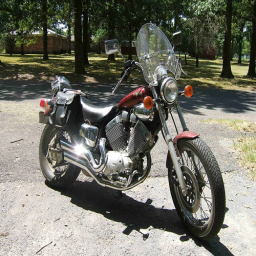} & \includegraphics[width=1.5cm, height=1.5cm]{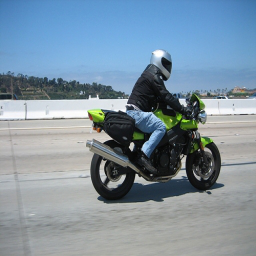} & \includegraphics[width=1.5cm, height=1.5cm]{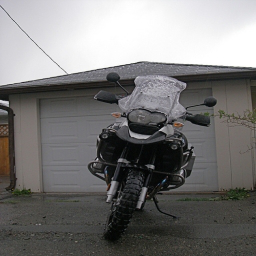} & \includegraphics[width=1.5cm, height=1.5cm]{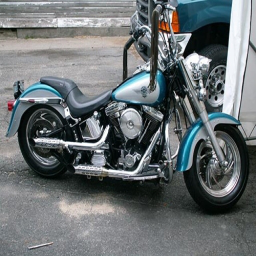} & \includegraphics[width=1.5cm, height=1.5cm]{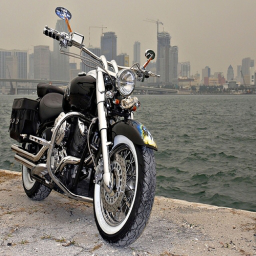} & \includegraphics[width=1.5cm, height=1.5cm]{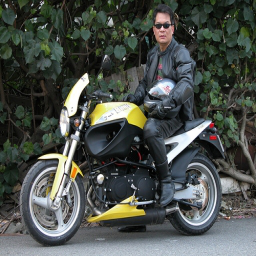} & \includegraphics[width=1.5cm, height=1.5cm]{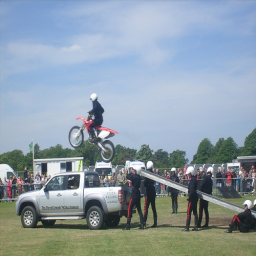} & \includegraphics[width=1.5cm, height=1.5cm]{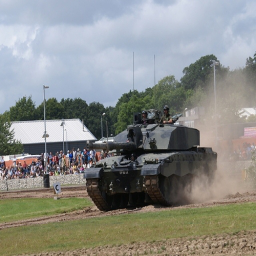} & \includegraphics[width=1.5cm, height=1.5cm]{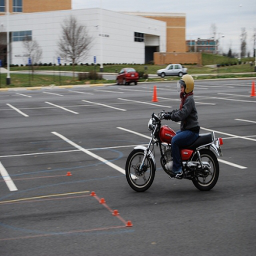} & \includegraphics[width=1.5cm, height=1.5cm]{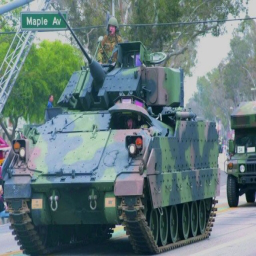} & \includegraphics[width=1.5cm, height=1.5cm]{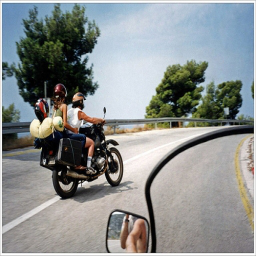} & \includegraphics[width=1.5cm, height=1.5cm]{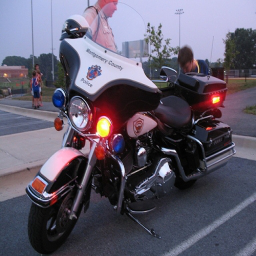} & \includegraphics[width=1.5cm, height=1.5cm]{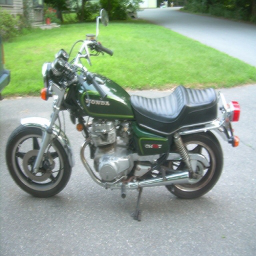} & \includegraphics[width=1.5cm, height=1.5cm]{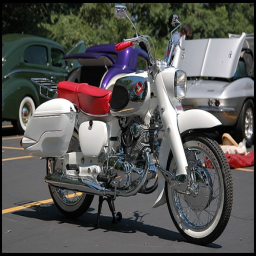} \\ 
 & \cmark & \cmark & \xmark & \cmark & \cmark & \cmark & \cmark & \cmark & \cmark & \cmark & \cmark & \cmark & \cmark & \xmark & \cmark & \xmark & \cmark & \cmark & \cmark & \cmark \\
\end{tabular}
}
\end{center}
\caption{Top-20 zero-shot SBIR results obtained by our SEM-PCYC model on the Sketchy (Extended) dataset are shown here according to the Euclidean distances, where the green ticks denote the correctly retrieved candidates, whereas the red crosses indicate the wrong retrievals. (best viewed in color)}
\label{fig:qual_results_sketchy}
\end{figure*}

\begin{figure*}[!t]
\begin{center}
\resizebox{\textwidth}{!}{
\begin{tabular}{@{}c@{}c@{}c@{}c@{}c@{}c@{}c@{}c@{}c@{}c@{}c@{}c@{}c@{}c@{}c@{}c@{}c@{}c@{}c@{}c@{}c}
\includegraphics[width=1.5cm, height=1.5cm]{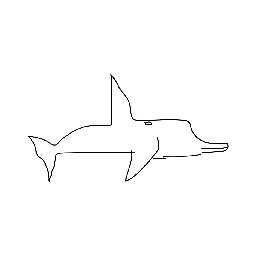} & \includegraphics[width=1.5cm, height=1.5cm]{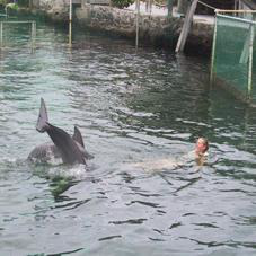} & \includegraphics[width=1.5cm, height=1.5cm]{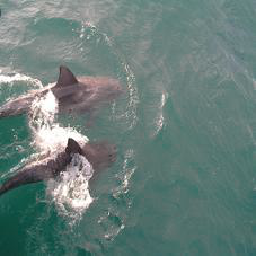} & \includegraphics[width=1.5cm, height=1.5cm]{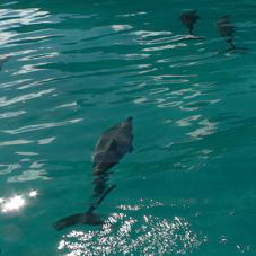} & \includegraphics[width=1.5cm, height=1.5cm]{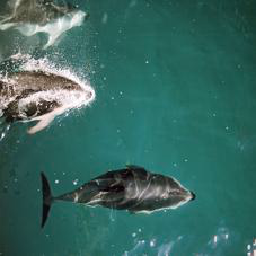} & \includegraphics[width=1.5cm, height=1.5cm]{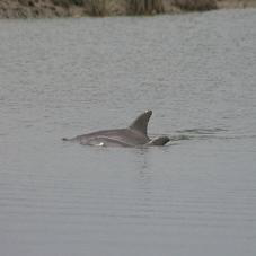} & \includegraphics[width=1.5cm, height=1.5cm]{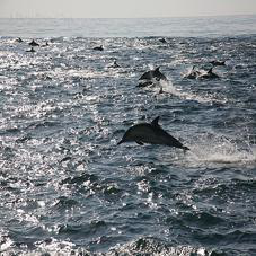} & \includegraphics[width=1.5cm, height=1.5cm]{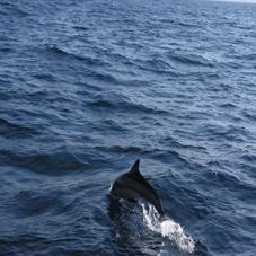} & \includegraphics[width=1.5cm, height=1.5cm]{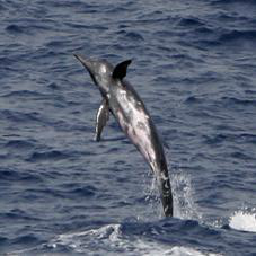} & \includegraphics[width=1.5cm, height=1.5cm]{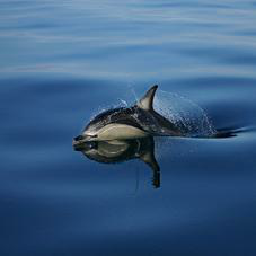} & \includegraphics[width=1.5cm, height=1.5cm]{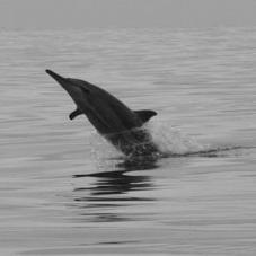} & \includegraphics[width=1.5cm, height=1.5cm]{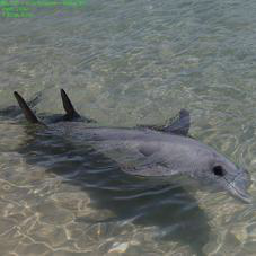} & \includegraphics[width=1.5cm, height=1.5cm]{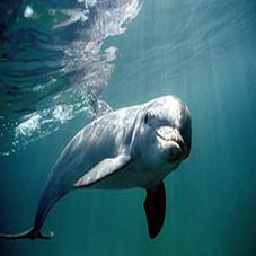} & \includegraphics[width=1.5cm, height=1.5cm]{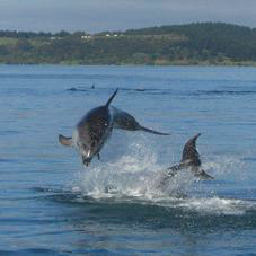} & \includegraphics[width=1.5cm, height=1.5cm]{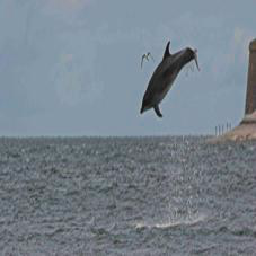} & \includegraphics[width=1.5cm, height=1.5cm]{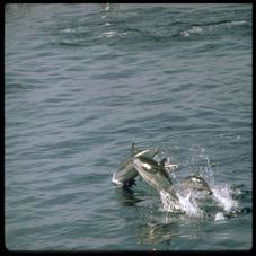} & \includegraphics[width=1.5cm, height=1.5cm]{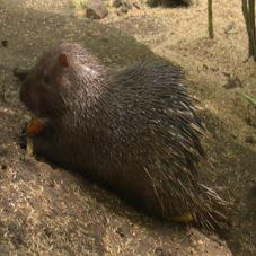} & \includegraphics[width=1.5cm, height=1.5cm]{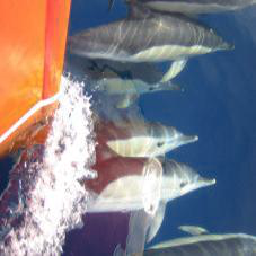} & \includegraphics[width=1.5cm, height=1.5cm]{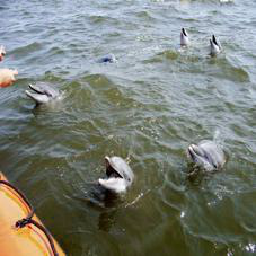} & \includegraphics[width=1.5cm, height=1.5cm]{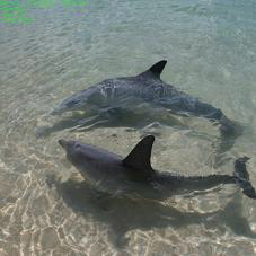} & \includegraphics[width=1.5cm, height=1.5cm]{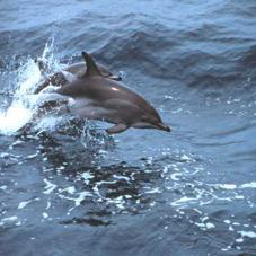} \\ 
 & \cmark & \cmark & \cmark & \cmark & \cmark & \cmark & \cmark & \cmark & \cmark & \cmark & \cmark & \cmark & \cmark & \cmark & \cmark & \xmark & \cmark & \cmark & \cmark & \cmark \\
\includegraphics[width=1.5cm, height=1.5cm]{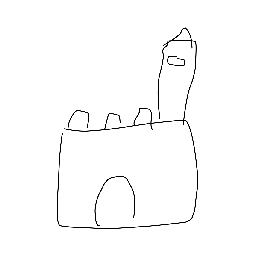} & \includegraphics[width=1.5cm, height=1.5cm]{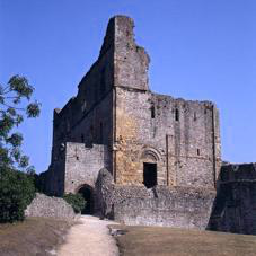} & \includegraphics[width=1.5cm, height=1.5cm]{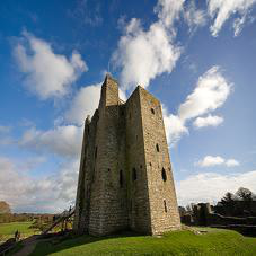} & \includegraphics[width=1.5cm, height=1.5cm]{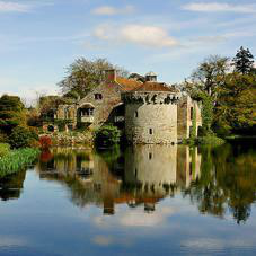} & \includegraphics[width=1.5cm, height=1.5cm]{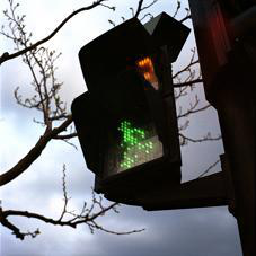} & \includegraphics[width=1.5cm, height=1.5cm]{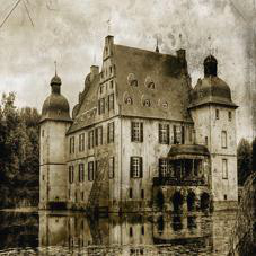} & \includegraphics[width=1.5cm, height=1.5cm]{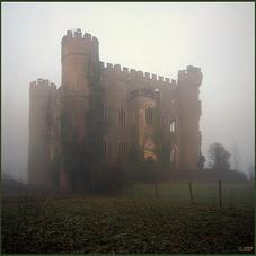} & \includegraphics[width=1.5cm, height=1.5cm]{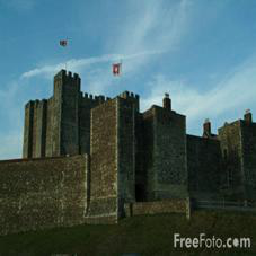} & \includegraphics[width=1.5cm, height=1.5cm]{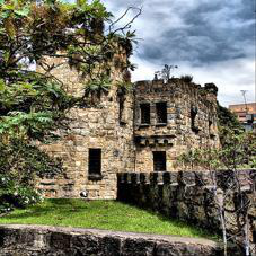} & \includegraphics[width=1.5cm, height=1.5cm]{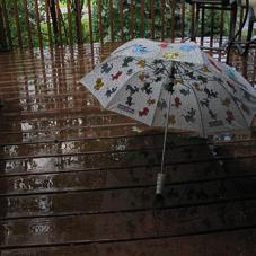} & \includegraphics[width=1.5cm, height=1.5cm]{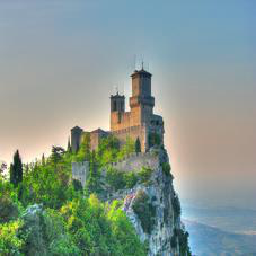} & \includegraphics[width=1.5cm, height=1.5cm]{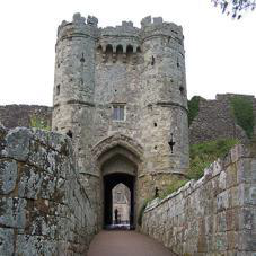} & \includegraphics[width=1.5cm, height=1.5cm]{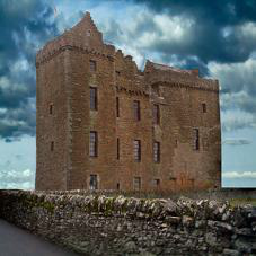} & \includegraphics[width=1.5cm, height=1.5cm]{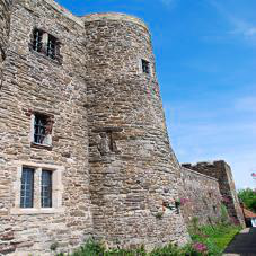} & \includegraphics[width=1.5cm, height=1.5cm]{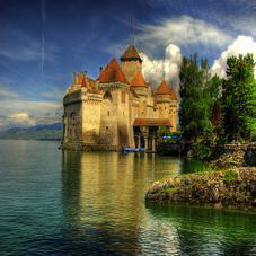} & \includegraphics[width=1.5cm, height=1.5cm]{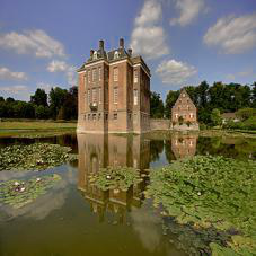} & \includegraphics[width=1.5cm, height=1.5cm]{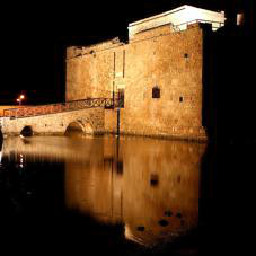} & \includegraphics[width=1.5cm, height=1.5cm]{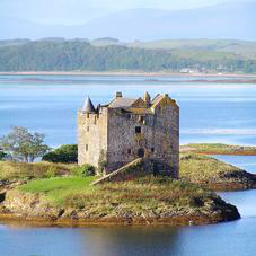} & \includegraphics[width=1.5cm, height=1.5cm]{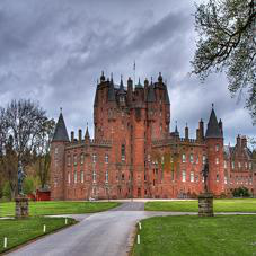} & \includegraphics[width=1.5cm, height=1.5cm]{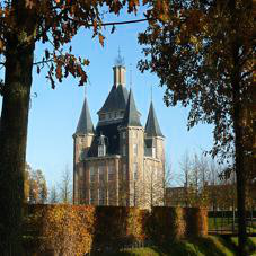} & \includegraphics[width=1.5cm, height=1.5cm]{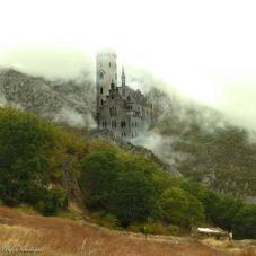} \\ 
 & \cmark & \cmark & \cmark & \xmark & \cmark & \cmark & \cmark & \cmark & \xmark & \cmark & \cmark & \cmark & \cmark & \cmark & \cmark & \cmark & \cmark & \cmark & \cmark & \cmark \\
\includegraphics[width=1.5cm, height=1.5cm]{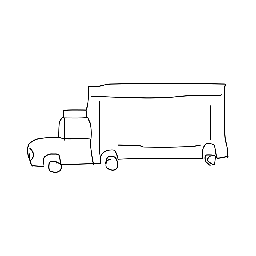} & \includegraphics[width=1.5cm, height=1.5cm]{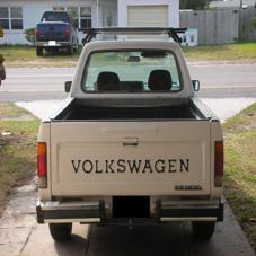} & \includegraphics[width=1.5cm, height=1.5cm]{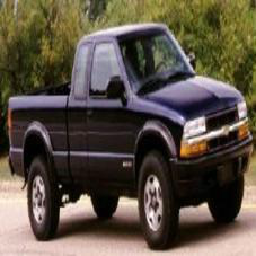} & \includegraphics[width=1.5cm, height=1.5cm]{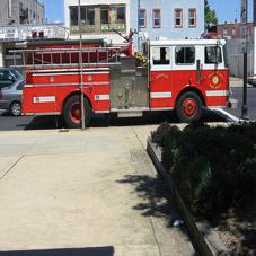} & \includegraphics[width=1.5cm, height=1.5cm]{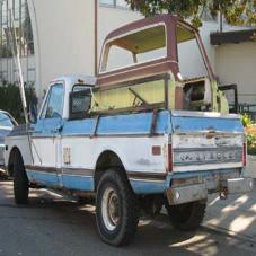} & \includegraphics[width=1.5cm, height=1.5cm]{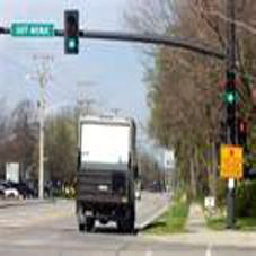} & \includegraphics[width=1.5cm, height=1.5cm]{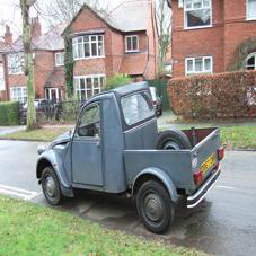} & \includegraphics[width=1.5cm, height=1.5cm]{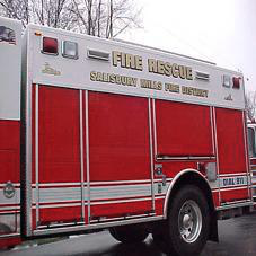} & \includegraphics[width=1.5cm, height=1.5cm]{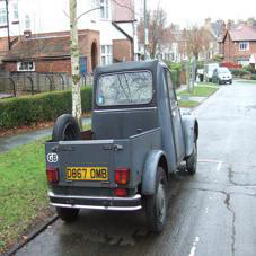} & \includegraphics[width=1.5cm, height=1.5cm]{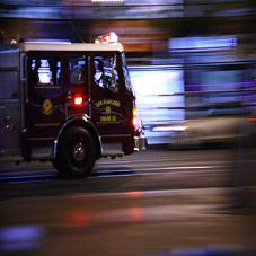} & \includegraphics[width=1.5cm, height=1.5cm]{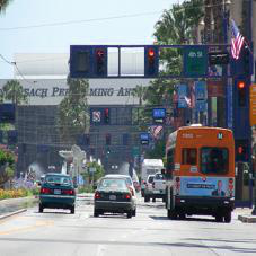} & \includegraphics[width=1.5cm, height=1.5cm]{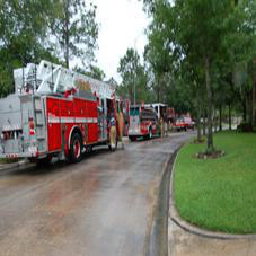} & \includegraphics[width=1.5cm, height=1.5cm]{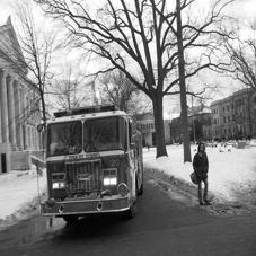} & \includegraphics[width=1.5cm, height=1.5cm]{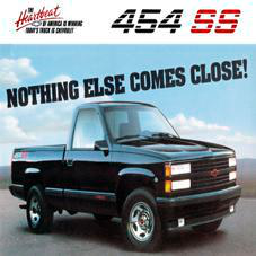} & \includegraphics[width=1.5cm, height=1.5cm]{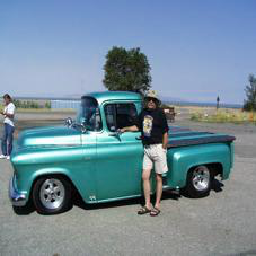} & \includegraphics[width=1.5cm, height=1.5cm]{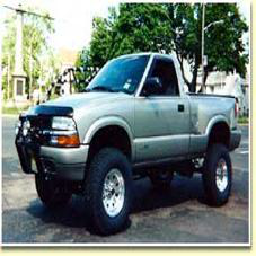} & \includegraphics[width=1.5cm, height=1.5cm]{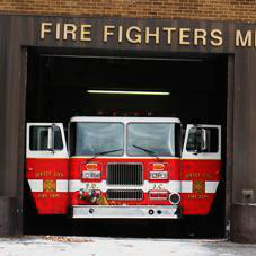} & \includegraphics[width=1.5cm, height=1.5cm]{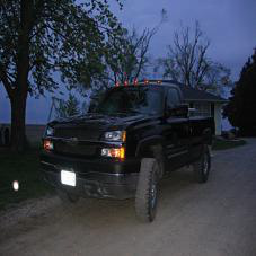} & \includegraphics[width=1.5cm, height=1.5cm]{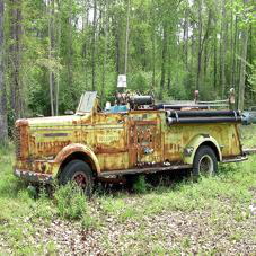} & \includegraphics[width=1.5cm, height=1.5cm]{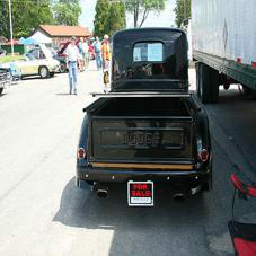} & \includegraphics[width=1.5cm, height=1.5cm]{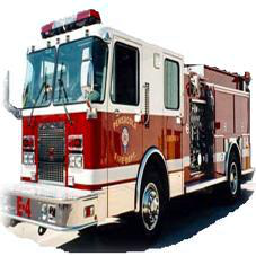} \\ 
 & \cmark & \cmark & \cmark & \cmark & \xmark & \cmark & \cmark & \cmark & \cmark & \xmark & \cmark & \cmark & \cmark & \cmark & \cmark & \cmark & \cmark & \cmark & \cmark & \cmark \\
\includegraphics[width=1.5cm, height=1.5cm]{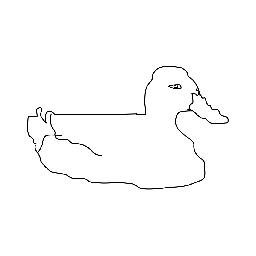} & \includegraphics[width=1.5cm, height=1.5cm]{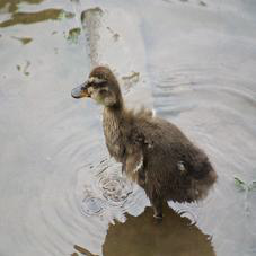} & \includegraphics[width=1.5cm, height=1.5cm]{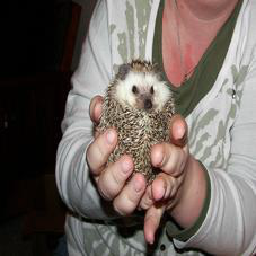} & \includegraphics[width=1.5cm, height=1.5cm]{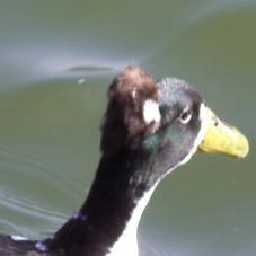} & \includegraphics[width=1.5cm, height=1.5cm]{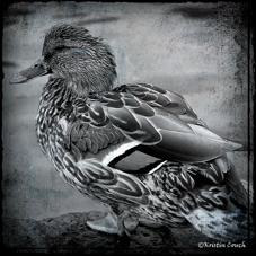} & \includegraphics[width=1.5cm, height=1.5cm]{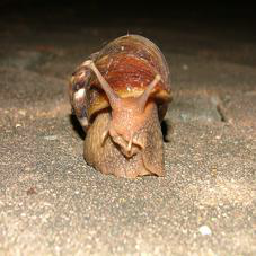} & \includegraphics[width=1.5cm, height=1.5cm]{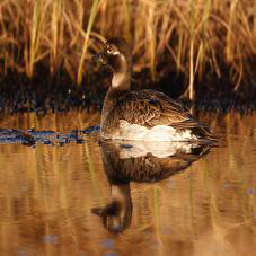} & \includegraphics[width=1.5cm, height=1.5cm]{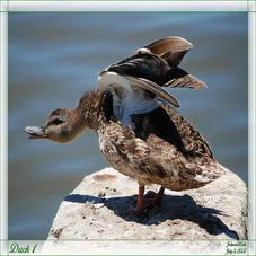} & \includegraphics[width=1.5cm, height=1.5cm]{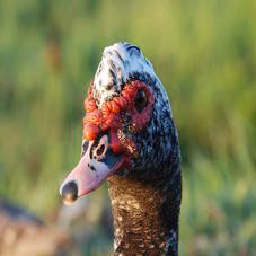} & \includegraphics[width=1.5cm, height=1.5cm]{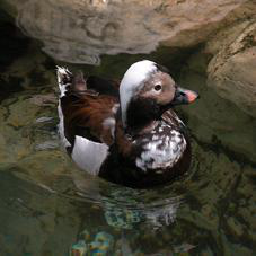} & \includegraphics[width=1.5cm, height=1.5cm]{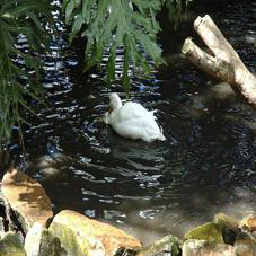} & \includegraphics[width=1.5cm, height=1.5cm]{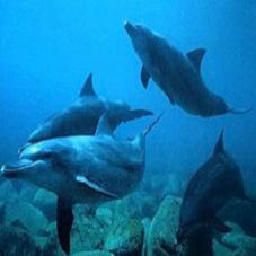} & \includegraphics[width=1.5cm, height=1.5cm]{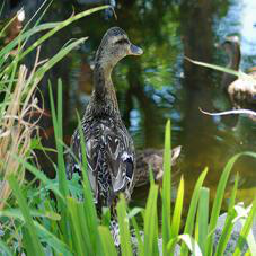} & \includegraphics[width=1.5cm, height=1.5cm]{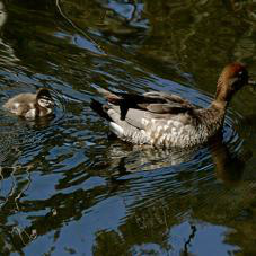} & \includegraphics[width=1.5cm, height=1.5cm]{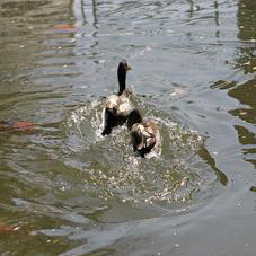} & \includegraphics[width=1.5cm, height=1.5cm]{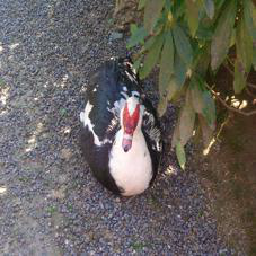} & \includegraphics[width=1.5cm, height=1.5cm]{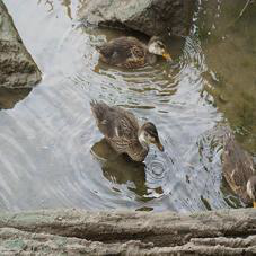} & \includegraphics[width=1.5cm, height=1.5cm]{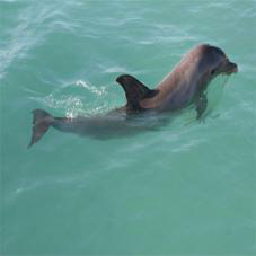} & \includegraphics[width=1.5cm, height=1.5cm]{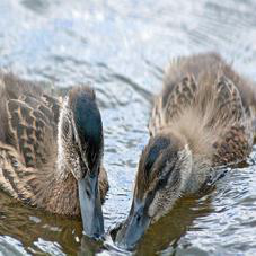} & \includegraphics[width=1.5cm, height=1.5cm]{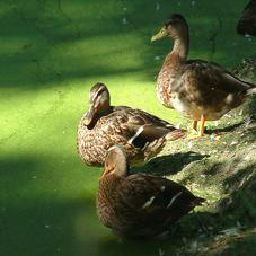} & \includegraphics[width=1.5cm, height=1.5cm]{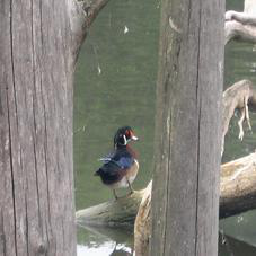} \\ 
 & \cmark & \xmark & \cmark & \cmark & \xmark & \cmark & \cmark & \cmark & \cmark & \cmark & \xmark & \cmark & \cmark & \cmark & \cmark & \cmark & \xmark & \cmark & \cmark & \cmark \\ 
\includegraphics[width=1.5cm, height=1.5cm]{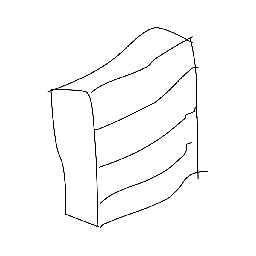} & \includegraphics[width=1.5cm, height=1.5cm]{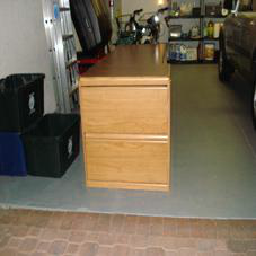} & \includegraphics[width=1.5cm, height=1.5cm]{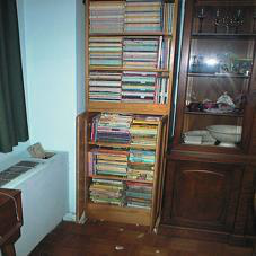} & \includegraphics[width=1.5cm, height=1.5cm]{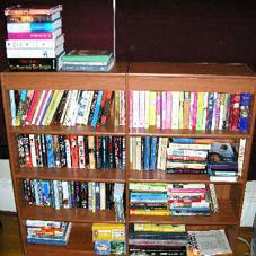} & \includegraphics[width=1.5cm, height=1.5cm]{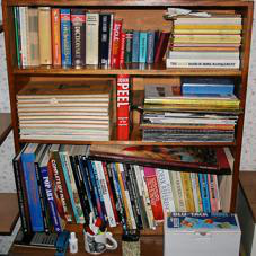} & \includegraphics[width=1.5cm, height=1.5cm]{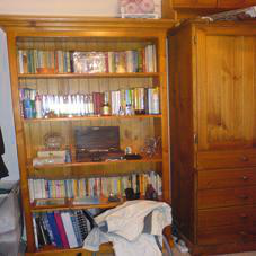} & \includegraphics[width=1.5cm, height=1.5cm]{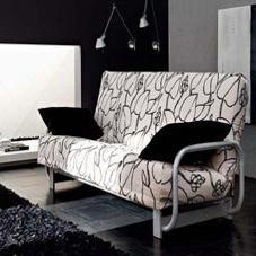} & \includegraphics[width=1.5cm, height=1.5cm]{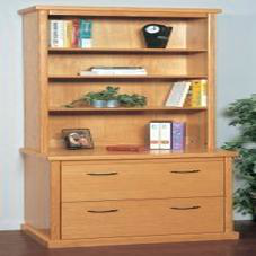} & \includegraphics[width=1.5cm, height=1.5cm]{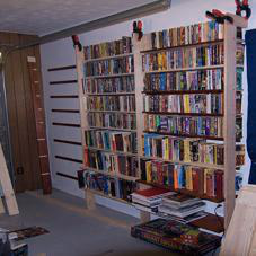} & \includegraphics[width=1.5cm, height=1.5cm]{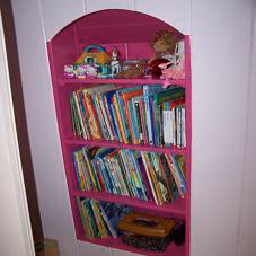} & \includegraphics[width=1.5cm, height=1.5cm]{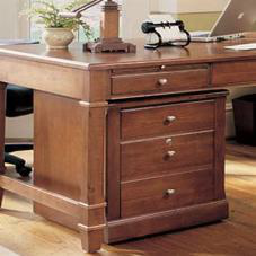} & \includegraphics[width=1.5cm, height=1.5cm]{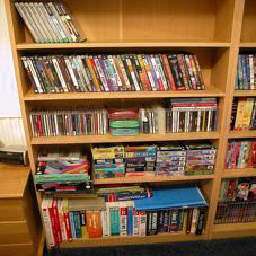} & \includegraphics[width=1.5cm, height=1.5cm]{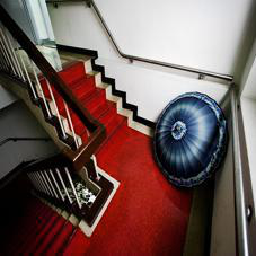} & \includegraphics[width=1.5cm, height=1.5cm]{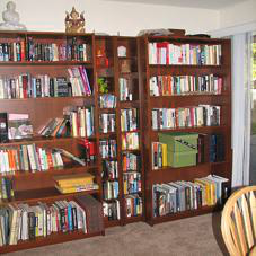} & \includegraphics[width=1.5cm, height=1.5cm]{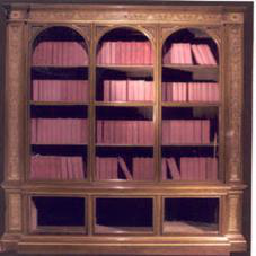} & \includegraphics[width=1.5cm, height=1.5cm]{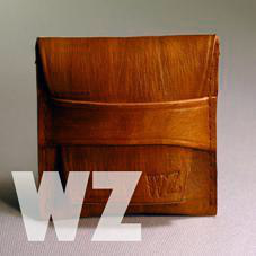} & \includegraphics[width=1.5cm, height=1.5cm]{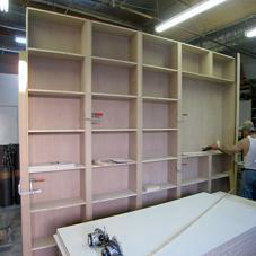} & \includegraphics[width=1.5cm, height=1.5cm]{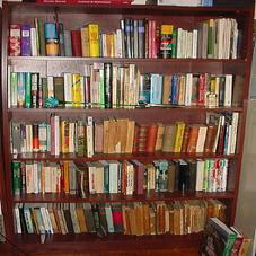} & \includegraphics[width=1.5cm, height=1.5cm]{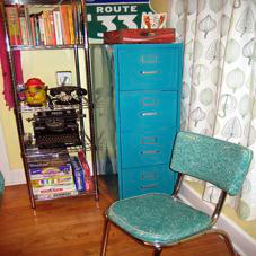} & \includegraphics[width=1.5cm, height=1.5cm]{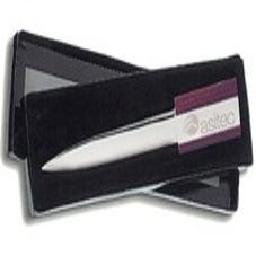} & \includegraphics[width=1.5cm, height=1.5cm]{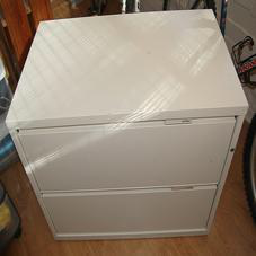} \\ 
 & \xmark & \cmark & \cmark & \cmark & \cmark & \xmark & \xmark & \cmark & \cmark & \xmark & \cmark & \xmark & \cmark & \cmark & \xmark & \cmark & \cmark & \xmark & \xmark & \xmark \\
\includegraphics[width=1.5cm, height=1.5cm]{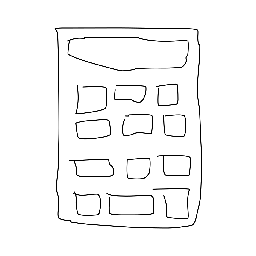} & \includegraphics[width=1.5cm, height=1.5cm]{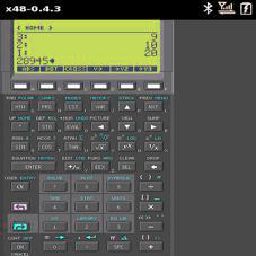} & \includegraphics[width=1.5cm, height=1.5cm]{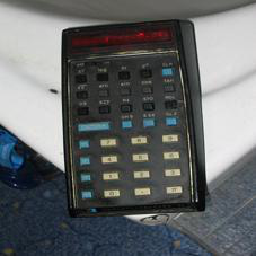} & \includegraphics[width=1.5cm, height=1.5cm]{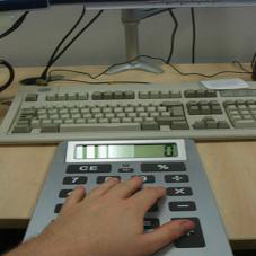} & \includegraphics[width=1.5cm, height=1.5cm]{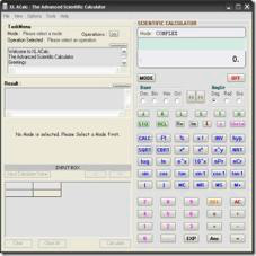} & \includegraphics[width=1.5cm, height=1.5cm]{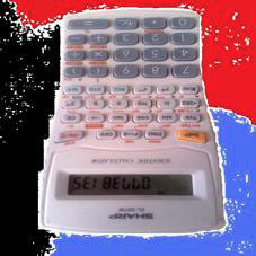} & \includegraphics[width=1.5cm, height=1.5cm]{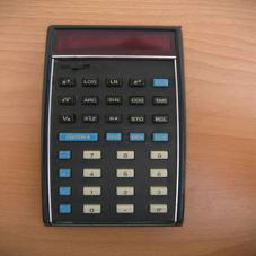} & \includegraphics[width=1.5cm, height=1.5cm]{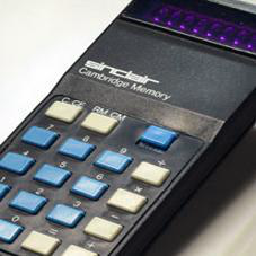} & \includegraphics[width=1.5cm, height=1.5cm]{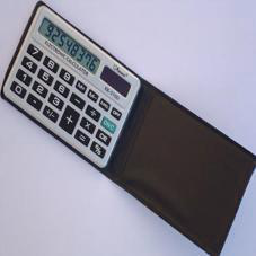} & \includegraphics[width=1.5cm, height=1.5cm]{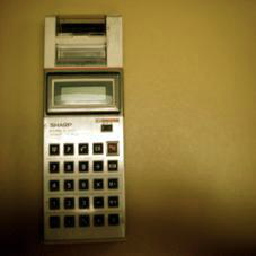} & \includegraphics[width=1.5cm, height=1.5cm]{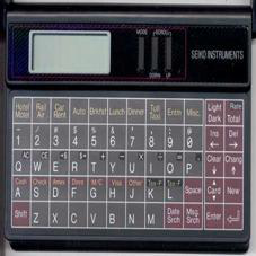} & \includegraphics[width=1.5cm, height=1.5cm]{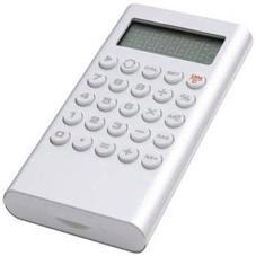} & \includegraphics[width=1.5cm, height=1.5cm]{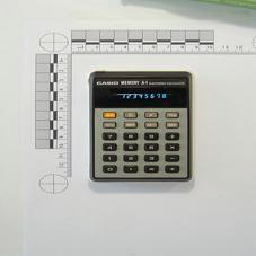} & \includegraphics[width=1.5cm, height=1.5cm]{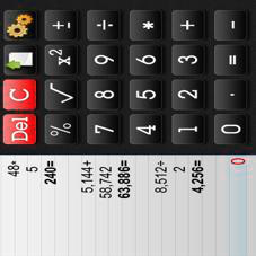} & \includegraphics[width=1.5cm, height=1.5cm]{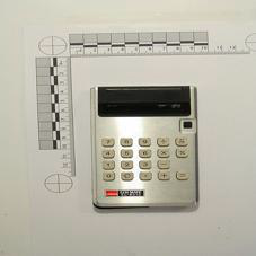} & \includegraphics[width=1.5cm, height=1.5cm]{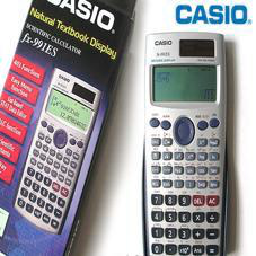} & \includegraphics[width=1.5cm, height=1.5cm]{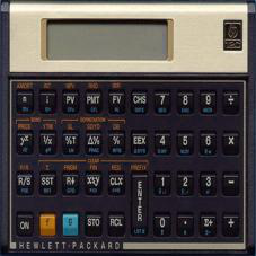} & \includegraphics[width=1.5cm, height=1.5cm]{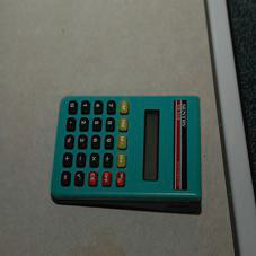} & \includegraphics[width=1.5cm, height=1.5cm]{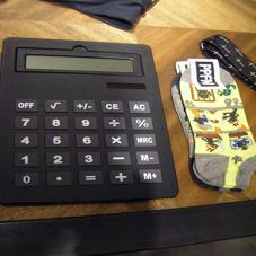} & \includegraphics[width=1.5cm, height=1.5cm]{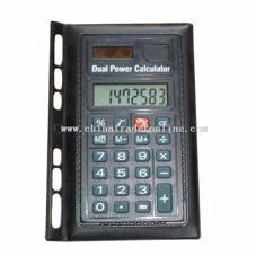} & \includegraphics[width=1.5cm, height=1.5cm]{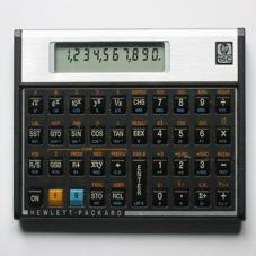} \\ 
 & \cmark & \cmark & \cmark & \cmark & \cmark & \cmark & \cmark & \cmark & \cmark & \cmark & \cmark & \cmark & \cmark & \cmark & \cmark & \cmark & \cmark & \cmark & \cmark & \cmark \\
\end{tabular}
}\end{center}
\caption{Top-20 zero-shot SBIR results obtained by our SEM-PCYC model on the TU-Berlin (Extended) dataset are shown here according to the Euclidean distances, where the green ticks denote the correctly retrieved candidates, whereas the red crosses indicate the wrong retrievals. (best viewed in color)}
\label{fig:qual_results_tu-berlin}
\end{figure*}

\begin{figure*}[!t]
\begin{center}
\resizebox{\textwidth}{!}{
\begin{tabular}{@{}c@{}c@{}c@{}c@{}c@{}c@{}c@{}c@{}c@{}c@{}c@{}c@{}c@{}c@{}c@{}c@{}c@{}c@{}c@{}c@{}c}
\includegraphics[width=1.5cm, height=1.5cm]{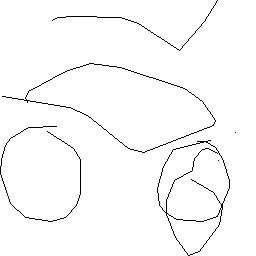} & \includegraphics[width=1.5cm, height=1.5cm]{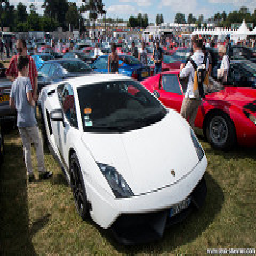} & \includegraphics[width=1.5cm, height=1.5cm]{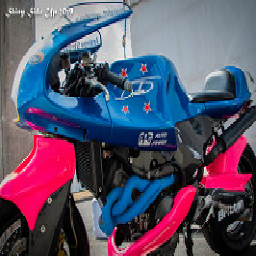} & \includegraphics[width=1.5cm, height=1.5cm]{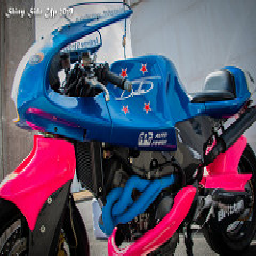} & \includegraphics[width=1.5cm, height=1.5cm]{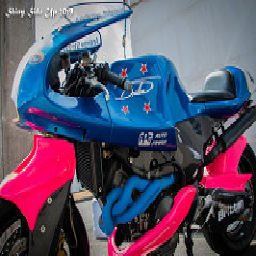} & \includegraphics[width=1.5cm, height=1.5cm]{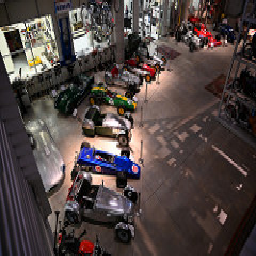} & \includegraphics[width=1.5cm, height=1.5cm]{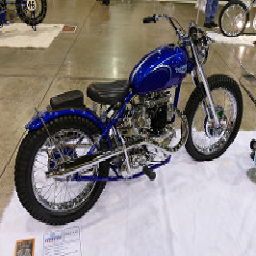} & \includegraphics[width=1.5cm, height=1.5cm]{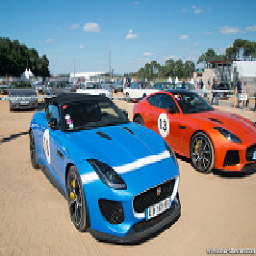} & \includegraphics[width=1.5cm, height=1.5cm]{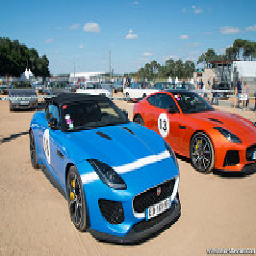} & \includegraphics[width=1.5cm, height=1.5cm]{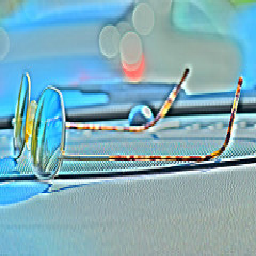} & \includegraphics[width=1.5cm, height=1.5cm]{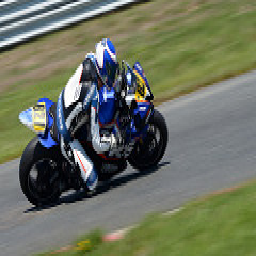} & \includegraphics[width=1.5cm, height=1.5cm]{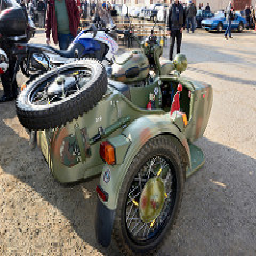} & \includegraphics[width=1.5cm, height=1.5cm]{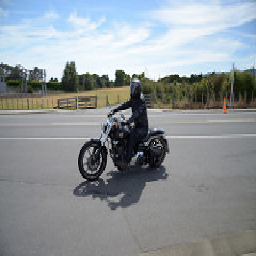} & \includegraphics[width=1.5cm, height=1.5cm]{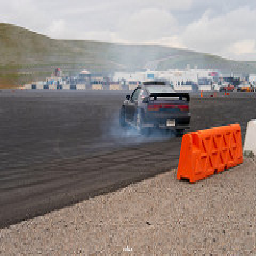} & \includegraphics[width=1.5cm, height=1.5cm]{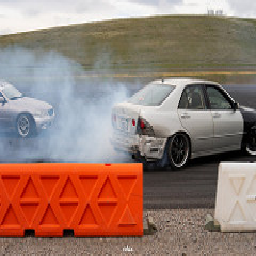} & \includegraphics[width=1.5cm, height=1.5cm]{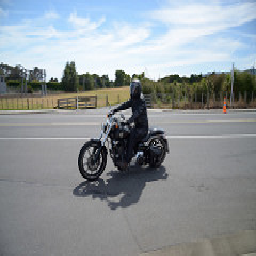} & \includegraphics[width=1.5cm, height=1.5cm]{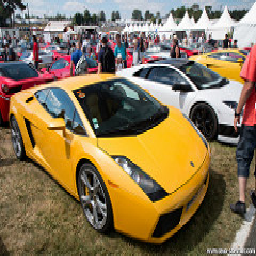} & \includegraphics[width=1.5cm, height=1.5cm]{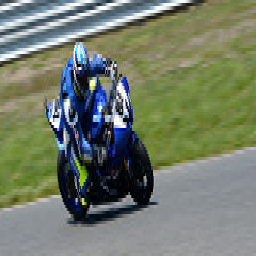} & \includegraphics[width=1.5cm, height=1.5cm]{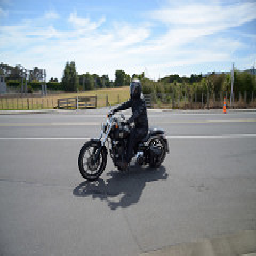} & \includegraphics[width=1.5cm, height=1.5cm]{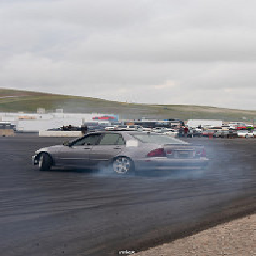} & \includegraphics[width=1.5cm, height=1.5cm]{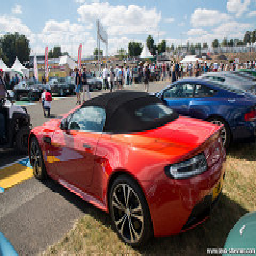} \\ 
 & \xmark & \cmark & \cmark & \cmark & \cmark & \cmark & \xmark & \xmark & \xmark & \cmark & \cmark & \cmark & \xmark & \xmark & \cmark & \xmark & \cmark & \cmark & \xmark & \xmark \\ 
\includegraphics[width=1.5cm, height=1.5cm]{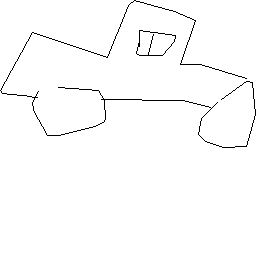} & \includegraphics[width=1.5cm, height=1.5cm]{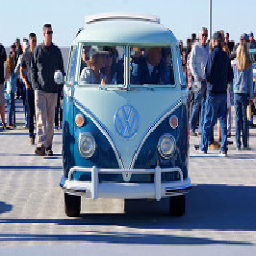} & \includegraphics[width=1.5cm, height=1.5cm]{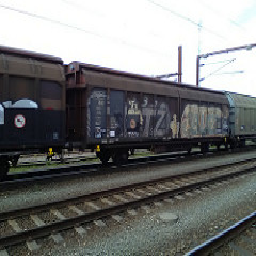} & \includegraphics[width=1.5cm, height=1.5cm]{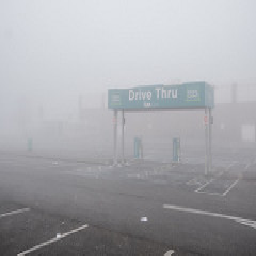} & \includegraphics[width=1.5cm, height=1.5cm]{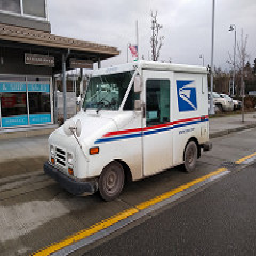} & \includegraphics[width=1.5cm, height=1.5cm]{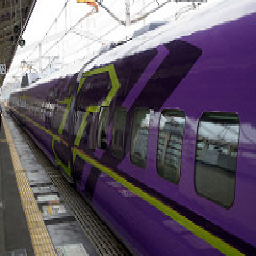} & \includegraphics[width=1.5cm, height=1.5cm]{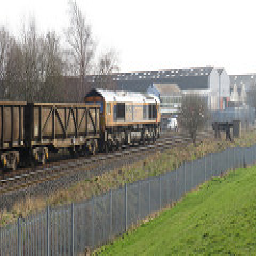} & \includegraphics[width=1.5cm, height=1.5cm]{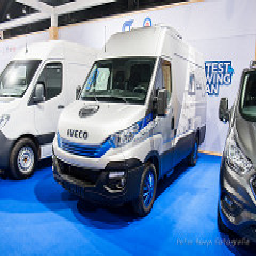} & \includegraphics[width=1.5cm, height=1.5cm]{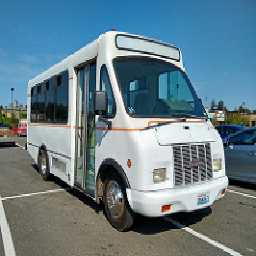} & \includegraphics[width=1.5cm, height=1.5cm]{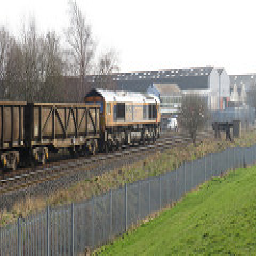} & \includegraphics[width=1.5cm, height=1.5cm]{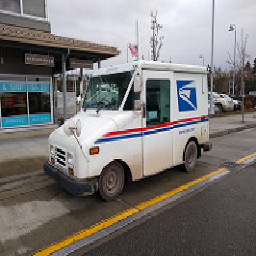} & \includegraphics[width=1.5cm, height=1.5cm]{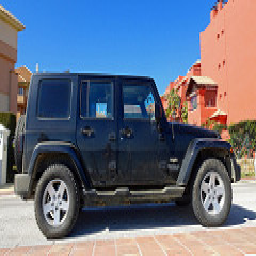} & \includegraphics[width=1.5cm, height=1.5cm]{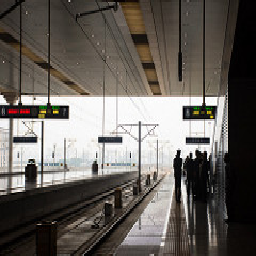} & \includegraphics[width=1.5cm, height=1.5cm]{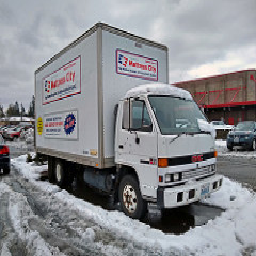} & \includegraphics[width=1.5cm, height=1.5cm]{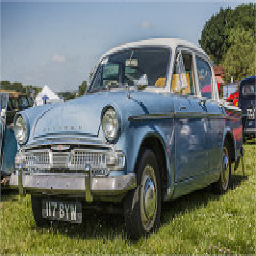} & \includegraphics[width=1.5cm, height=1.5cm]{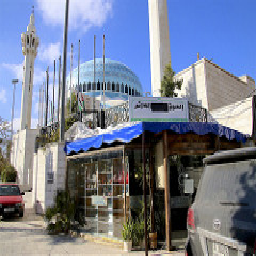} & \includegraphics[width=1.5cm, height=1.5cm]{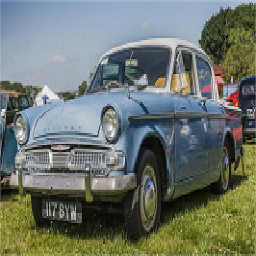} & \includegraphics[width=1.5cm, height=1.5cm]{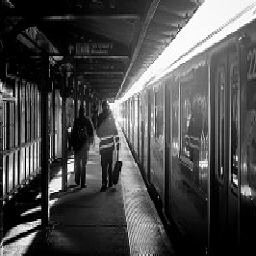} & \includegraphics[width=1.5cm, height=1.5cm]{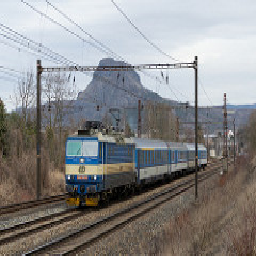} & \includegraphics[width=1.5cm, height=1.5cm]{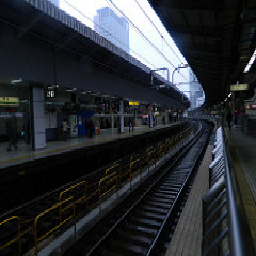} & \includegraphics[width=1.5cm, height=1.5cm]{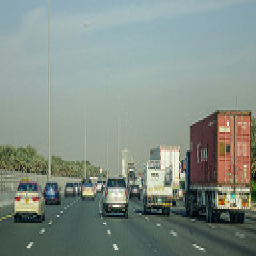} \\ 
 & \cmark & \cmark & \cmark & \cmark & \xmark & \cmark & \cmark & \cmark & \cmark & \cmark & \cmark & \xmark & \cmark & \cmark & \cmark & \cmark & \cmark & \xmark & \xmark & \cmark \\ 
\includegraphics[width=1.5cm, height=1.5cm]{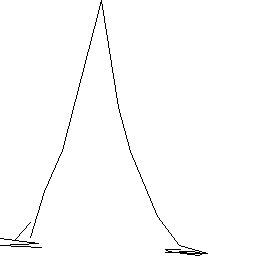} & \includegraphics[width=1.5cm, height=1.5cm]{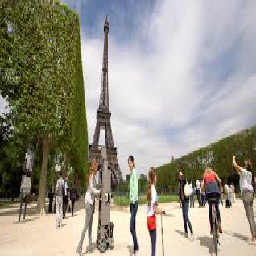} & \includegraphics[width=1.5cm, height=1.5cm]{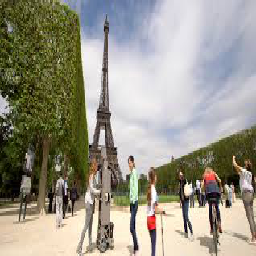} & \includegraphics[width=1.5cm, height=1.5cm]{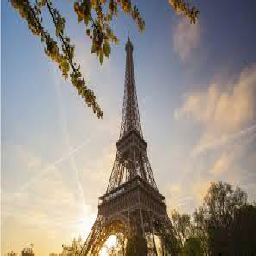} & \includegraphics[width=1.5cm, height=1.5cm]{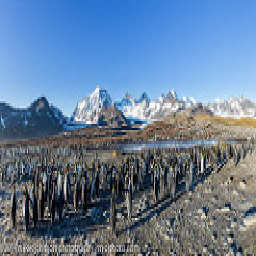} & \includegraphics[width=1.5cm, height=1.5cm]{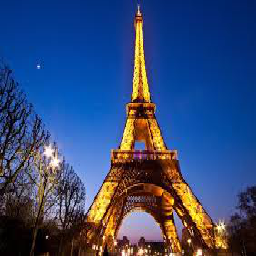} & \includegraphics[width=1.5cm, height=1.5cm]{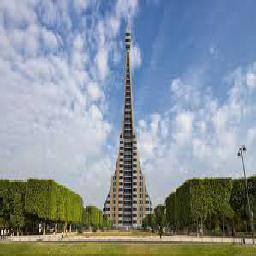} & \includegraphics[width=1.5cm, height=1.5cm]{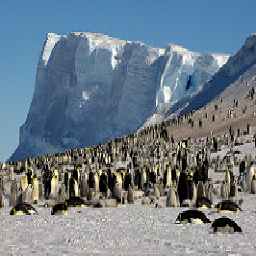} & \includegraphics[width=1.5cm, height=1.5cm]{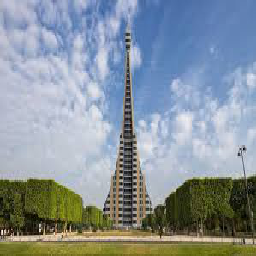} & \includegraphics[width=1.5cm, height=1.5cm]{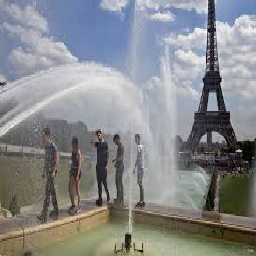} & \includegraphics[width=1.5cm, height=1.5cm]{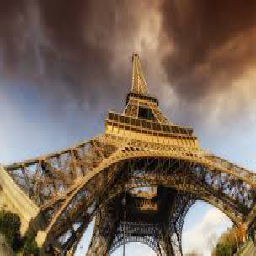} & \includegraphics[width=1.5cm, height=1.5cm]{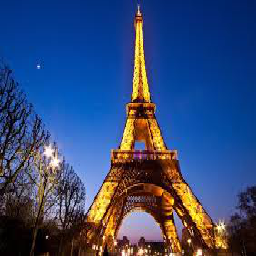} & \includegraphics[width=1.5cm, height=1.5cm]{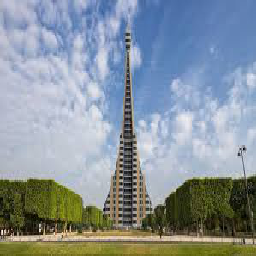} & \includegraphics[width=1.5cm, height=1.5cm]{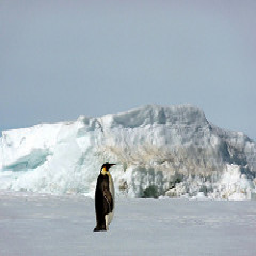} & \includegraphics[width=1.5cm, height=1.5cm]{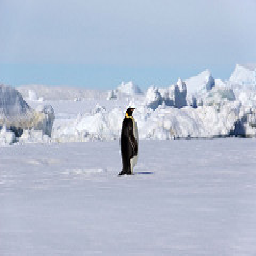} & \includegraphics[width=1.5cm, height=1.5cm]{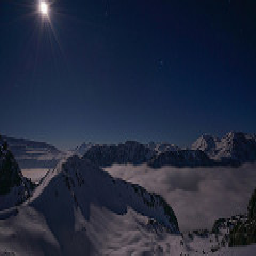} & \includegraphics[width=1.5cm, height=1.5cm]{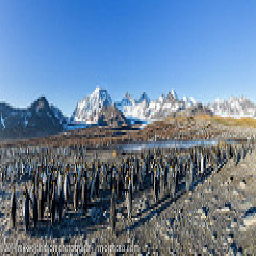} & \includegraphics[width=1.5cm, height=1.5cm]{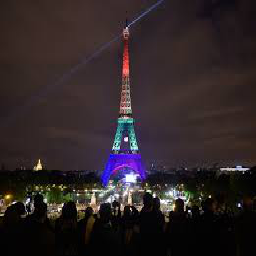} & \includegraphics[width=1.5cm, height=1.5cm]{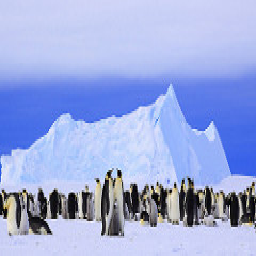} & \includegraphics[width=1.5cm, height=1.5cm]{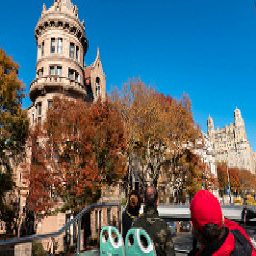} & \includegraphics[width=1.5cm, height=1.5cm]{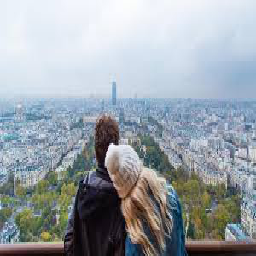} \\ 
 & \cmark & \cmark & \cmark & \xmark & \cmark & \cmark & \xmark & \cmark & \cmark & \cmark & \cmark & \cmark & \xmark & \xmark & \xmark & \xmark & \cmark & \xmark & \xmark & \cmark \\ 
\includegraphics[width=1.5cm, height=1.5cm]{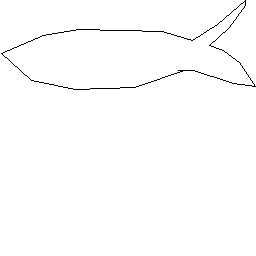} & \includegraphics[width=1.5cm, height=1.5cm]{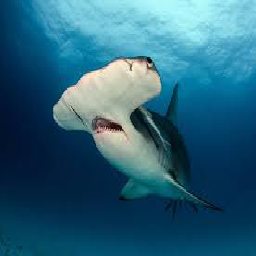} & \includegraphics[width=1.5cm, height=1.5cm]{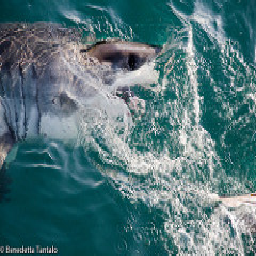} & \includegraphics[width=1.5cm, height=1.5cm]{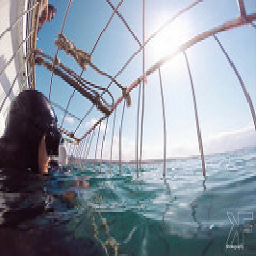} & \includegraphics[width=1.5cm, height=1.5cm]{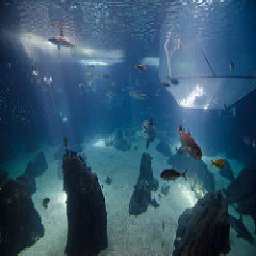} & \includegraphics[width=1.5cm, height=1.5cm]{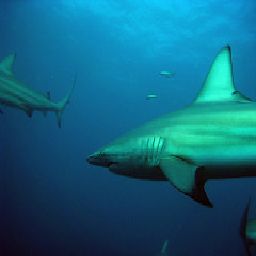} & \includegraphics[width=1.5cm, height=1.5cm]{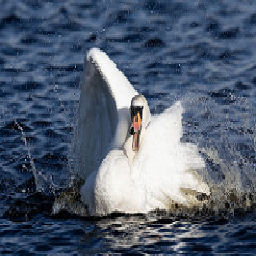} & \includegraphics[width=1.5cm, height=1.5cm]{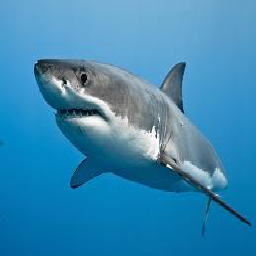} & \includegraphics[width=1.5cm, height=1.5cm]{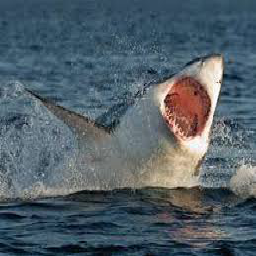} & \includegraphics[width=1.5cm, height=1.5cm]{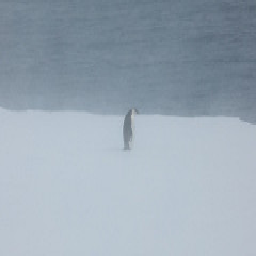} & \includegraphics[width=1.5cm, height=1.5cm]{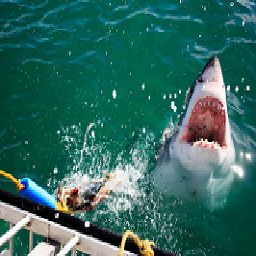} & \includegraphics[width=1.5cm, height=1.5cm]{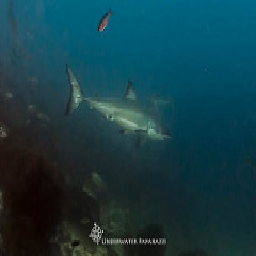} & \includegraphics[width=1.5cm, height=1.5cm]{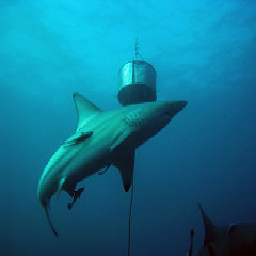} & \includegraphics[width=1.5cm, height=1.5cm]{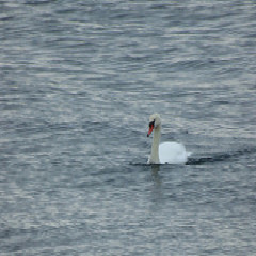} & \includegraphics[width=1.5cm, height=1.5cm]{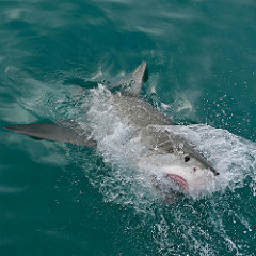} & \includegraphics[width=1.5cm, height=1.5cm]{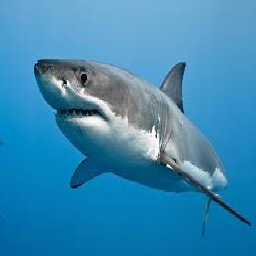} & \includegraphics[width=1.5cm, height=1.5cm]{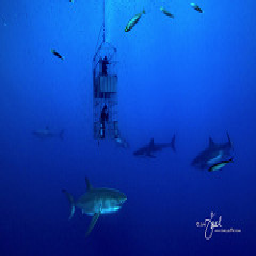} & \includegraphics[width=1.5cm, height=1.5cm]{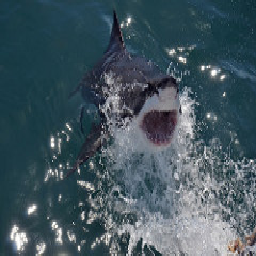} & \includegraphics[width=1.5cm, height=1.5cm]{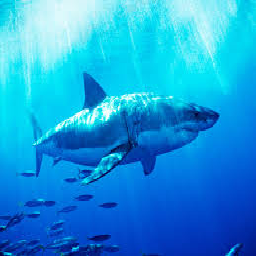} & \includegraphics[width=1.5cm, height=1.5cm]{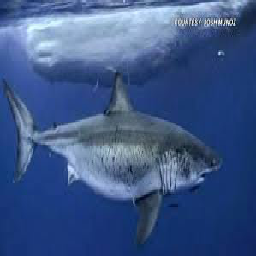} & \includegraphics[width=1.5cm, height=1.5cm]{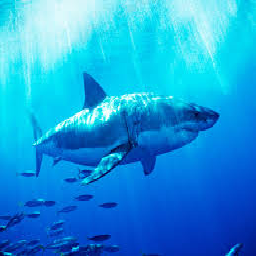} \\ 
 & \cmark & \cmark & \cmark & \cmark & \cmark & \xmark & \cmark & \cmark & \xmark & \cmark & \cmark & \cmark & \xmark & \cmark & \cmark & \cmark & \cmark & \cmark & \cmark & \cmark \\ 
\includegraphics[width=1.5cm, height=1.5cm]{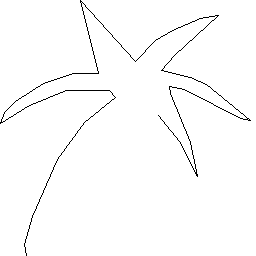} & \includegraphics[width=1.5cm, height=1.5cm]{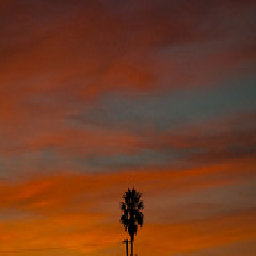} & \includegraphics[width=1.5cm, height=1.5cm]{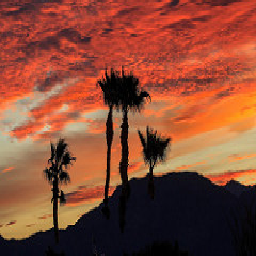} & \includegraphics[width=1.5cm, height=1.5cm]{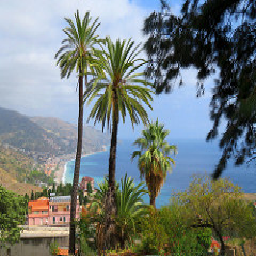} & \includegraphics[width=1.5cm, height=1.5cm]{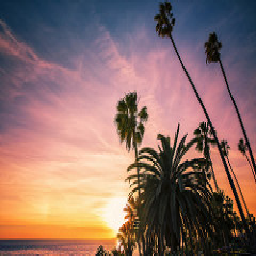} & \includegraphics[width=1.5cm, height=1.5cm]{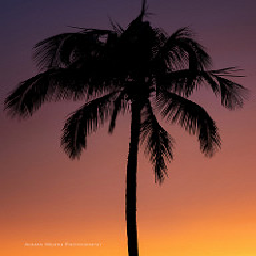} & \includegraphics[width=1.5cm, height=1.5cm]{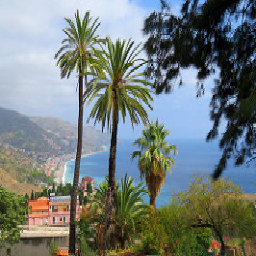} & \includegraphics[width=1.5cm, height=1.5cm]{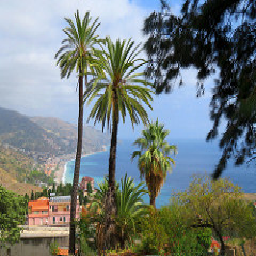} & \includegraphics[width=1.5cm, height=1.5cm]{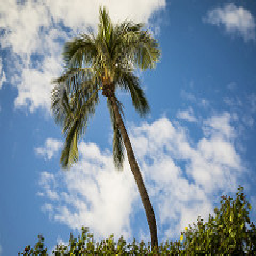} & \includegraphics[width=1.5cm, height=1.5cm]{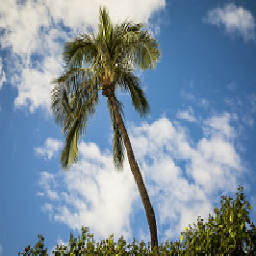} & \includegraphics[width=1.5cm, height=1.5cm]{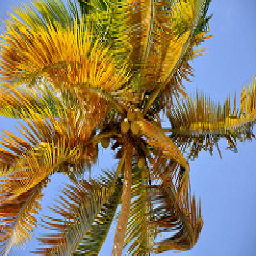} & \includegraphics[width=1.5cm, height=1.5cm]{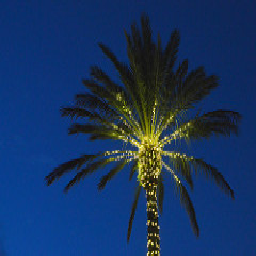} & \includegraphics[width=1.5cm, height=1.5cm]{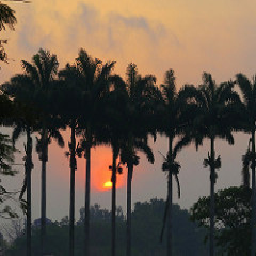} & \includegraphics[width=1.5cm, height=1.5cm]{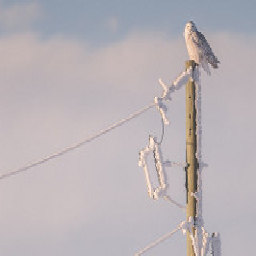} & \includegraphics[width=1.5cm, height=1.5cm]{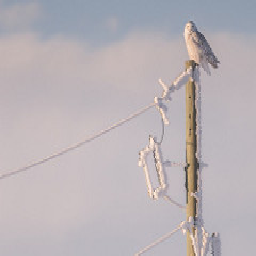} & \includegraphics[width=1.5cm, height=1.5cm]{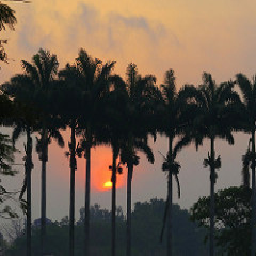} & \includegraphics[width=1.5cm, height=1.5cm]{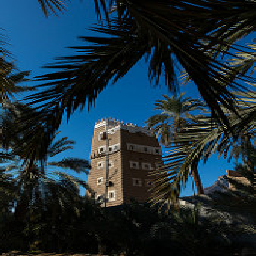} & \includegraphics[width=1.5cm, height=1.5cm]{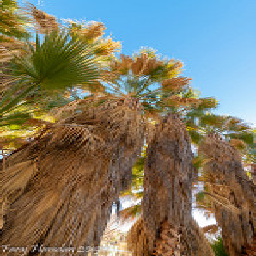} & \includegraphics[width=1.5cm, height=1.5cm]{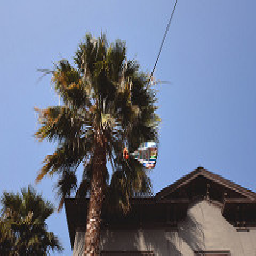} & \includegraphics[width=1.5cm, height=1.5cm]{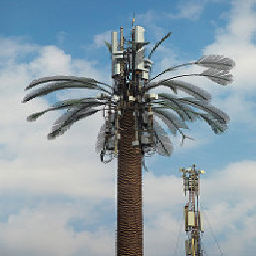} & \includegraphics[width=1.5cm, height=1.5cm]{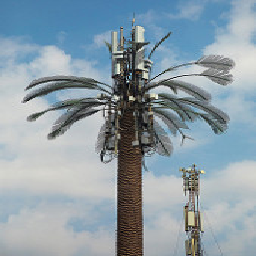} \\ 
 & \cmark & \cmark & \cmark & \cmark & \cmark & \cmark & \cmark & \cmark & \cmark & \cmark & \cmark & \xmark & \xmark & \xmark & \xmark & \cmark & \cmark & \cmark & \cmark & \cmark \\ 
\includegraphics[width=1.5cm, height=1.5cm]{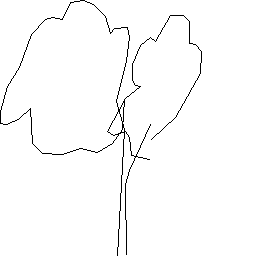} & \includegraphics[width=1.5cm, height=1.5cm]{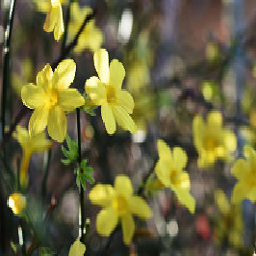} & \includegraphics[width=1.5cm, height=1.5cm]{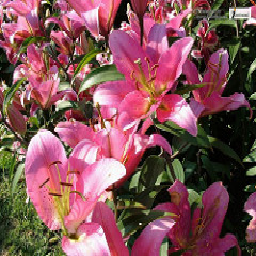} & \includegraphics[width=1.5cm, height=1.5cm]{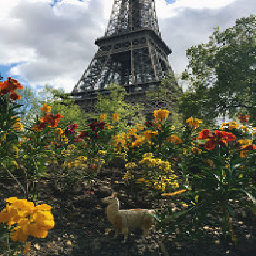} & \includegraphics[width=1.5cm, height=1.5cm]{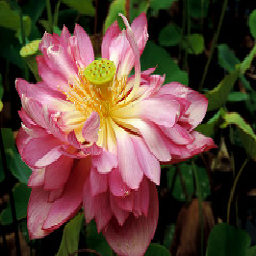} & \includegraphics[width=1.5cm, height=1.5cm]{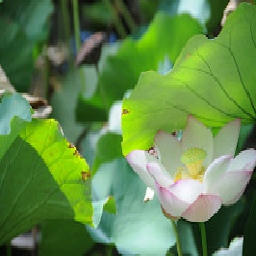} & \includegraphics[width=1.5cm, height=1.5cm]{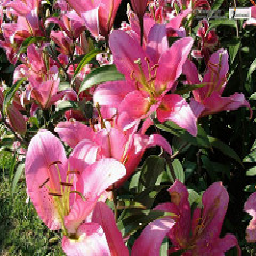} & \includegraphics[width=1.5cm, height=1.5cm]{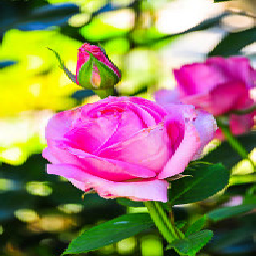} & \includegraphics[width=1.5cm, height=1.5cm]{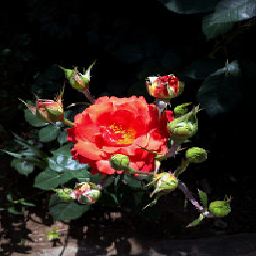} & \includegraphics[width=1.5cm, height=1.5cm]{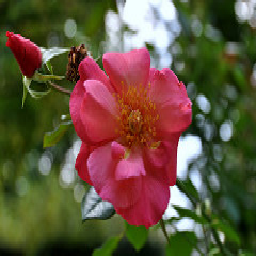} & \includegraphics[width=1.5cm, height=1.5cm]{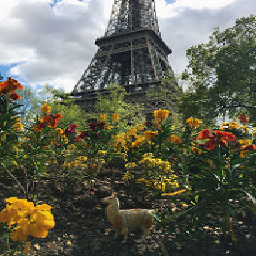} & \includegraphics[width=1.5cm, height=1.5cm]{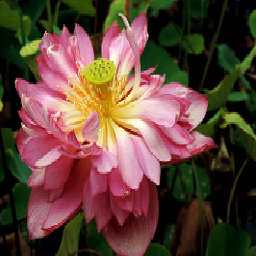} & \includegraphics[width=1.5cm, height=1.5cm]{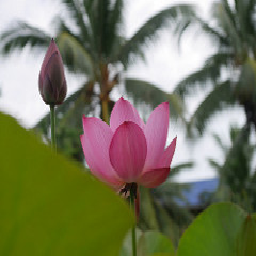} & \includegraphics[width=1.5cm, height=1.5cm]{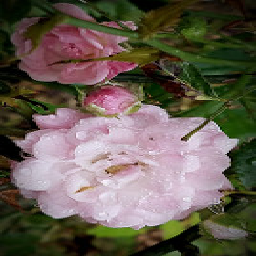} & \includegraphics[width=1.5cm, height=1.5cm]{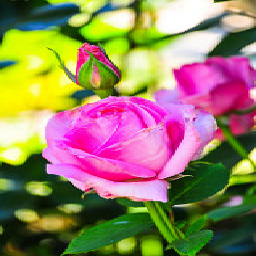} & \includegraphics[width=1.5cm, height=1.5cm]{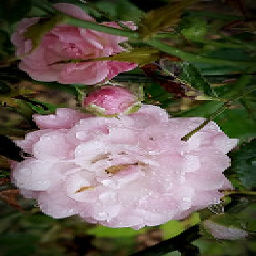} & \includegraphics[width=1.5cm, height=1.5cm]{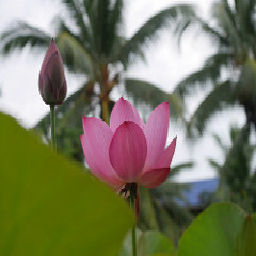} & \includegraphics[width=1.5cm, height=1.5cm]{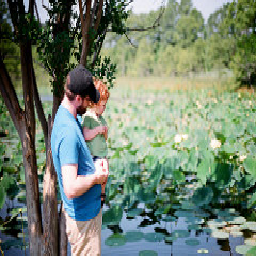} & \includegraphics[width=1.5cm, height=1.5cm]{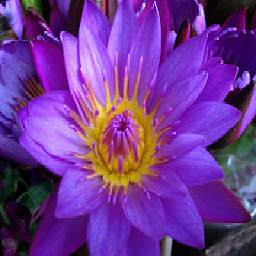} & \includegraphics[width=1.5cm, height=1.5cm]{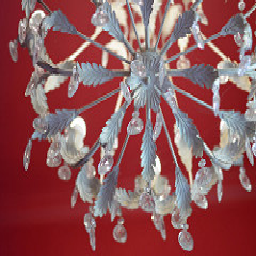} & \includegraphics[width=1.5cm, height=1.5cm]{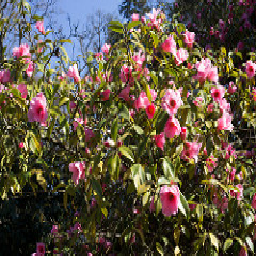} \\ 
 & \xmark & \cmark & \xmark & \cmark & \cmark & \cmark & \cmark & \cmark & \cmark & \xmark & \cmark & \cmark & \cmark & \cmark & \cmark & \cmark & \cmark & \cmark & \xmark & \xmark \\ 
\end{tabular}
}\end{center}
\caption{Top-20 zero-shot SBIR results obtained by our SEM-PCYC model on the QuickDraw (Extended) dataset are shown here according to the Euclidean distances, where the green ticks denote the correctly retrieved candidates, whereas the red crosses indicate the wrong retrievals. (best viewed in color)}
\label{fig:qual_results_quickdraw}
\end{figure*}

\myparagraph{Qualitative Results.} 
We analyze the retrieval performance of our proposed model qualitatively in~\fig{fig:qual_results_sketchy}, ~\fig{fig:qual_results_tu-berlin} and \fig{fig:qual_results_quickdraw}. Some notable examples are as follows. Sketch query of \texttt{tank} retrieves some examples of \texttt{motorcycle} probably because both of them have wheels in common (row $1$ of~\fig{fig:qual_results_sketchy}). Similar explanation can be given in the case of \texttt{car} and \texttt{motorcycle} (row $1$ of~\fig{fig:qual_results_quickdraw}). For having visual and semantic similarity, sketching \texttt{guitar} retrieves some \texttt{violin}s (row $2$ of~\fig{fig:qual_results_sketchy}). This can also be observed in case of \texttt{train} and \texttt{van} in row $2$ of~\fig{fig:qual_results_quickdraw}.

For having visual and semantic similarity, querying \texttt{bear} retrieves some \texttt{squirrel}s (row $3$ of~\fig{fig:qual_results_sketchy}). Querying objects with wheel (\eg, \texttt{wheelchair}, \texttt{motorcycle}) sometime wrongly retrieves other vehicles, probably because of having wheels in common (row $6$ of~\fig{fig:qual_results_sketchy}). Sketch query of \texttt{spoon} retrieves some examples of \texttt{racket} (row $4$ of~\fig{fig:qual_results_sketchy}), possibly for having significant visual similarity. Sketch of \texttt{burger} retrieves some examples of \texttt{jack-o-lantern} (row $5$ of~\fig{fig:qual_results_sketchy}), perhaps for having same shape. Querying \texttt{castle}, retrieves images having large portion of sky (row $2$ of~\fig{fig:qual_results_tu-berlin}), because the images of its semantically similar classes, such as, \texttt{skyscraper}, \texttt{church}, are mostly captured with sky in background. Similar phenomenon can be observed in case of \texttt{tree} and \texttt{electrical post} in row $5$ of~\fig{fig:qual_results_quickdraw}. Querying \texttt{duck}, retrieves images of \texttt{swan} or \texttt{shark} (row $4$ of~\fig{fig:qual_results_tu-berlin}), probably for having watery background in common. Sketch of \texttt{pickup truck} retrieves some images from \texttt{traffic light} class for having a truck like object in the scene (row $3$ of~\fig{fig:qual_results_tu-berlin}). Sketching \texttt{bookshelf} retrieves some examples of \texttt{cabinet} for having significant visual and semantic similarity (row $5$ of~\fig{fig:qual_results_tu-berlin}).

Sometimes too much abstraction in sketches can produce wrong retrieval results. For example, in row $3$ of~\fig{fig:qual_results_quickdraw}, it is difficult to understand whether the sketch is of \texttt{eiffel tower} or any other tower or a hill. Furthermore, we have observed certain ambiguities in annotation of images in QuickDraw dataset. Currently, the images are much complex, which often contain two or more objects, and most of the currently available SBIR datasets provide single object annotation ignoring the object in background. For example see row $6$ of~\fig{fig:qual_results_quickdraw}, many of the wrongly retrieved images truly contain \texttt{flower}, whereas some of them are annotated as \texttt{tower} or \texttt{trees} etc. Additionally, as the images from QuickDraw dataset are collected from the Flickr website, it contains many subsequent captures which can be confused as identical frames. Hence, although some retrievals on QuickDraw dataset appear identical, they are not in terms of the actual pixel values.

In general, we observe that the wrongly retrieved candidates mostly have a closer visual and semantic relevance with the queried ones. This effect is more prominent in TU-Berlin dataset, which may be due to the inter-class similarity of sketches between different classes. As shown in \fig{tab:sketches_tu_berlin}, the classes \texttt{swan}, \texttt{duck} and \texttt{owl}, \texttt{penguin} have substantial visual similarity, and all of them are \texttt{standing bird} which is a separate class of the same dataset. Therefore, for TU-Berlin dataset, it is challenging to generalize the \emph{unseen} classes from the learned representation of \emph{seen} classes.

\begin{table*}[!ht]
\centering
\resizebox{\textwidth}{!}{
\begin{tabular}{ccc|ccc|ccc|ccc|ccc}
\hline
\multicolumn{3}{c|}{\textbf{Text Embedding}} & \multicolumn{3}{c|}{\textbf{Hierarchical Embedding}} & \multicolumn{3}{c|}{\textbf{Sketchy (Extended)}} & \multicolumn{3}{c|}{\textbf{TU-Berlin (Extended)}} & \multicolumn{3}{c}{\textbf{QuickDraw (Extended)}}\\
\textbf{GloVe}~\cite{Pennington2014GloVe} & \textbf{Word2Vec}~\cite{Mikolov2013} & \textbf{FastText}~\cite{Joulin2016FastText} & \textbf{Path} & \textbf{Lin}~\cite{Lin1998ITSim} & \textbf{Ji-Cn}~\cite{Jiang1997SemSim} & $32$ dim & $64$ dim & $128$ dim & $32$ dim & $64$ dim & $128$ dim & $32$ dim & $64$ dim & $128$ dim\\
\hline
\checkmark &  &  &  &  &  & $0.237$ & $0.284$ & $0.321$ & $0.193$ & $0.228$ & $0.239$ & $0.127$ & $0.149$ & $0.165$ \\
 & \checkmark &  &  &  &  & $0.279$ & $0.330$ & $0.365$ & $0.199$ & $0.232$ & $0.243$ & $0.124$ & $0.132$ & $0.167$ \\
 &  & \checkmark &  &  &  & $0.264$ & $0.344$ & $0.343$ & $0.219$ & $0.262$ & $0.265$ & $0.127$ & $0.155$ & $0.165$ \\
 &  &  & \checkmark &  &  & $0.290$ & $0.314$ & $0.365$ & $0.201$ & $0.224$ & $0.255$ & $0.121$ & $0.138$ & $0.155$ \\
 &  &  &  & \checkmark &  & $0.201$ & $0.248$ & $0.264$ & $0.152$ & $0.169$ & $0.182$ & $0.130$ & $0.149$ & $0.158$ \\
 &  &  &  &  & \checkmark & $0.263$ & $0.308$ & $0.352$ & $0.208$ & $0.227$ & $0.239$ & $0.151$ & $0.146$ & $0.152$ \\
 \hline
\checkmark &  &  & \checkmark &  &  & $0.259$ & $0.338$ & $0.356$ & $0.238$ & $0.276$ & $0.281$ & $0.129$ & $0.176$ & $0.158$ \\
\checkmark &  &  &  & \checkmark &  & $0.275$ & $0.299$ & $0.318$ & $0.241$ & $0.253$ & $0.264$ & $0.130$ & $\mathbf{0.177}$ & $0.175$ \\
\checkmark &  &  &  &  & \checkmark & $0.273$ & $0.285$ & $0.291$ & $0.238$ & $0.243$ & $0.251$ & $\mathbf{0.149}$ & $0.163$ & $0.165$ \\
 & \checkmark &  & \checkmark &  &  & $0.298$ & $0.340$ & $0.368$ & $\mathbf{0.278}$ & $\mathbf{0.297}$ & $\mathbf{0.301}$ & $0.145$ & $0.150$ & $0.164$ \\
 & \checkmark &  &  & \checkmark &  & $0.282$ & $0.288$ & $0.306$ & $0.253$ & $0.264$ & $0.282$ & $0.142$ & $0.169$ & $0.175$ \\
 & \checkmark &  &  &  & \checkmark & $\mathbf{0.307}$ & $\mathbf{0.349}$ & $0.372$ & $0.273$ & $0.291$ & $0.298$ & $0.145$ & $0.155$ & $\mathbf{0.184}$ \\
 &  & \checkmark & \checkmark &  &  & $0.329$ & $\mathbf{0.349}$ & $\mathbf{0.400}$ & $0.242$ & $\mathbf{0.297}$ & $0.289$ & $0.137$ & $0.151$ & $0.153$ \\
 &  & \checkmark &  & \checkmark &  & $0.304$ & $0.344$ & $0.352$ & $0.254$ & $0.296$ & $0.286$ & $0.129$ & $0.150$ & $0.147$ \\
 &  & \checkmark &  &  & \checkmark & $0.317$ & $0.299$ & $0.381$ & $0.246$ & $0.279$ & $0.326$ & $0.124$ & $0.144$ & $0.182$ \\
\hline
\end{tabular}}
\caption{Zero-shot SBIR mAP@all using different semantic embeddings (top) and their combinations (bottom) with $32$, $64$ and $128$ dimension.}
\label{tab:res_sem}
\end{table*}

\myparagraph{Effect of Side-Information.} In zero-shot learning, side information is as important as the visual information as it is the only means the model can discover similarities between classes. As the type of side information has a high effect in performance of any method, we analyze the effect of side-information and present zero-shot SBIR results by considering different side information and their combinations. We compare the effect of using GloVe~\cite{Pennington2014GloVe}, Word2Vec~\cite{Mikolov2013a} and FastText~\cite{Joulin2016FastText} as text-based model, and three similarity measurements, \ie~path, Lin~\cite{Lin1998ITSim} and Jiang-Conrath~\cite{Jiang1997SemSim} for constructing three different side information that are based on WordNet hierarchy. \tab{tab:res_sem} contains the quantitative results on Sketchy, TU-Berlin and QuickDraw datasets with different side information mentioned and their combinations, where we set $M=32, 64, 128$. We have observed that in majority of cases combining different side information increases the performance by $1\%$ to $3\%$. 

On Sketchy, the combination of Word2Vec and Jiang-Conrath hierarchical similarity as well as FastText and Path reach the highest mAP of $0.349$ with 64d embedding while on TU Berlin dataset, in addition to the combination of Word2Vec and path similarity, FastText and Path lead with $0.297$ mAP with 64d, and for QuickDraw the combination of GloVe and Lin hierarchical similarity reaches to $0.177$ for $64$d. We conclude from these experiments that indeed text-based and hierarchy-based class embeddings are complementary.

\myparagraph{Effect of Visual Features.} Visual features are also crucial for the zero-shot SBIR task. For having some overview on that, addition to VGG-16~\cite{Simonyan2014} features obtained before the last \texttt{fc} layer, we also consider SE-ResNet-50~\cite{Hu2019SENet,He2015ResNet} features, and perform zero-shot SBIR experiments on the Sketchy, TU-Berlin and QuickDraw datasets with different semantic models mentioned above. In \tab{tab:res_vis_sem}, we present the mAP@all values obtained by the considered visual features and semantic models, where we observe that SE-ResNet-50 features work consistently better than VGG-16 on all the three datasets. Especially, the performance gain on the challenging TU-Berlin dataset should be noted, which we speculate as the benefit of feature calibration strategy involved in the SE blocks, that effectively produces robust features minimizing inter-class confusion as presented in \fig{tab:sketches_tu_berlin}.

\begin{table*}[!ht]
\centering
\begin{tabular}{l|l|ccc}
\hline
\textbf{Visual} & \textbf{Semantic} & \textbf{Sketchy} & \textbf{TU-Berlin} & \textbf{QuickDraw} \\
\textbf{Features} & \textbf{Model} & \textbf{(Extended)} & \textbf{(Extended)} & \textbf{(Extended)} \\
\hline
\multirow{6}{*}{VGG-16~\cite{Simonyan2014}} & GloVe~\cite{Pennington2014GloVe} & $0.284$ & $0.228$ & $0.149$ \\
 & Word2Vec~\cite{Mikolov2013} & $0.330$ & $0.232$ & $0.132$ \\
 & FastText~\cite{Joulin2016FastText} & $\mathbf{0.344}$ & $\mathbf{0.262}$ & $\mathbf{0.155}$ \\
 & Path & $0.314$ & $0.224$ & $0.138$ \\
 & Lin~\cite{Lin1998ITSim} & $0.248$ & $0.169$ & $0.149$ \\
 & Ji-Cn~\cite{Jiang1997SemSim} & $0.308$ & $0.227$ & $0.146$ \\
\hline
\multirow{6}{*}{SE-ResNet-50~\cite{Hu2019SENet,He2015ResNet}} & GloVe~\cite{Pennington2014GloVe} & $0.344$ & $0.329$ & $\mathbf{0.172}$ \\
 & Word2Vec~\cite{Mikolov2013} & $\mathbf{0.385}$ & $0.305$ & $0.151$ \\
 & FastText~\cite{Joulin2016FastText} & $0.368$ & $\mathbf{0.349}$ & $0.171$ \\
 & Path & $0.330$ & $0.317$ & $0.156$ \\
 & Lin~\cite{Lin1998ITSim} & $0.362$ & $0.318$ & $0.146$ \\
 & Ji-Cn~\cite{Jiang1997SemSim} & $0.384$ & $0.306$ & $0.161$ \\
\hline
\end{tabular}
\caption{Zero-shot SBIR mAP@all using different semantic embeddings either with VGG-16 or ResNet-50 visual features while the dimension is kept equal to $64$.}
\label{tab:res_vis_sem}
\end{table*}

\myparagraph{Model Ablations.} The baselines of our ablation study are built by modifying some parts of the SEM-PCYC model and analyze the effect of different losses of our model. First, we train the model only with adversarial loss, and then alternatively add cycle consistency and classification loss for the training. Second, we train our model by only withdrawing the adversarial loss for the semantic domain, which should indicate the effect of side information in our case. We also train the model without the side information selection mechanism, for that, we only take the original text or hierarchical embedding or their combination as side information, which can give an idea on the advantage of selecting side information via the auto-encoder. Next, we experiment reducing the dimensionality of the class embedding to a percentage of the full dimensionality. Finally, to demonstrate the effectiveness of the regularizer used in the auto-encoder for selecting discriminative side information, we experiment by making $\lambda=0$ in \eq{eqn:aenc_loss}.

{
\setlength{\tabcolsep}{3pt}
\renewcommand{\arraystretch}{1.2} 
\begin{table*}[!t]
\centering
\begin{tabular}{lccc}
\hline
\multirow{2}{*}{\textbf{Description}} & \textbf{Sketchy} & \textbf{TU-Berlin} & \textbf{QuickDraw} \\
  & \textbf{(Extended)} & \textbf{(Extended)} & \textbf{(Extended)} \\
\hline
Only adversarial loss & $0.128$ & $0.109$ & $0.065$ \\
Adversarial + cycle consistency loss & $0.147$ & $0.131$ & $0.078$ \\
Adversarial + classification loss & $0.140$ & $0.127$ & $0.076$ \\
Adversarial (sketch + image) + cycle consistency + classification loss & $0.213$ & $0.154$ & $0.075$ \\
Without selecting side information & $0.382$ & $0.299$ & $0.185$ \\
Without regularizer in~\eq{eqn:aenc_loss} & $0.323$ & $0.273$ & $0.158$ \\
\textbf{SEM-PCYC (full model)} & $\mathbf{0.349}$ & $\mathbf{0.297}$ & $\mathbf{0.177}$ \\
\hline
\end{tabular}
\caption{Ablation study on our SEM-PCYC model ($64$d) on three datasets (measured with mAP@all).}
\label{tab:ablation_study_zero_shot}
\end{table*}
}

The mAP@all values obtained by respective baselines mentioned above are shown in~\tab{tab:ablation_study_zero_shot}. We consider the best side information setting according to~\tab{tab:res_sem} depending on the dataset. The assessed baselines have typically underperformed the full SEM-PCYC model. Only with adversarial losses, the performance of our system drops significantly. We suspect that only adversarial training although maps sketch and image input to a semantic space, there is no guarantee that sketch-image pairs of same category are matched. This is because adversarial training only ensures the mapping of input modality to target modality that matches its empirical distribution~\cite{Zhu2017CycleGAN}, but does not guarantee an individual input and output are paired up. 

Imposing cycle-consistency constraint ensures the one-to-one correspondence of sketch-image categories. However, the performance of our system does not improve substantially while the model is trained both with adversarial and cycle consistency loss. We speculate that this issue could be due to the lack of inter-category discriminating power of the learned embedding functions; for that, we set a classification criteria to train discriminating cross-modal embedding functions. We further observe that only imposing classification criteria together with adversarial loss, neither improves the retrieval results. We conjecture that in this case the learned embedding could be very discriminative but the two modalities might be matched in wrong way. Hence, it can be concluded that all these three losses are complimentary to each other and absolutely essential for effective zero-shot SBIR. 

Next, we analyze the effect of side information and notice that without the adversarial loss for the semantic domain, our model performs better than the previously mentioned three  configurations but does not reach near to the full model. This is due to the fact that without semantic mapping, the resulting embeddings are not semantically related to each other, which do not help in cross modal retrieval in zero-shot scenario. We further observe that without the encoded and compact side information, we achieve better mAP@all with a compromise on retrieval time, as the original dimension ($354+300=654$d for Sketchy, $664+300=964$d for TU-Berlin and $344+300=644$d for QuickDraw) of considered side information is much higher than the encoded ones ($64$d). We further investigate by reducing its dimension as a percentage of the original one (see \fig{fig:plots}(c)), and we have observed that at the beginning, reducing a small part (mostly $5\%$ to $30\%$) usually leads to a better performance, which reveals that not all the side information are necessary for effective zero-shot SBIR and some of them are even harmful. In fact, the first removed ones have low information content, and can be regarded as noise. 

We have also perceived that removing more side information (beyond $20\%$ to $40\%$) deteriorates the performance of the system, which is quite justifiable because the compressing mechanism of auto-encoder progressively removes important and predictable side information. However, it can be observed that with highly compressed side information as well, our model provides a very good deal with performance and retrieval time. 

Finally, without using the regularizer in \eq{eqn:aenc_loss} although our system performs reasonably, the mAP@all value is still lower than the best obtained performance. We explain this as a benefit of using $\ell_{21}$-norm based regularizer that effectively select representative side information.

\begin{figure*}[!t]
\resizebox{\textwidth}{!}{
\begin{tabular}{ccc}
\includegraphics[width=0.33\textwidth]{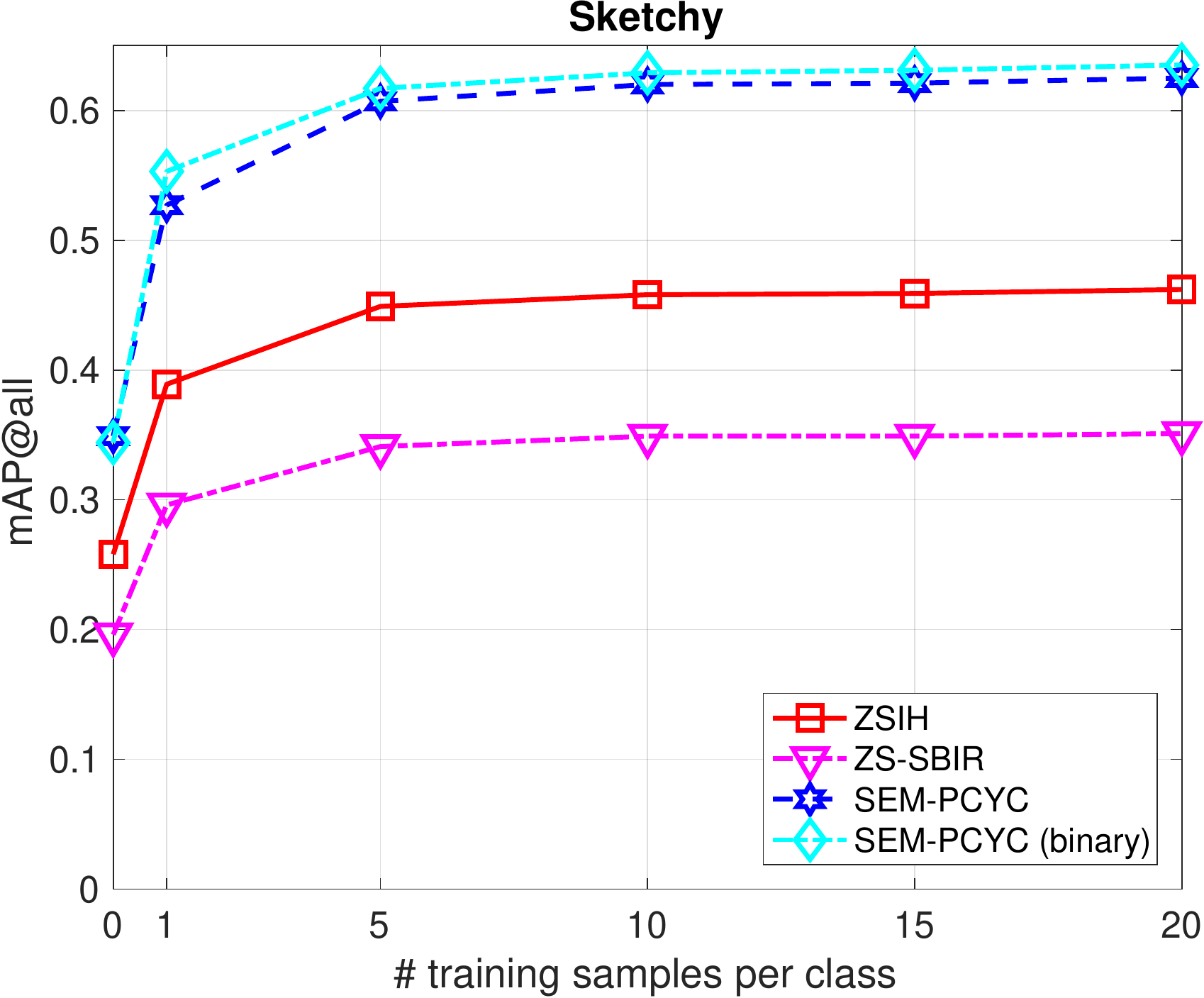} & \includegraphics[width=0.33\textwidth]{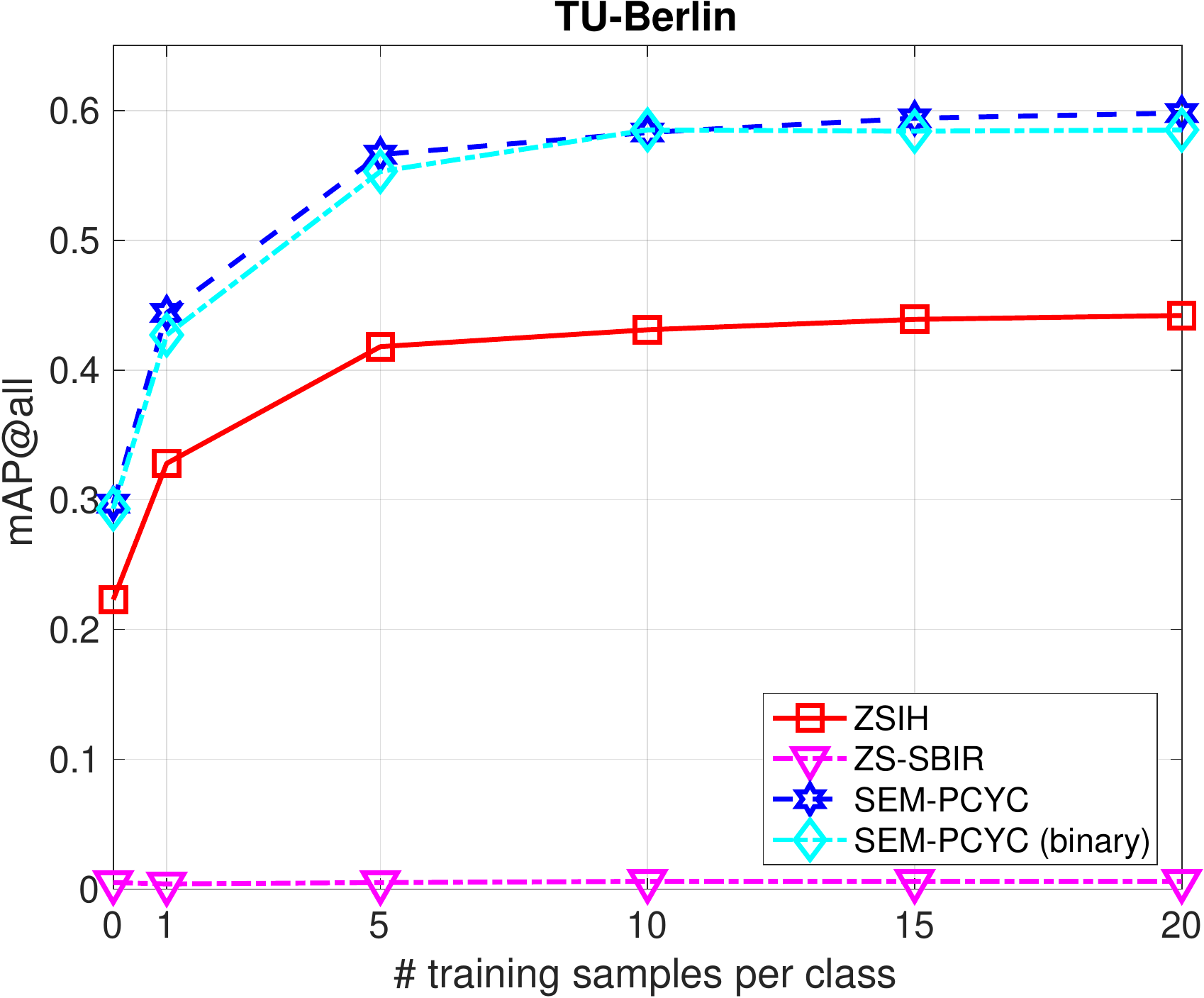} & \includegraphics[width=0.33\textwidth]{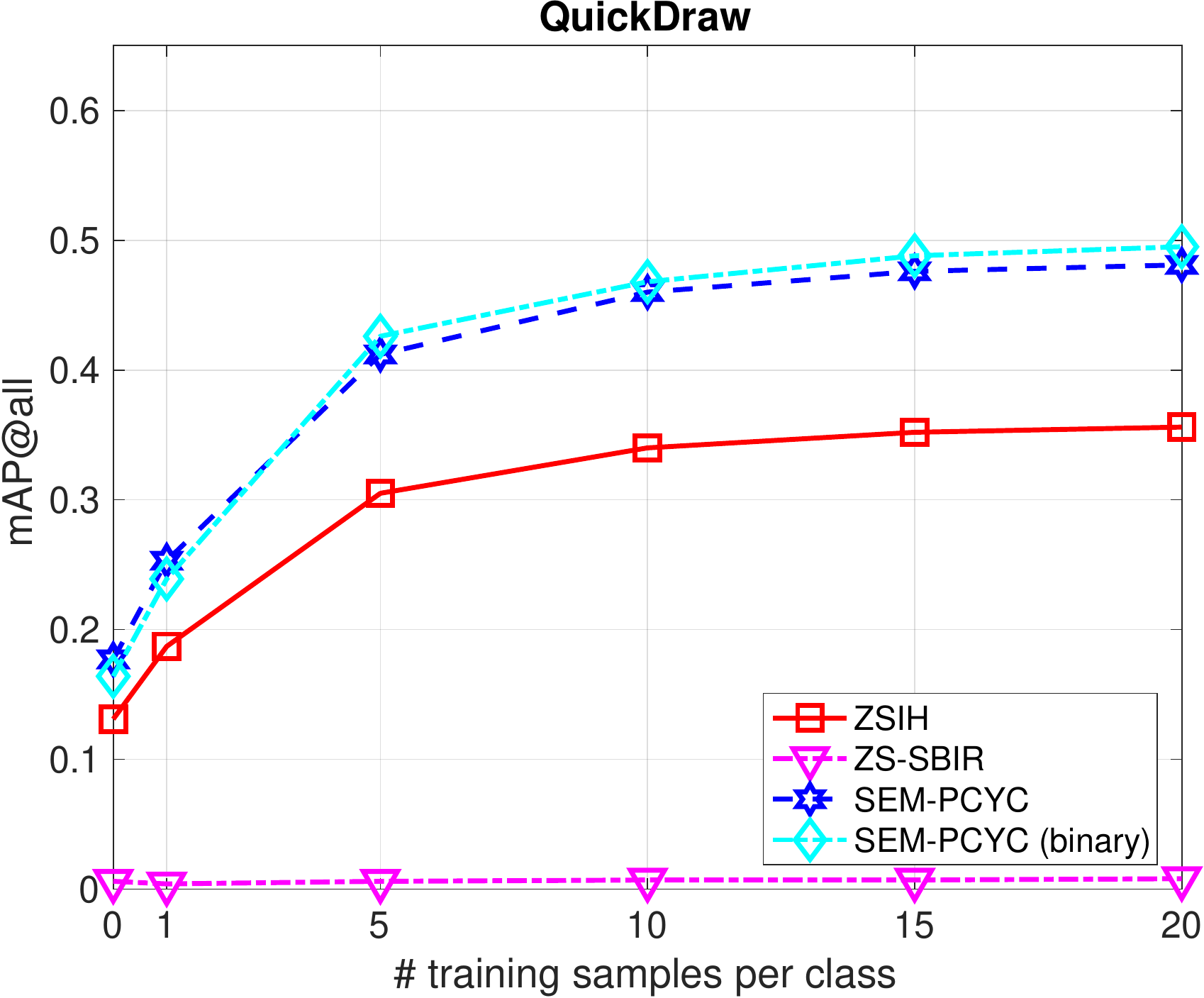} \\
(a) & (b) & (c) \\
\includegraphics[width=0.33\textwidth]{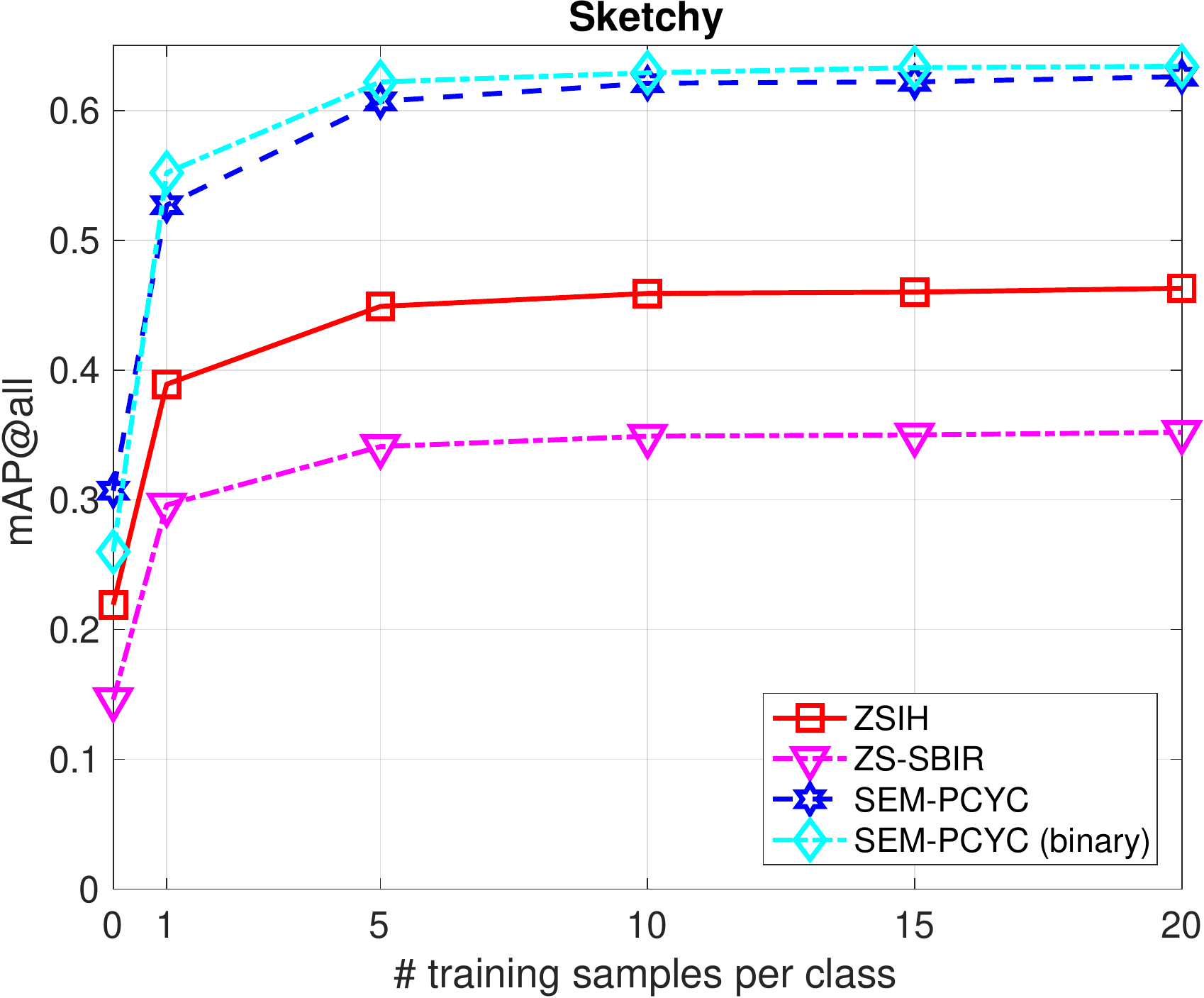} & \includegraphics[width=0.33\textwidth]{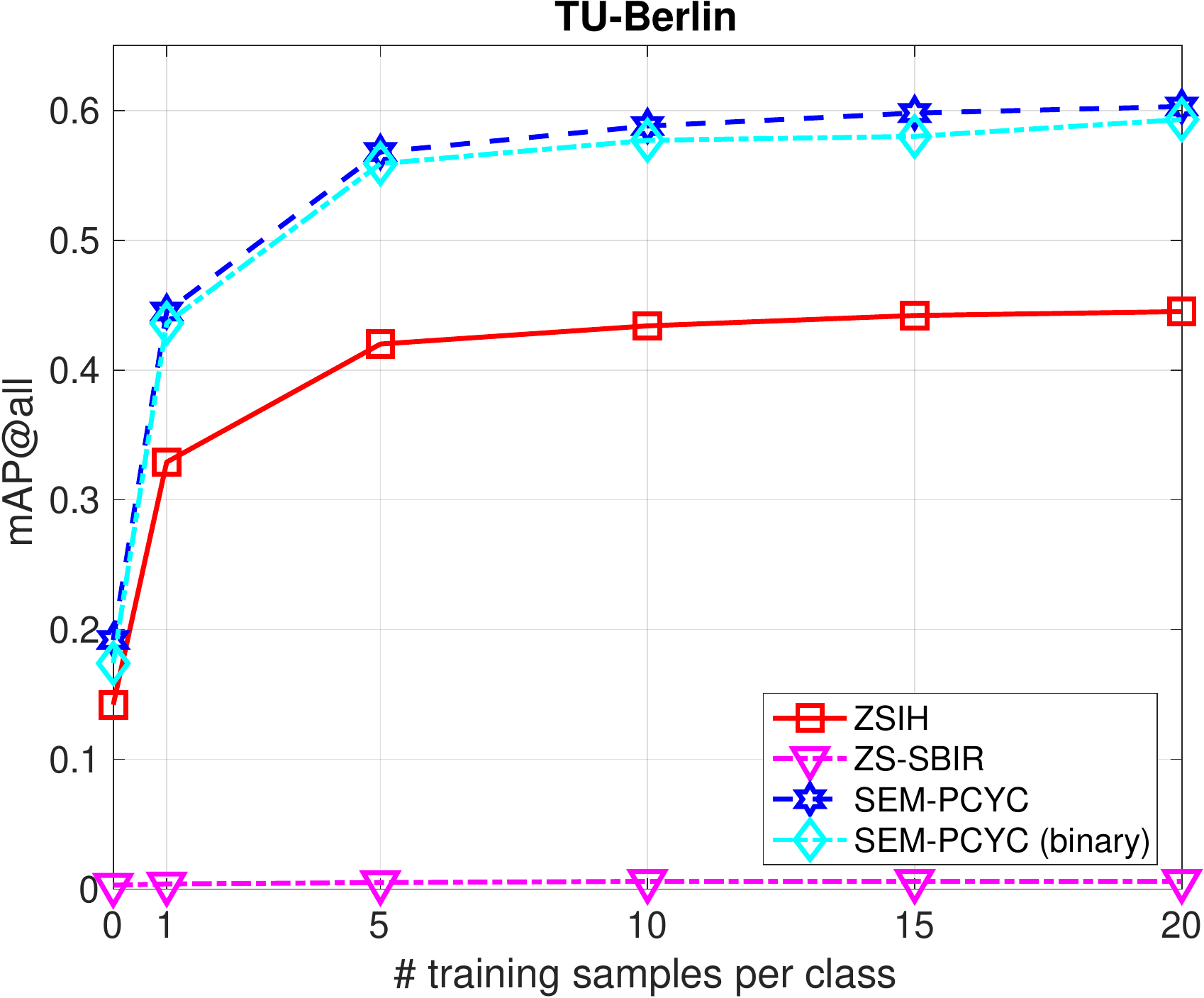} & \includegraphics[width=0.33\textwidth]{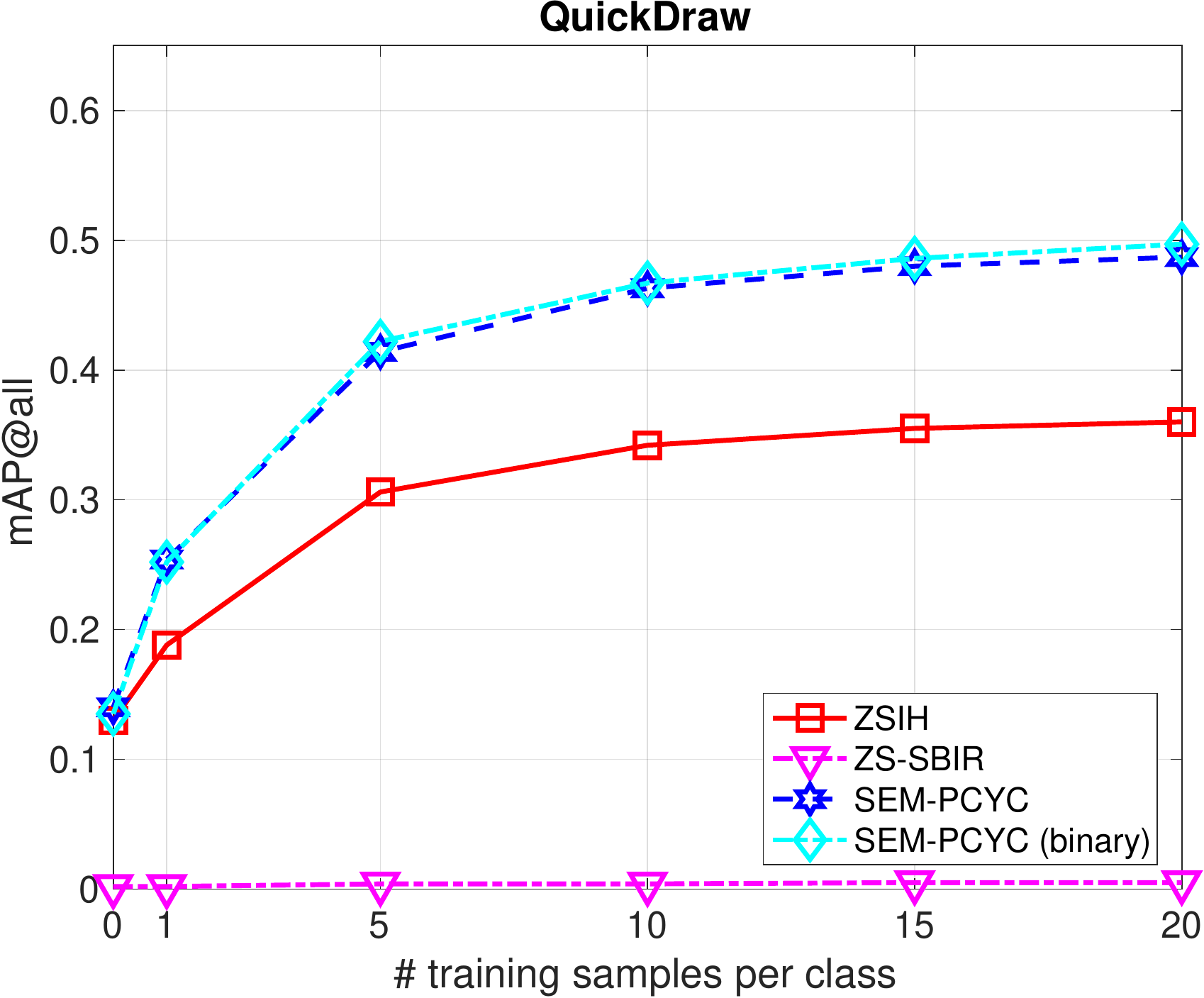} \\
(d) & (e) & (f) \\
\end{tabular}}
\caption{Few-shot sketch-based image retrieval (k=0,1,5,10,15,20) performance comparison with three existing state-of-the-art methods on Sketchy, TU-Berlin and Quickdraw datasets. Top: Few-shot Sketch Based Image Retrieval results, Bottom: Generalized Few-Shot Sketch-Based Image Retrieval results.}
\label{fig:few_shot}
\end{figure*}

\begin{figure*}
\begin{center}
\resizebox{\textwidth}{!}{
\begin{tabular}{lc@{}c@{}c@{}c@{}c@{}cc@{}c@{}c@{}c@{}c@{}cc@{}c@{}c@{}c@{}c@{}cc@{}c@{}c@{}c@{}c@{}c}
\Huge{Query} && \multicolumn{5}{c}{\Huge 0-shot} && \multicolumn{5}{c}{\Huge 1-shot} && \multicolumn{5}{c}{\Huge 5-shot} && \multicolumn{5}{c}{\Huge 10-shot} \\
\includegraphics[width=1.5cm, height=1.5cm]{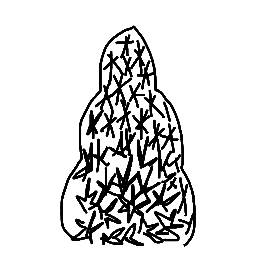} && \includegraphics[width=1.5cm, height=1.5cm]{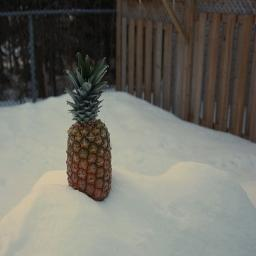} & \includegraphics[width=1.5cm, height=1.5cm]{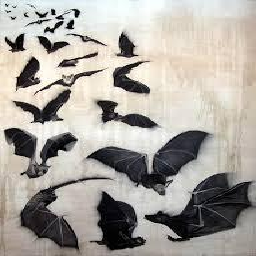} & \includegraphics[width=1.5cm, height=1.5cm]{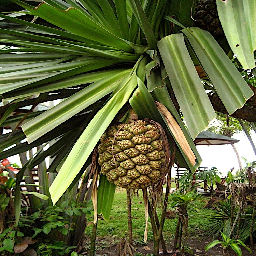} & \includegraphics[width=1.5cm, height=1.5cm]{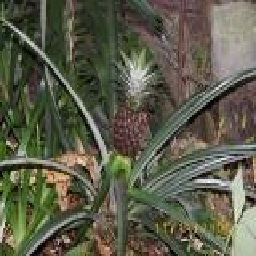} & \includegraphics[width=1.5cm, height=1.5cm]{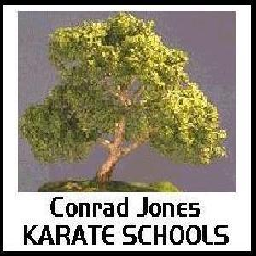} && \includegraphics[width=1.5cm, height=1.5cm]{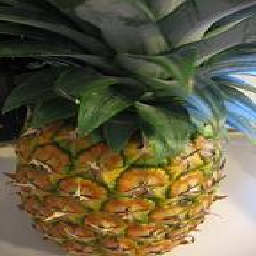} & \includegraphics[width=1.5cm, height=1.5cm]{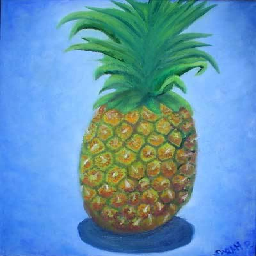} & \includegraphics[width=1.5cm, height=1.5cm]{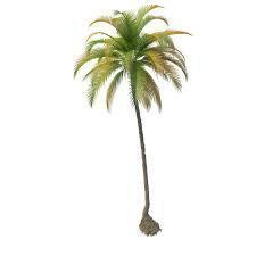} & \includegraphics[width=1.5cm, height=1.5cm]{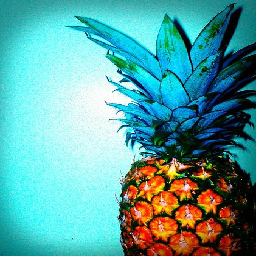} & \includegraphics[width=1.5cm, height=1.5cm]{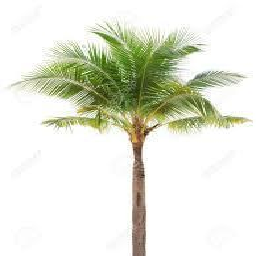} && \includegraphics[width=1.5cm, height=1.5cm]{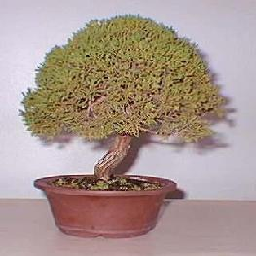} & \includegraphics[width=1.5cm, height=1.5cm]{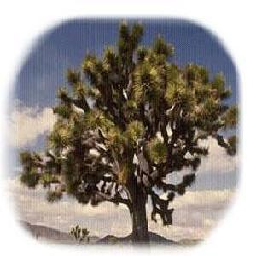} & \includegraphics[width=1.5cm, height=1.5cm]{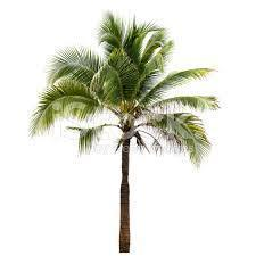} & \includegraphics[width=1.5cm, height=1.5cm]{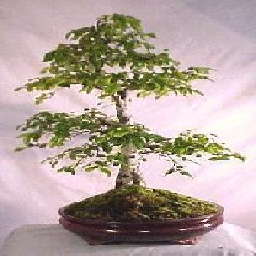} & \includegraphics[width=1.5cm, height=1.5cm]{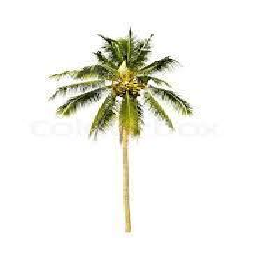} && \includegraphics[width=1.5cm, height=1.5cm]{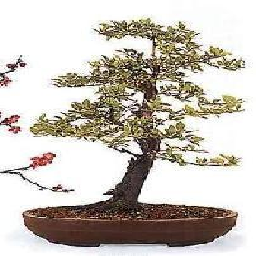} & \includegraphics[width=1.5cm, height=1.5cm]{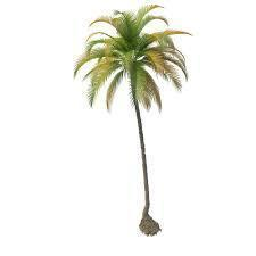} & \includegraphics[width=1.5cm, height=1.5cm]{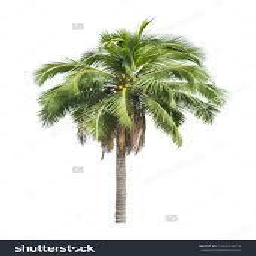} & \includegraphics[width=1.5cm, height=1.5cm]{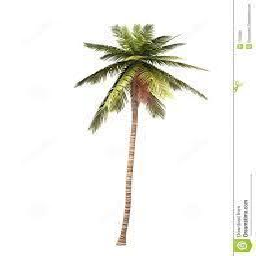} & \includegraphics[width=1.5cm, height=1.5cm]{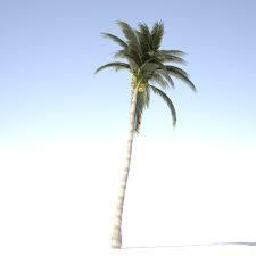} \\ 
&& \xmark & \xmark & \xmark & \xmark & \cmark && \xmark & \xmark & \cmark & \xmark & \cmark && \cmark & \cmark & \cmark & \cmark & \cmark && \cmark & \cmark & \cmark & \cmark & \cmark \\ 
\includegraphics[width=1.5cm, height=1.5cm]{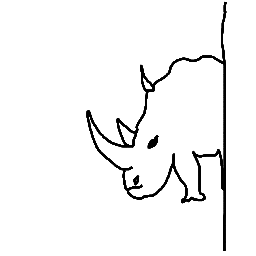} && \includegraphics[width=1.5cm, height=1.5cm]{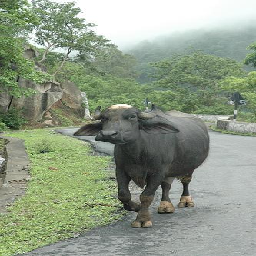} & \includegraphics[width=1.5cm, height=1.5cm]{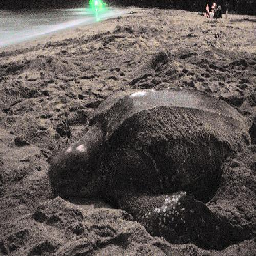} & \includegraphics[width=1.5cm, height=1.5cm]{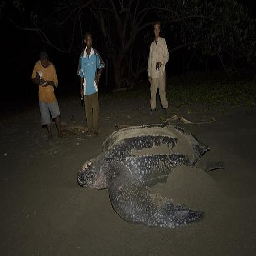} & \includegraphics[width=1.5cm, height=1.5cm]{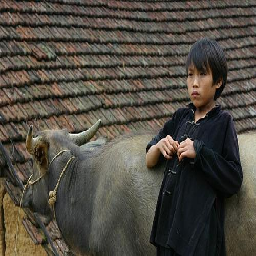} & \includegraphics[width=1.5cm, height=1.5cm]{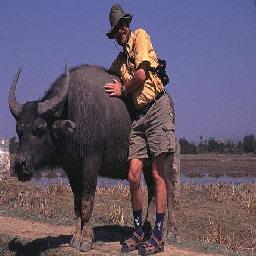} && \includegraphics[width=1.5cm, height=1.5cm]{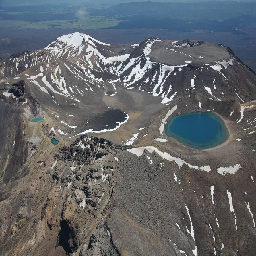} & \includegraphics[width=1.5cm, height=1.5cm]{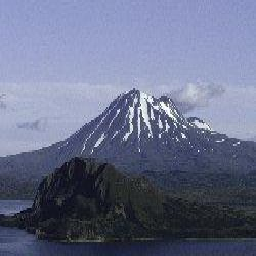} & \includegraphics[width=1.5cm, height=1.5cm]{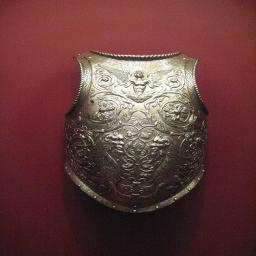} & \includegraphics[width=1.5cm, height=1.5cm]{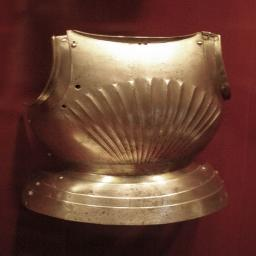} & \includegraphics[width=1.5cm, height=1.5cm]{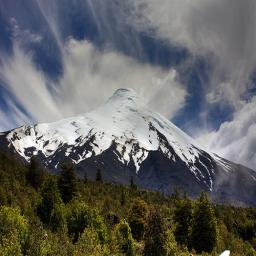} && \includegraphics[width=1.5cm, height=1.5cm]{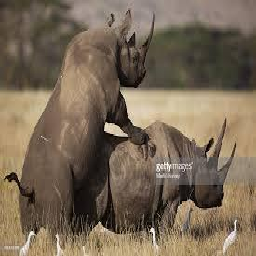} & \includegraphics[width=1.5cm, height=1.5cm]{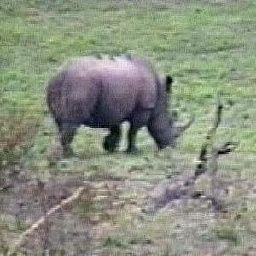} & \includegraphics[width=1.5cm, height=1.5cm]{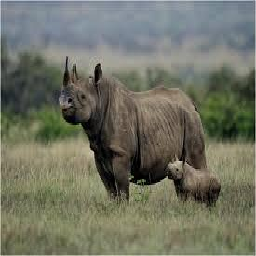} & \includegraphics[width=1.5cm, height=1.5cm]{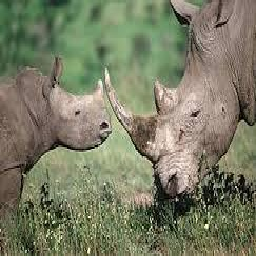} & \includegraphics[width=1.5cm, height=1.5cm]{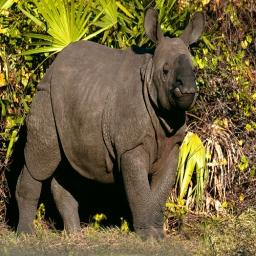} && \includegraphics[width=1.5cm, height=1.5cm]{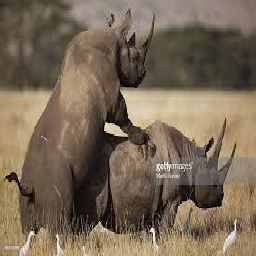} & \includegraphics[width=1.5cm, height=1.5cm]{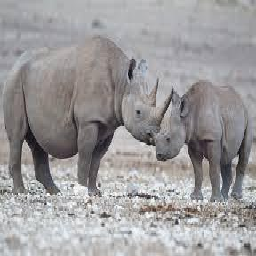} & \includegraphics[width=1.5cm, height=1.5cm]{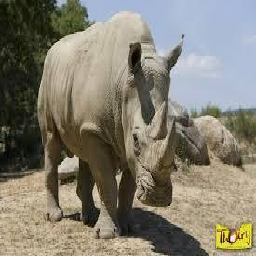} & \includegraphics[width=1.5cm, height=1.5cm]{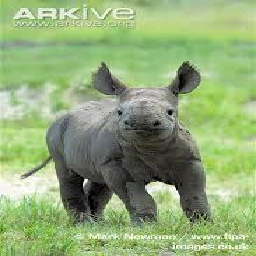} & \includegraphics[width=1.5cm, height=1.5cm]{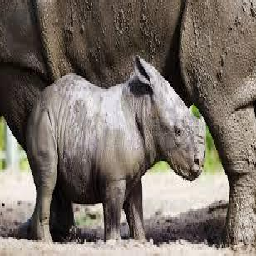} \\ 
&& \xmark & \xmark & \xmark & \xmark & \xmark && \xmark & \xmark & \xmark & \xmark & \xmark && \cmark & \cmark & \cmark & \cmark & \cmark && \cmark & \cmark & \cmark & \cmark & \cmark \\ 
\includegraphics[width=1.5cm, height=1.5cm]{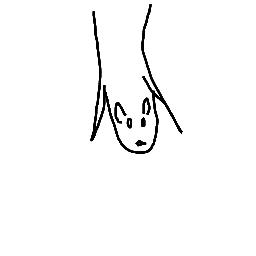} && \includegraphics[width=1.5cm, height=1.5cm]{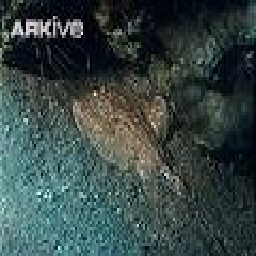} & \includegraphics[width=1.5cm, height=1.5cm]{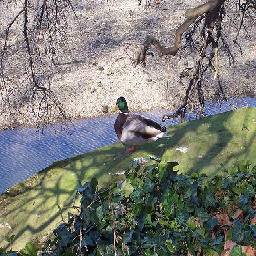} & \includegraphics[width=1.5cm, height=1.5cm]{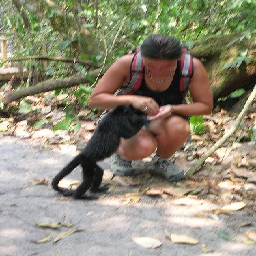} & \includegraphics[width=1.5cm, height=1.5cm]{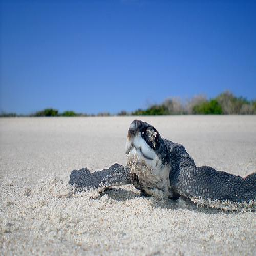} & \includegraphics[width=1.5cm, height=1.5cm]{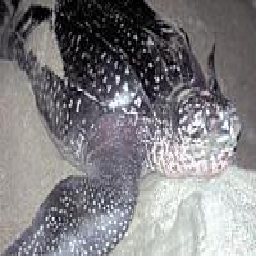} && \includegraphics[width=1.5cm, height=1.5cm]{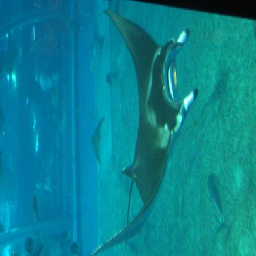} & \includegraphics[width=1.5cm, height=1.5cm]{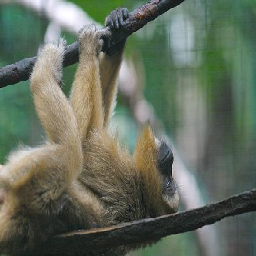} & \includegraphics[width=1.5cm, height=1.5cm]{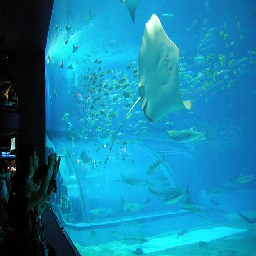} & \includegraphics[width=1.5cm, height=1.5cm]{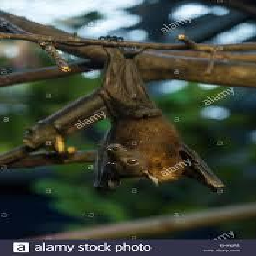} & \includegraphics[width=1.5cm, height=1.5cm]{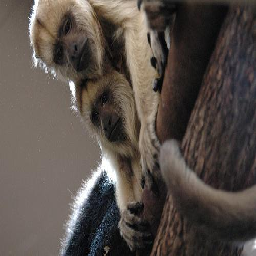} && \includegraphics[width=1.5cm, height=1.5cm]{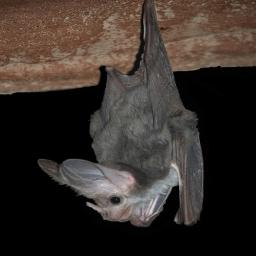} & \includegraphics[width=1.5cm, height=1.5cm]{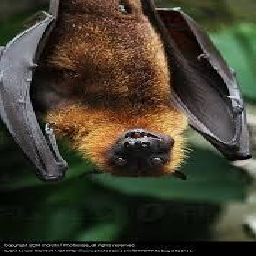} & \includegraphics[width=1.5cm, height=1.5cm]{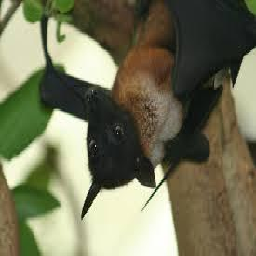} & \includegraphics[width=1.5cm, height=1.5cm]{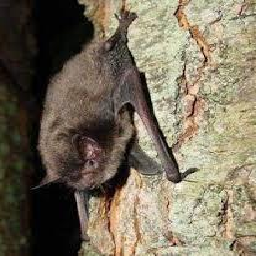} & \includegraphics[width=1.5cm, height=1.5cm]{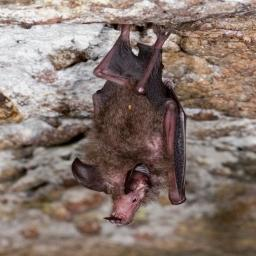} && \includegraphics[width=1.5cm, height=1.5cm]{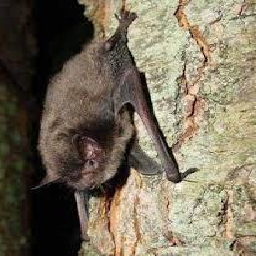} & \includegraphics[width=1.5cm, height=1.5cm]{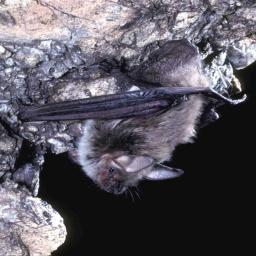} & \includegraphics[width=1.5cm, height=1.5cm]{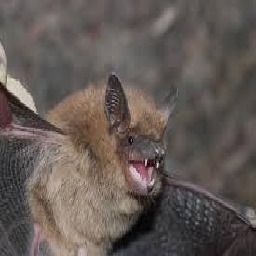} & \includegraphics[width=1.5cm, height=1.5cm]{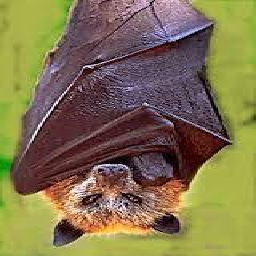} & \includegraphics[width=1.5cm, height=1.5cm]{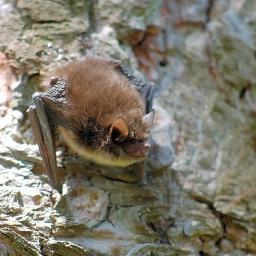} \\ 
&& \xmark & \xmark & \xmark & \xmark & \xmark && \xmark & \xmark & \xmark & \cmark & \xmark && \cmark & \cmark & \cmark & \cmark & \cmark && \cmark & \cmark & \cmark & \cmark & \cmark \\
\includegraphics[width=1.5cm, height=1.5cm]{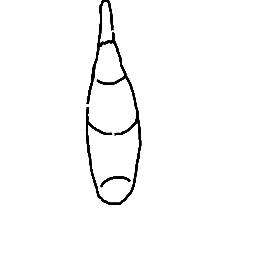} && \includegraphics[width=1.5cm, height=1.5cm]{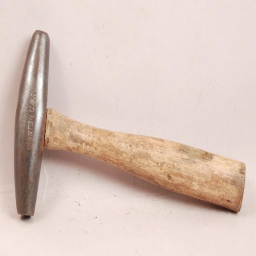} & \includegraphics[width=1.5cm, height=1.5cm]{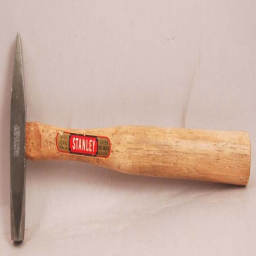} & \includegraphics[width=1.5cm, height=1.5cm]{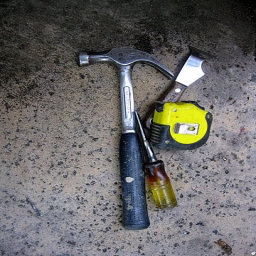} & \includegraphics[width=1.5cm, height=1.5cm]{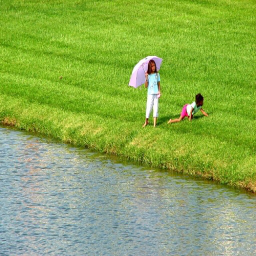} & \includegraphics[width=1.5cm, height=1.5cm]{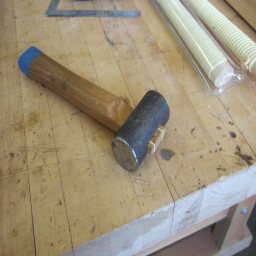} && \includegraphics[width=1.5cm, height=1.5cm]{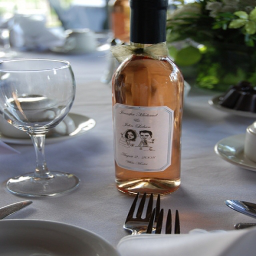} & \includegraphics[width=1.5cm, height=1.5cm]{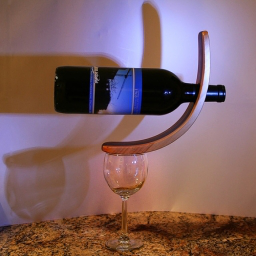} & \includegraphics[width=1.5cm, height=1.5cm]{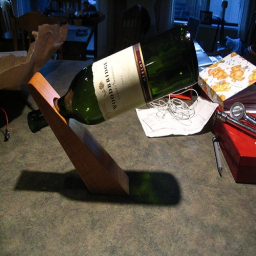} & \includegraphics[width=1.5cm, height=1.5cm]{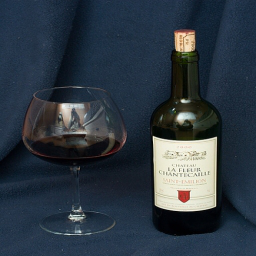} & \includegraphics[width=1.5cm, height=1.5cm]{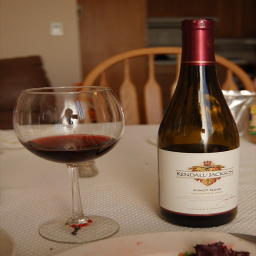} && \includegraphics[width=1.5cm, height=1.5cm]{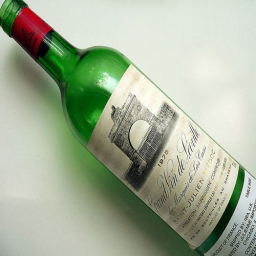} & \includegraphics[width=1.5cm, height=1.5cm]{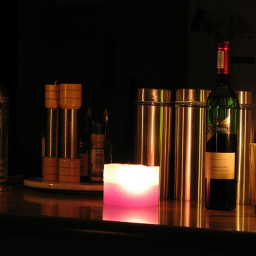} & \includegraphics[width=1.5cm, height=1.5cm]{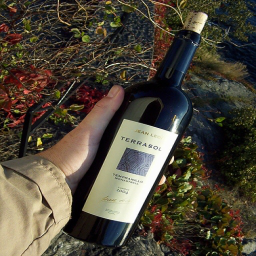} & \includegraphics[width=1.5cm, height=1.5cm]{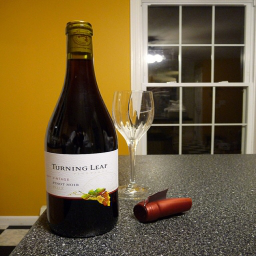} & \includegraphics[width=1.5cm, height=1.5cm]{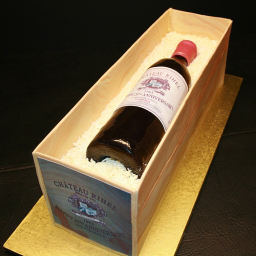} && \includegraphics[width=1.5cm, height=1.5cm]{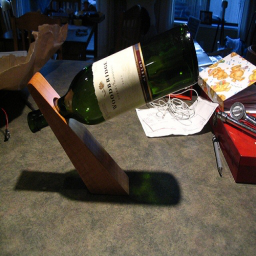} & \includegraphics[width=1.5cm, height=1.5cm]{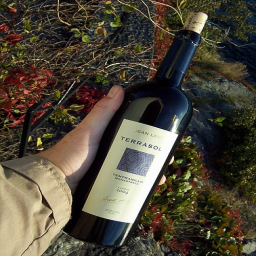} & \includegraphics[width=1.5cm, height=1.5cm]{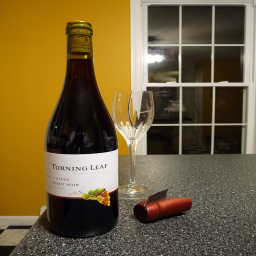} & \includegraphics[width=1.5cm, height=1.5cm]{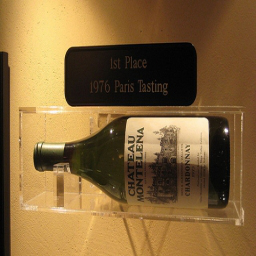} & \includegraphics[width=1.5cm, height=1.5cm]{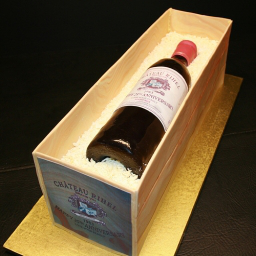} \\
&& \xmark & \xmark & \xmark & \xmark & \xmark && \cmark & \cmark & \cmark & \cmark & \cmark && \cmark & \cmark & \cmark & \cmark & \cmark && \cmark & \cmark & \cmark & \cmark & \cmark \\
\includegraphics[width=1.5cm, height=1.5cm]{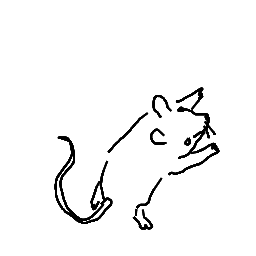} && \includegraphics[width=1.5cm, height=1.5cm]{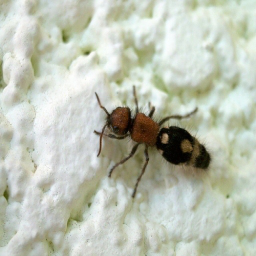} & \includegraphics[width=1.5cm, height=1.5cm]{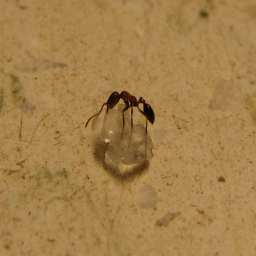} & \includegraphics[width=1.5cm, height=1.5cm]{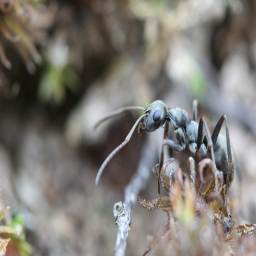} &  \includegraphics[width=1.5cm, height=1.5cm]{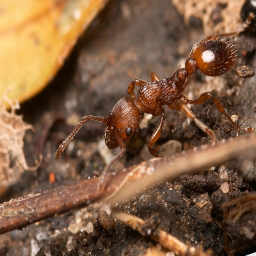} & \includegraphics[width=1.5cm, height=1.5cm]{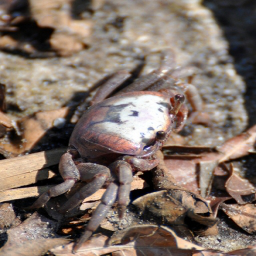} && \includegraphics[width=1.5cm, height=1.5cm]{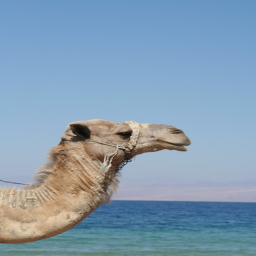} & \includegraphics[width=1.5cm, height=1.5cm]{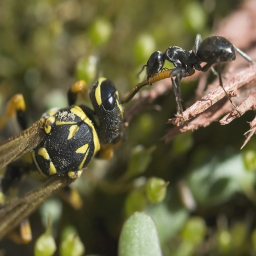} & \includegraphics[width=1.5cm, height=1.5cm]{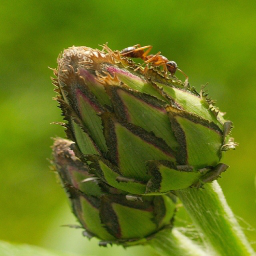} & \includegraphics[width=1.5cm, height=1.5cm]{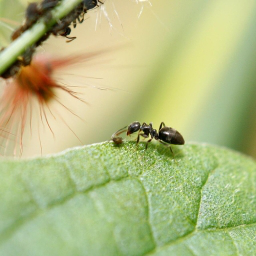} &  \includegraphics[width=1.5cm, height=1.5cm]{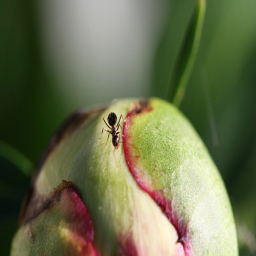} && \includegraphics[width=1.5cm, height=1.5cm]{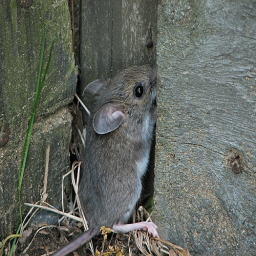} & \includegraphics[width=1.5cm, height=1.5cm]{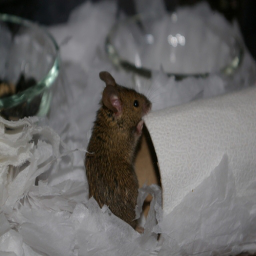} & \includegraphics[width=1.5cm, height=1.5cm]{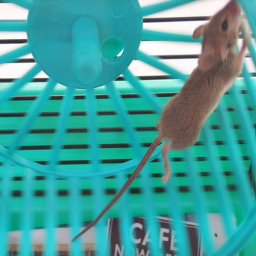} & \includegraphics[width=1.5cm, height=1.5cm]{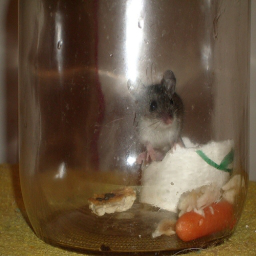} &  \includegraphics[width=1.5cm, height=1.5cm]{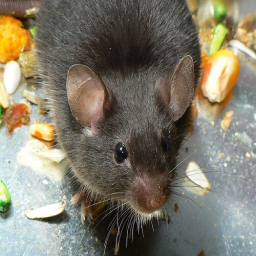} && \includegraphics[width=1.5cm, height=1.5cm]{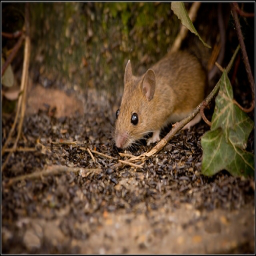} & \includegraphics[width=1.5cm, height=1.5cm]{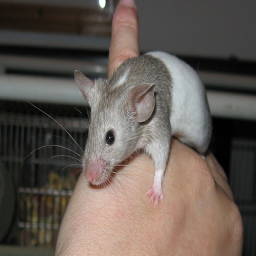} & \includegraphics[width=1.5cm, height=1.5cm]{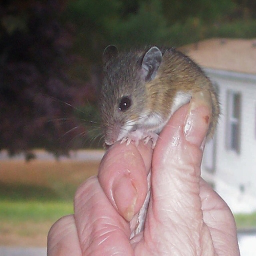} & \includegraphics[width=1.5cm, height=1.5cm]{few_shot/sketchy/40/10/4_1} & \includegraphics[width=1.5cm, height=1.5cm]{few_shot/sketchy/40/10/5_1}\\
&& \xmark & \xmark & \xmark & \xmark & \xmark && \xmark & \xmark & \xmark & \xmark & \xmark && \cmark & \cmark & \cmark & \cmark & \cmark && \cmark & \cmark & \cmark & \cmark & \cmark
\end{tabular}}
\end{center}
\caption{Top-5 $k$-shot ($k=0, 1, 5, 10$) SBIR results obtained by our SEM-PCYC model on the Sketchy (Extended) dataset are shown here according to the Euclidean distances, where the green ticks denote the correctly retrieved candidates, whereas the red crosses indicate the wrong retrievals. (best viewed in color)}
\label{fig:qual_results_fewshot_sketchy}
\end{figure*}

\begin{figure*}
\begin{center}
\resizebox{\textwidth}{!}{
\begin{tabular}{lc c@{}c@{}c@{}c@{}c@{}cc@{}c@{}c@{}c@{}c@{}cc@{}c@{}c@{}c@{}c@{}cc@{}c@{}c@{}c@{}c@{}c}
& \Huge{Query} && \multicolumn{5}{c}{\Huge 0-shot} && \multicolumn{5}{c}{\Huge 1-shot} && \multicolumn{5}{c}{\Huge 5-shot} && \multicolumn{5}{c}{\Huge 10-shot} \\
& \includegraphics[width=1.5cm, height=1.5cm]{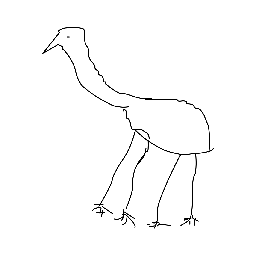} && \includegraphics[width=1.5cm, height=1.5cm]{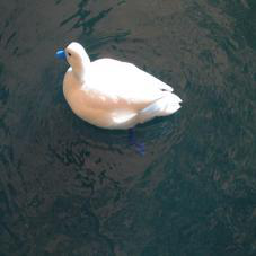} & \includegraphics[width=1.5cm, height=1.5cm]{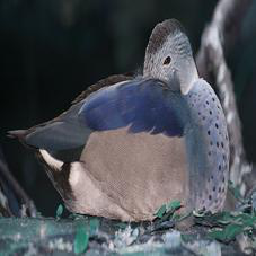} & \includegraphics[width=1.5cm, height=1.5cm]{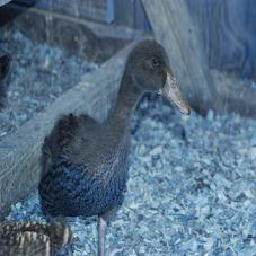} & \includegraphics[width=1.5cm, height=1.5cm]{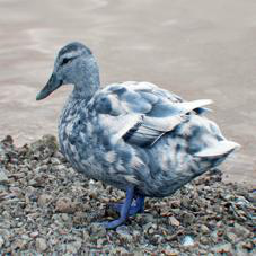} & \includegraphics[width=1.5cm, height=1.5cm]{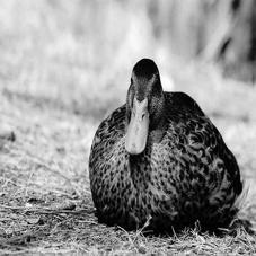} && \includegraphics[width=1.5cm, height=1.5cm]{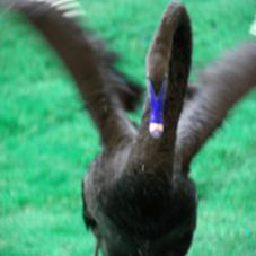} & \includegraphics[width=1.5cm, height=1.5cm]{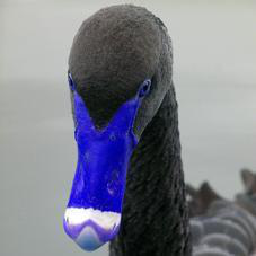} & \includegraphics[width=1.5cm, height=1.5cm]{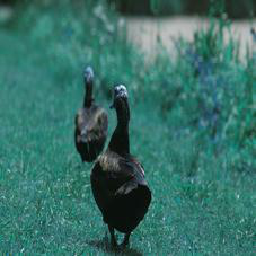} & \includegraphics[width=1.5cm, height=1.5cm]{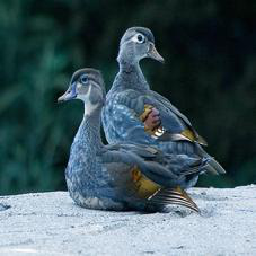} & \includegraphics[width=1.5cm, height=1.5cm]{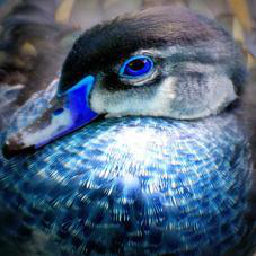} && \includegraphics[width=1.5cm, height=1.5cm]{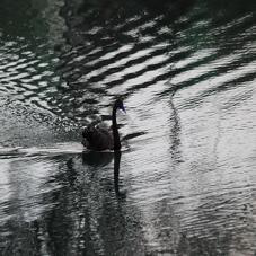} & \includegraphics[width=1.5cm, height=1.5cm]{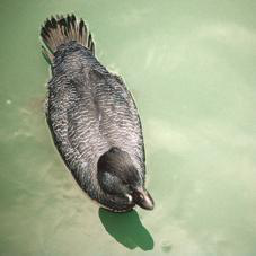} & \includegraphics[width=1.5cm, height=1.5cm]{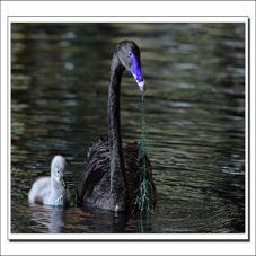} & \includegraphics[width=1.5cm, height=1.5cm]{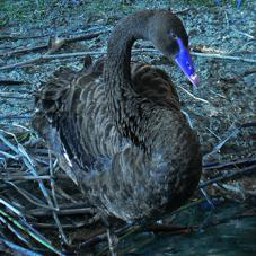} & \includegraphics[width=1.5cm, height=1.5cm]{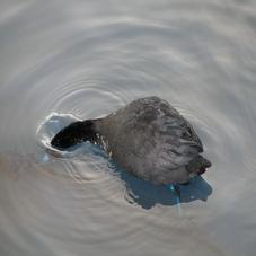} && \includegraphics[width=1.5cm, height=1.5cm]{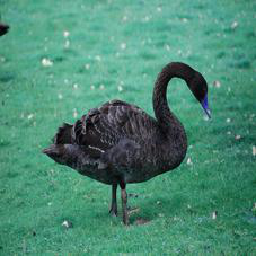} & \includegraphics[width=1.5cm, height=1.5cm]{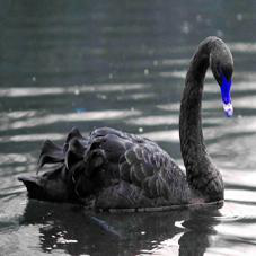} & \includegraphics[width=1.5cm, height=1.5cm]{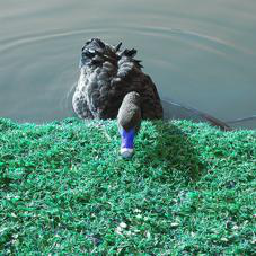} & \includegraphics[width=1.5cm, height=1.5cm]{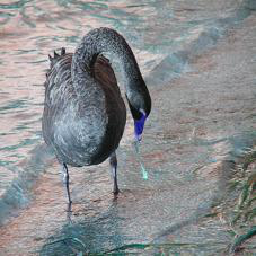} & \includegraphics[width=1.5cm, height=1.5cm]{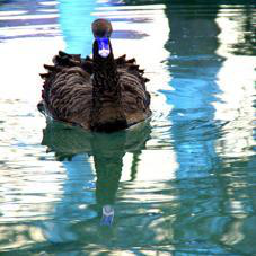} \\
&&& \xmark & \xmark & \xmark & \xmark & \xmark && \cmark & \cmark & \xmark & \xmark & \xmark && \cmark & \xmark & \cmark & \cmark & \xmark && \cmark & \cmark & \cmark & \cmark & \cmark \\
& \includegraphics[width=1.5cm, height=1.5cm]{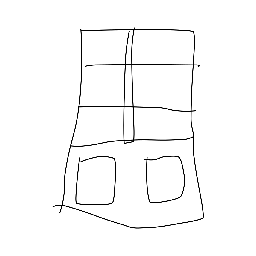} && \includegraphics[width=1.5cm, height=1.5cm]{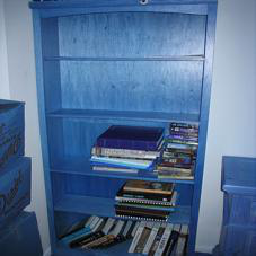} & \includegraphics[width=1.5cm, height=1.5cm]{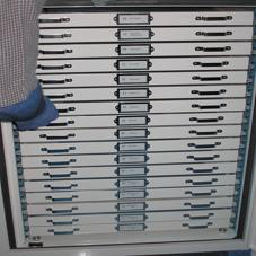} & \includegraphics[width=1.5cm, height=1.5cm]{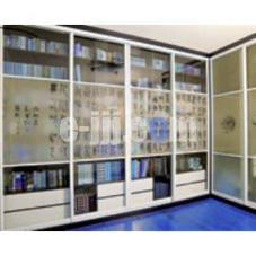} & \includegraphics[width=1.5cm, height=1.5cm]{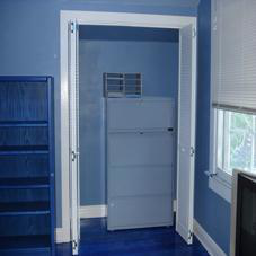} & \includegraphics[width=1.5cm, height=1.5cm]{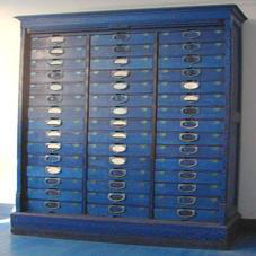} && \includegraphics[width=1.5cm, height=1.5cm]{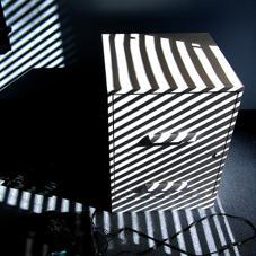} & \includegraphics[width=1.5cm, height=1.5cm]{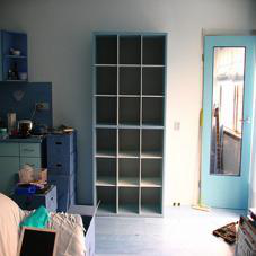} & \includegraphics[width=1.5cm, height=1.5cm]{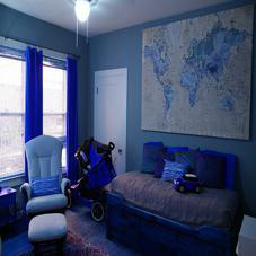} & \includegraphics[width=1.5cm, height=1.5cm]{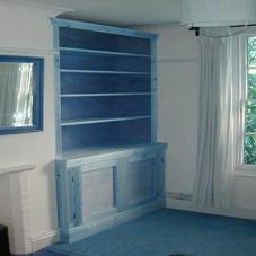} & \includegraphics[width=1.5cm, height=1.5cm]{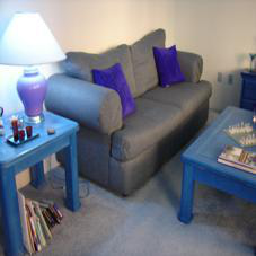} && \includegraphics[width=1.5cm, height=1.5cm]{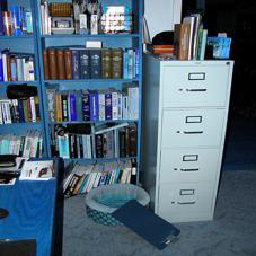} & \includegraphics[width=1.5cm, height=1.5cm]{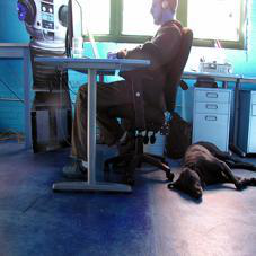} & \includegraphics[width=1.5cm, height=1.5cm]{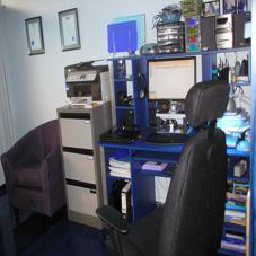} & \includegraphics[width=1.5cm, height=1.5cm]{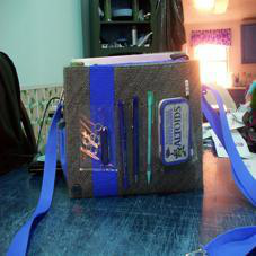} & \includegraphics[width=1.5cm, height=1.5cm]{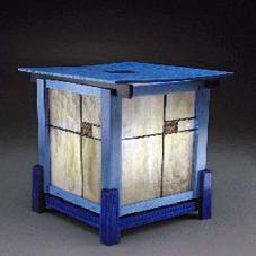} && \includegraphics[width=1.5cm, height=1.5cm]{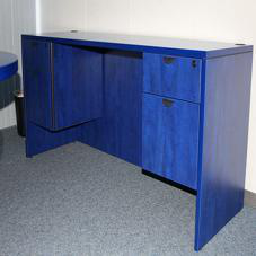} & \includegraphics[width=1.5cm, height=1.5cm]{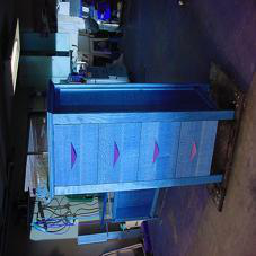} & \includegraphics[width=1.5cm, height=1.5cm]{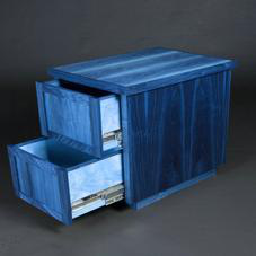} & \includegraphics[width=1.5cm, height=1.5cm]{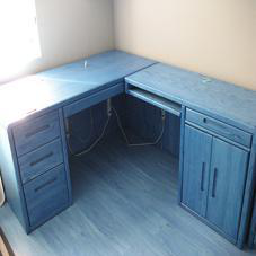} & \includegraphics[width=1.5cm, height=1.5cm]{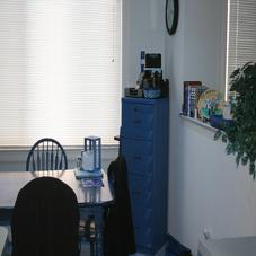} \\
&&& \xmark & \cmark & \xmark & \cmark & \cmark && \cmark & \xmark & \xmark & \xmark & \xmark && \cmark & \cmark & \cmark & \xmark & \xmark && \cmark & \cmark & \cmark & \cmark & \cmark \\
& \includegraphics[width=1.5cm, height=1.5cm]{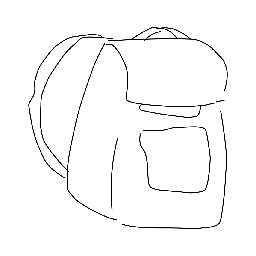} && \includegraphics[width=1.5cm, height=1.5cm]{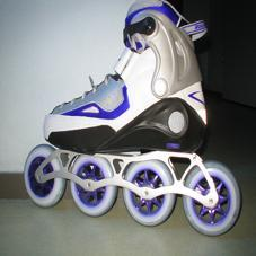} & \includegraphics[width=1.5cm, height=1.5cm]{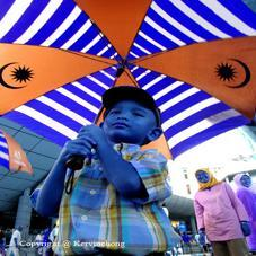} & \includegraphics[width=1.5cm, height=1.5cm]{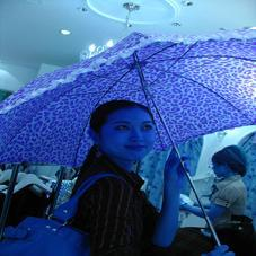} & \includegraphics[width=1.5cm, height=1.5cm]{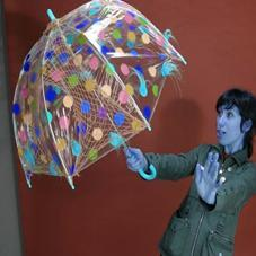} & \includegraphics[width=1.5cm, height=1.5cm]{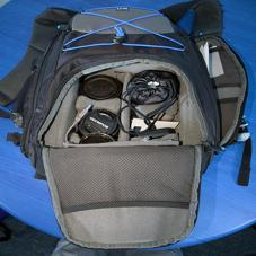} && \includegraphics[width=1.5cm, height=1.5cm]{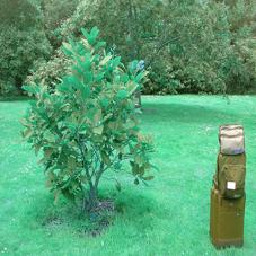} & \includegraphics[width=1.5cm, height=1.5cm]{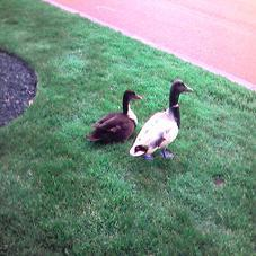} & \includegraphics[width=1.5cm, height=1.5cm]{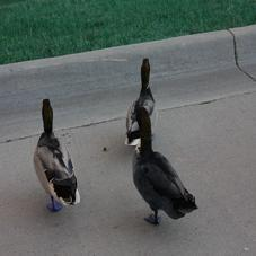} & \includegraphics[width=1.5cm, height=1.5cm]{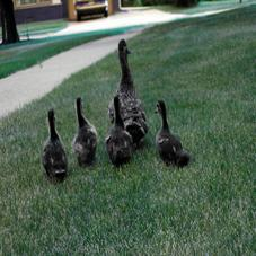} & \includegraphics[width=1.5cm, height=1.5cm]{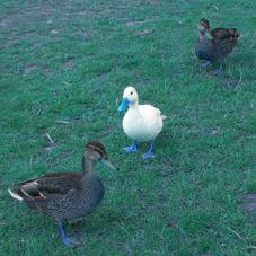} && \includegraphics[width=1.5cm, height=1.5cm]{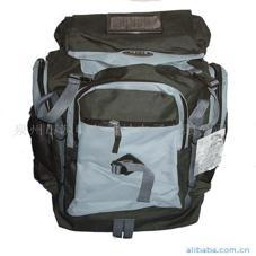} & \includegraphics[width=1.5cm, height=1.5cm]{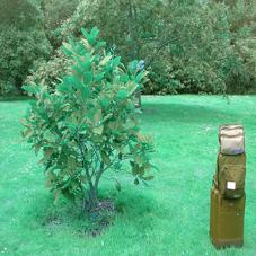} & \includegraphics[width=1.5cm, height=1.5cm]{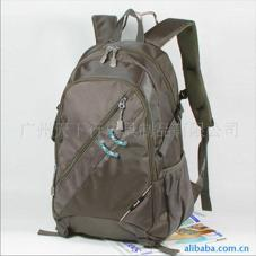} & \includegraphics[width=1.5cm, height=1.5cm]{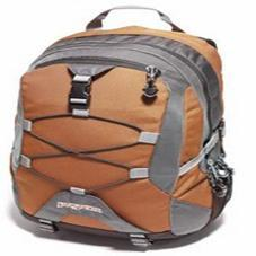} & \includegraphics[width=1.5cm, height=1.5cm]{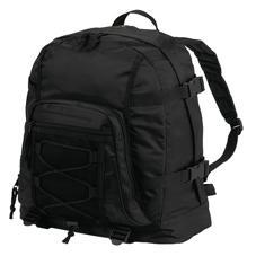} && \includegraphics[width=1.5cm, height=1.5cm]{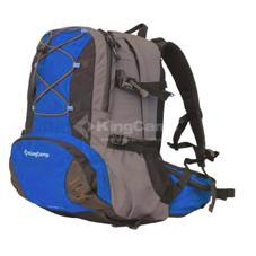} & \includegraphics[width=1.5cm, height=1.5cm]{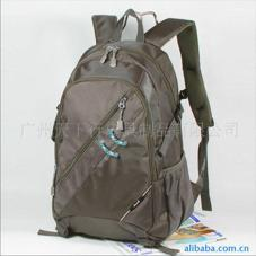} & \includegraphics[width=1.5cm, height=1.5cm]{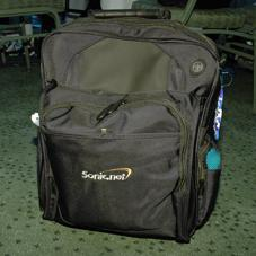} & \includegraphics[width=1.5cm, height=1.5cm]{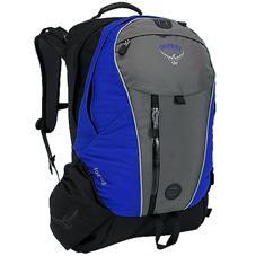} & \includegraphics[width=1.5cm, height=1.5cm]{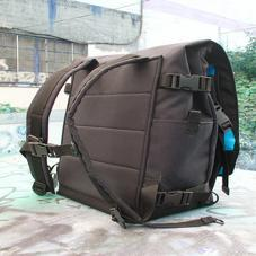} \\ 
&&& \xmark & \xmark & \xmark & \xmark & \cmark && \cmark & \xmark & \xmark & \xmark & \xmark && \cmark & \cmark & \cmark & \cmark & \cmark && \cmark & \cmark & \cmark & \cmark & \cmark \\
& \includegraphics[width=1.5cm, height=1.5cm]{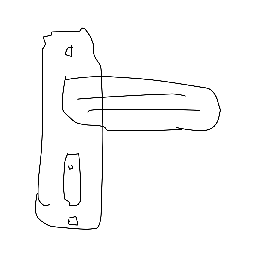} && \includegraphics[width=1.5cm, height=1.5cm]{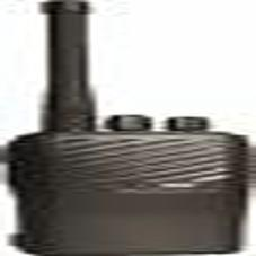} & \includegraphics[width=1.5cm, height=1.5cm]{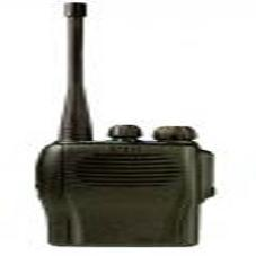} & \includegraphics[width=1.5cm, height=1.5cm]{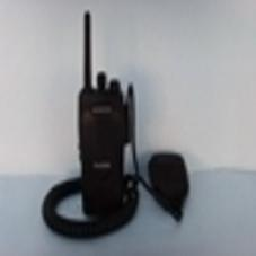} & \includegraphics[width=1.5cm, height=1.5cm]{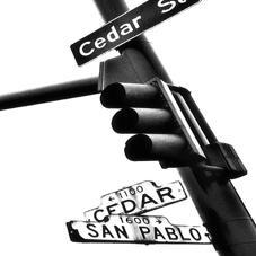} & \includegraphics[width=1.5cm, height=1.5cm]{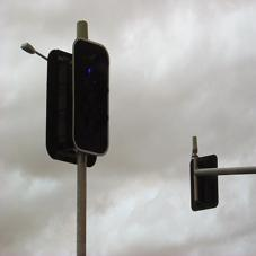} && \includegraphics[width=1.5cm, height=1.5cm]{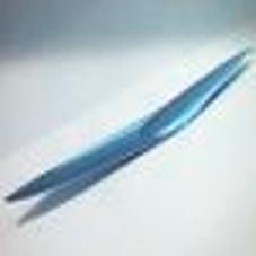} & \includegraphics[width=1.5cm, height=1.5cm]{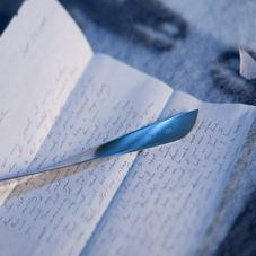} & \includegraphics[width=1.5cm, height=1.5cm]{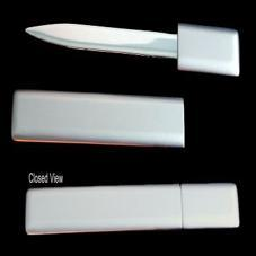} & \includegraphics[width=1.5cm, height=1.5cm]{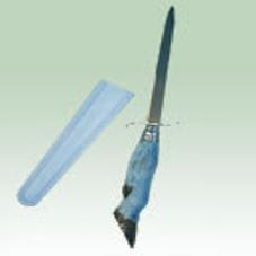} & \includegraphics[width=1.5cm, height=1.5cm]{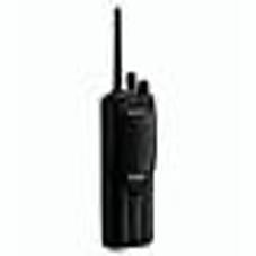} && \includegraphics[width=1.5cm, height=1.5cm]{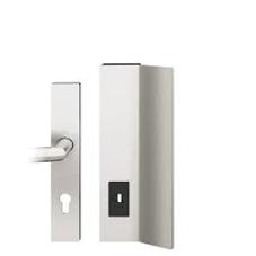} & \includegraphics[width=1.5cm, height=1.5cm]{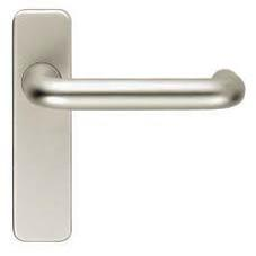} & \includegraphics[width=1.5cm, height=1.5cm]{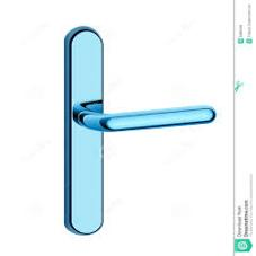} & \includegraphics[width=1.5cm, height=1.5cm]{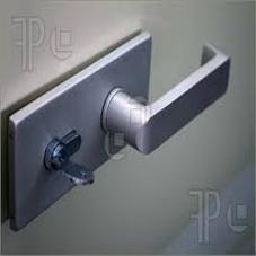} & \includegraphics[width=1.5cm, height=1.5cm]{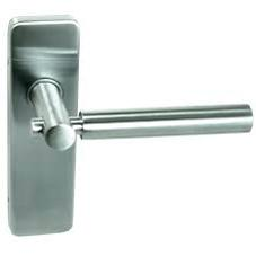} && \includegraphics[width=1.5cm, height=1.5cm]{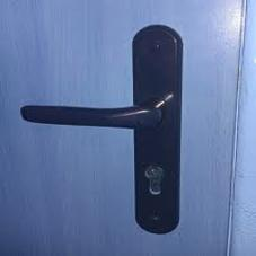} & \includegraphics[width=1.5cm, height=1.5cm]{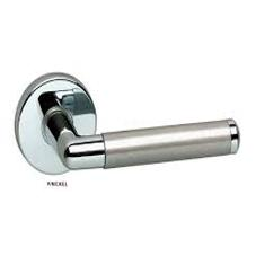} & \includegraphics[width=1.5cm, height=1.5cm]{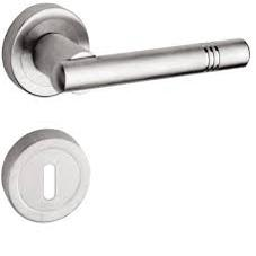} & \includegraphics[width=1.5cm, height=1.5cm]{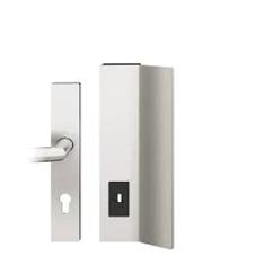} & \includegraphics[width=1.5cm, height=1.5cm]{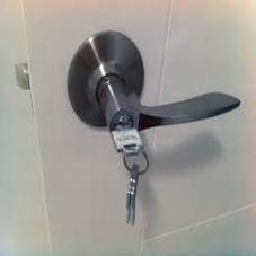} \\ 
&&& \xmark & \xmark & \xmark & \xmark & \xmark && \xmark & \xmark & \xmark & \xmark & \xmark && \cmark & \cmark & \cmark & \cmark & \cmark && \cmark & \cmark & \cmark & \cmark & \cmark \\
& \includegraphics[width=1.5cm, height=1.5cm]{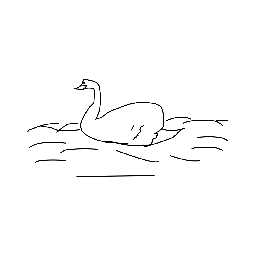} && \includegraphics[width=1.5cm, height=1.5cm]{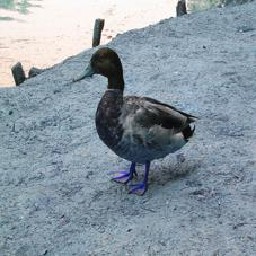} & \includegraphics[width=1.5cm, height=1.5cm]{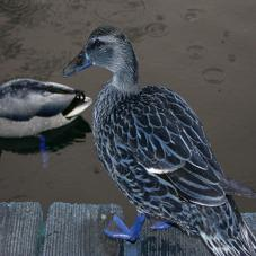} & \includegraphics[width=1.5cm, height=1.5cm]{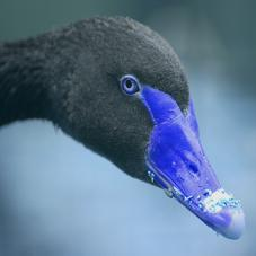} & \includegraphics[width=1.5cm, height=1.5cm]{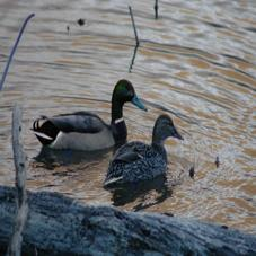} & \includegraphics[width=1.5cm, height=1.5cm]{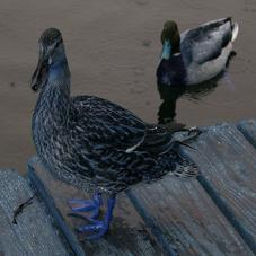} && \includegraphics[width=1.5cm, height=1.5cm]{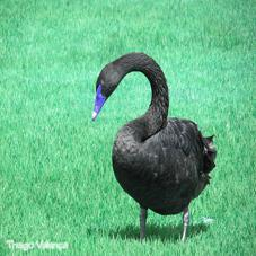} & \includegraphics[width=1.5cm, height=1.5cm]{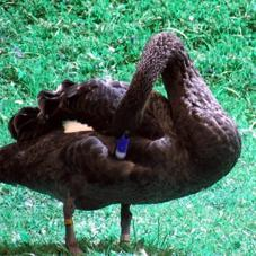} & \includegraphics[width=1.5cm, height=1.5cm]{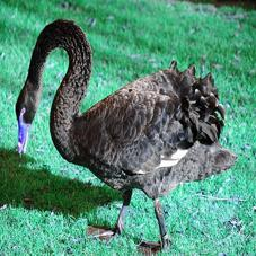} & \includegraphics[width=1.5cm, height=1.5cm]{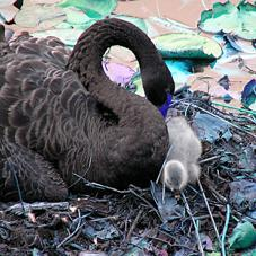} & \includegraphics[width=1.5cm, height=1.5cm]{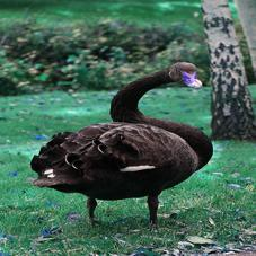} && \includegraphics[width=1.5cm, height=1.5cm]{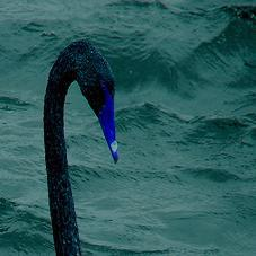} & \includegraphics[width=1.5cm, height=1.5cm]{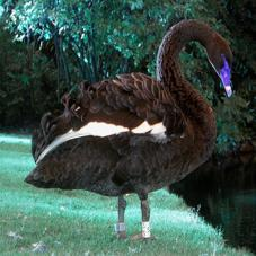} & \includegraphics[width=1.5cm, height=1.5cm]{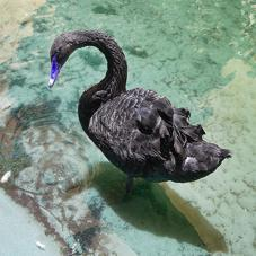} & \includegraphics[width=1.5cm, height=1.5cm]{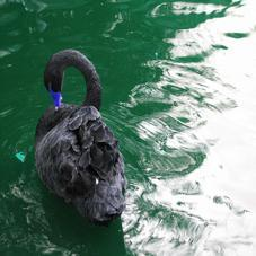} & \includegraphics[width=1.5cm, height=1.5cm]{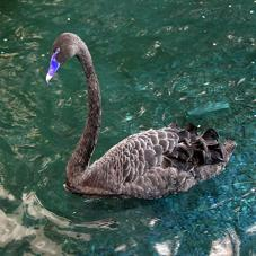} && \includegraphics[width=1.5cm, height=1.5cm]{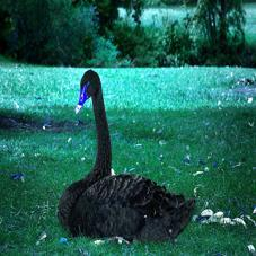} & \includegraphics[width=1.5cm, height=1.5cm]{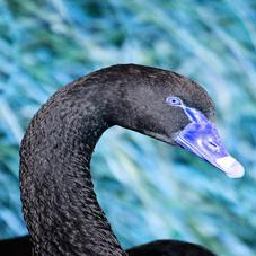} & \includegraphics[width=1.5cm, height=1.5cm]{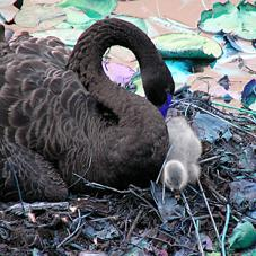} & \includegraphics[width=1.5cm, height=1.5cm]{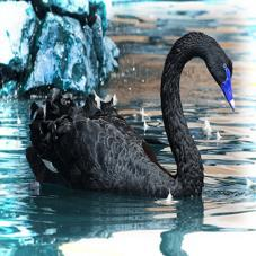} & \includegraphics[width=1.5cm, height=1.5cm]{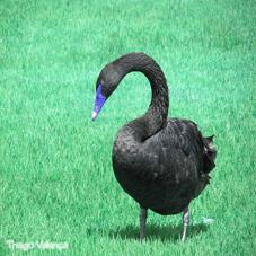}\\
&&& \xmark & \xmark & \cmark & \xmark & \xmark && \cmark & \cmark & \cmark & \cmark & \cmark && \cmark & \cmark & \cmark & \cmark & \cmark && \cmark & \cmark & \cmark & \cmark & \cmark
\end{tabular}}
\end{center}
\caption{Top-5 $k$-shot ($k=0, 1, 5, 10$) SBIR results obtained by our SEM-PCYC model on the TU-Berlin (Extended) dataset are shown here according to the Euclidean distances, where the green ticks denote the correctly retrieved candidates, whereas the red crosses indicate the wrong retrievals. (best viewed in color)}
\label{fig:qual_results_fewshot_tu-berlin}
\end{figure*}

\begin{figure*}
\begin{center}
\resizebox{\textwidth}{!}{
\begin{tabular}{lc c@{}c@{}c@{}c@{}c@{}cc@{}c@{}c@{}c@{}c@{}cc@{}c@{}c@{}c@{}c@{}cc@{}c@{}c@{}c@{}c@{}c}
& \Huge{Query} && \multicolumn{5}{c}{\Huge 0-shot} && \multicolumn{5}{c}{\Huge 1-shot} && \multicolumn{5}{c}{\Huge 5-shot} && \multicolumn{5}{c}{\Huge 10-shot} \\
& \includegraphics[width=1.5cm, height=1.5cm]{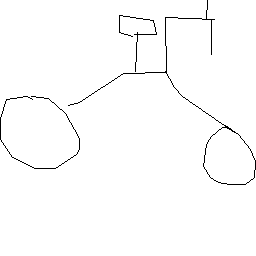} && \includegraphics[width=1.5cm, height=1.5cm]{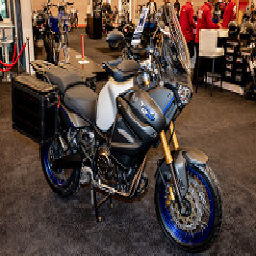} & \includegraphics[width=1.5cm, height=1.5cm]{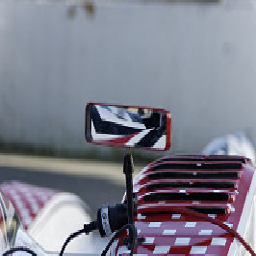} & \includegraphics[width=1.5cm, height=1.5cm]{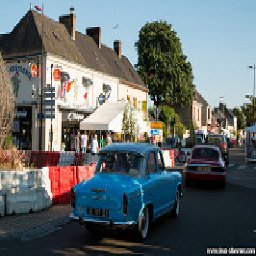} & \includegraphics[width=1.5cm, height=1.5cm]{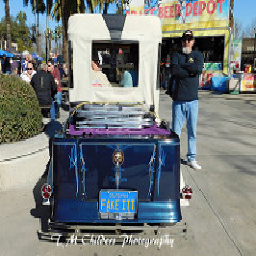} & \includegraphics[width=1.5cm, height=1.5cm]{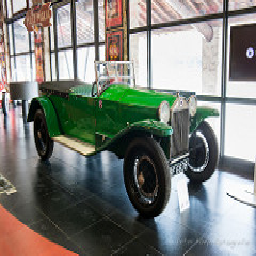} && \includegraphics[width=1.5cm, height=1.5cm]{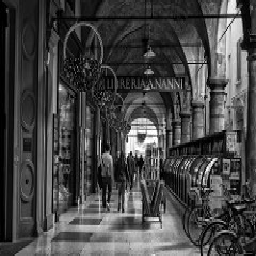} & \includegraphics[width=1.5cm, height=1.5cm]{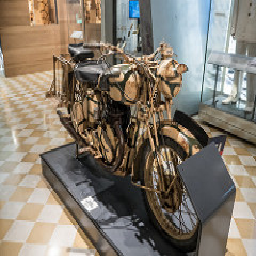} & \includegraphics[width=1.5cm, height=1.5cm]{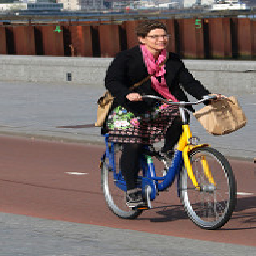} & \includegraphics[width=1.5cm, height=1.5cm]{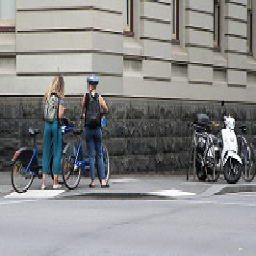} & \includegraphics[width=1.5cm, height=1.5cm]{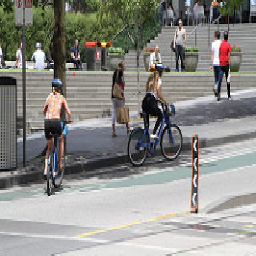} && \includegraphics[width=1.5cm, height=1.5cm]{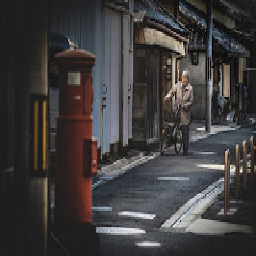} & \includegraphics[width=1.5cm, height=1.5cm]{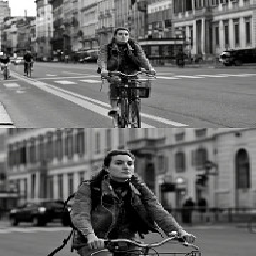} & \includegraphics[width=1.5cm, height=1.5cm]{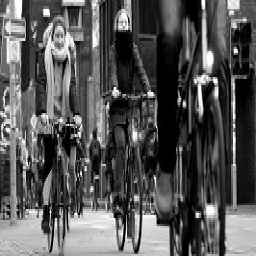} & \includegraphics[width=1.5cm, height=1.5cm]{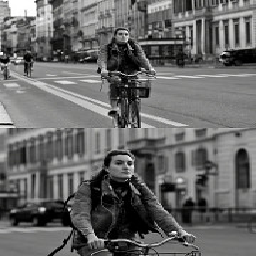} & \includegraphics[width=1.5cm, height=1.5cm]{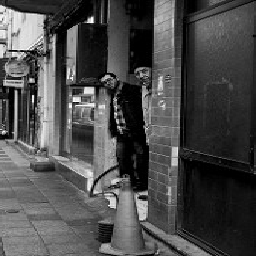} && \includegraphics[width=1.5cm, height=1.5cm]{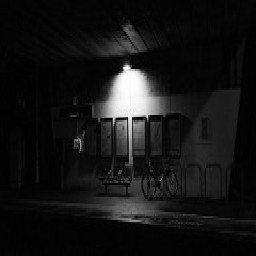} & \includegraphics[width=1.5cm, height=1.5cm]{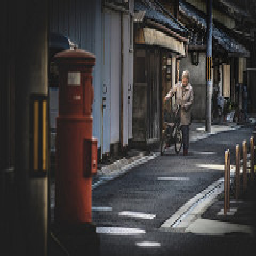} & \includegraphics[width=1.5cm, height=1.5cm]{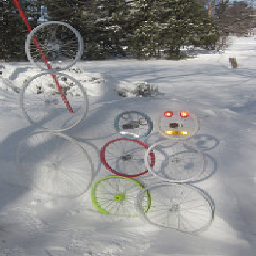} & \includegraphics[width=1.5cm, height=1.5cm]{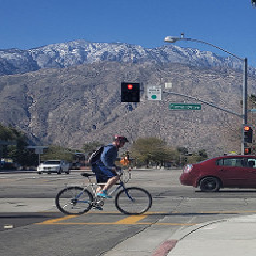} & \includegraphics[width=1.5cm, height=1.5cm]{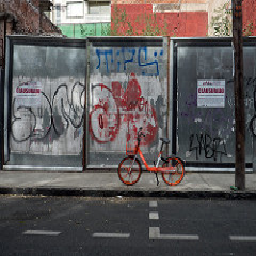} \\
&&& \xmark & \xmark & \xmark & \xmark & \xmark && \cmark & \xmark & \cmark & \cmark & \cmark && \cmark & \cmark & \cmark & \cmark & \xmark && \cmark & \cmark & \cmark & \cmark & \cmark \\
& \includegraphics[width=1.5cm, height=1.5cm]{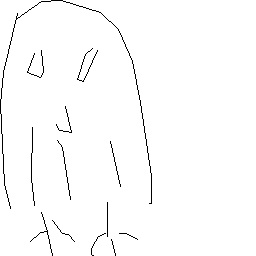} && \includegraphics[width=1.5cm, height=1.5cm]{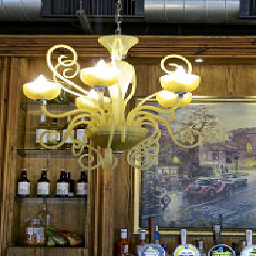} & \includegraphics[width=1.5cm, height=1.5cm]{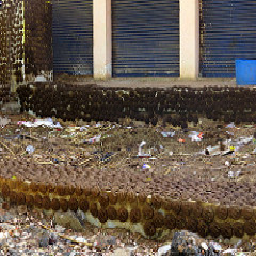} & \includegraphics[width=1.5cm, height=1.5cm]{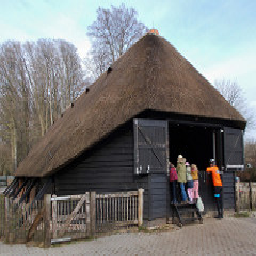} & \includegraphics[width=1.5cm, height=1.5cm]{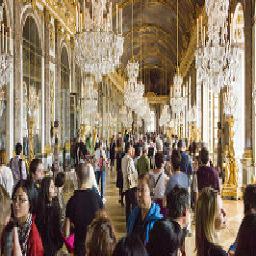} & \includegraphics[width=1.5cm, height=1.5cm]{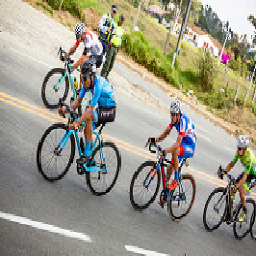} && \includegraphics[width=1.5cm, height=1.5cm]{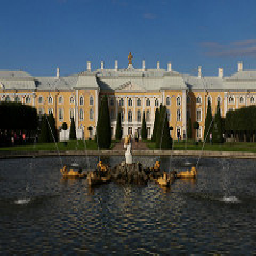} & \includegraphics[width=1.5cm, height=1.5cm]{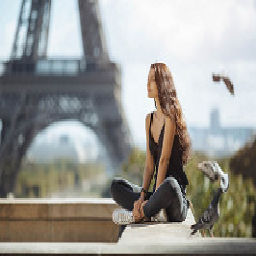} & \includegraphics[width=1.5cm, height=1.5cm]{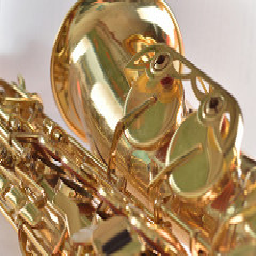} & \includegraphics[width=1.5cm, height=1.5cm]{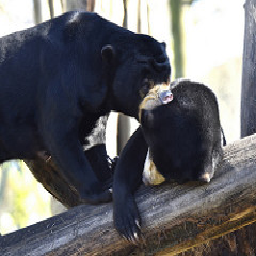} & \includegraphics[width=1.5cm, height=1.5cm]{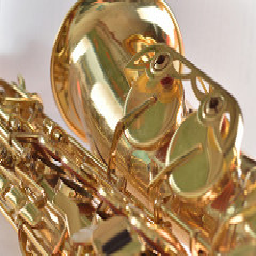} && \includegraphics[width=1.5cm, height=1.5cm]{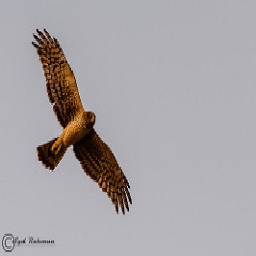} & \includegraphics[width=1.5cm, height=1.5cm]{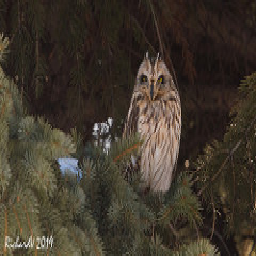} & \includegraphics[width=1.5cm, height=1.5cm]{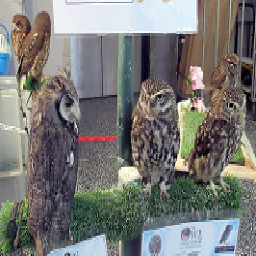} & \includegraphics[width=1.5cm, height=1.5cm]{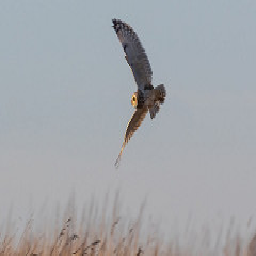} & \includegraphics[width=1.5cm, height=1.5cm]{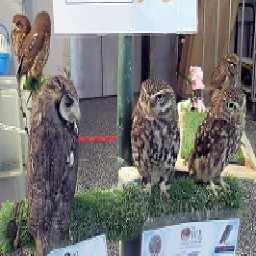} && \includegraphics[width=1.5cm, height=1.5cm]{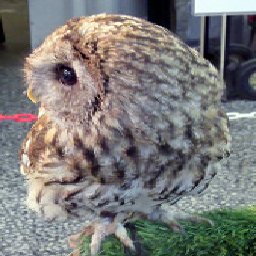} & \includegraphics[width=1.5cm, height=1.5cm]{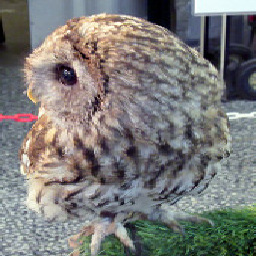} & \includegraphics[width=1.5cm, height=1.5cm]{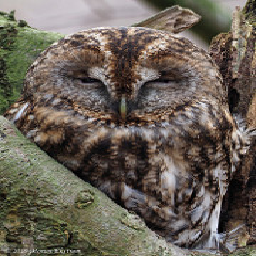} & \includegraphics[width=1.5cm, height=1.5cm]{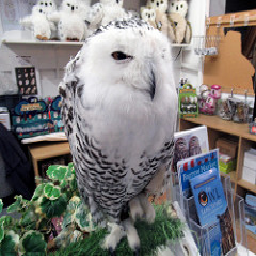} & \includegraphics[width=1.5cm, height=1.5cm]{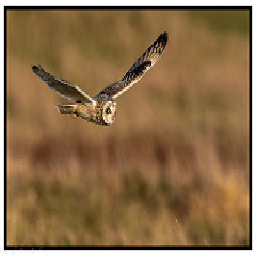} \\
&&& \xmark & \xmark & \xmark & \xmark & \xmark && \xmark & \xmark & \xmark & \xmark & \xmark && \cmark & \cmark & \cmark & \cmark & \cmark && \cmark & \cmark & \cmark & \cmark & \cmark \\
& \includegraphics[width=1.5cm, height=1.5cm]{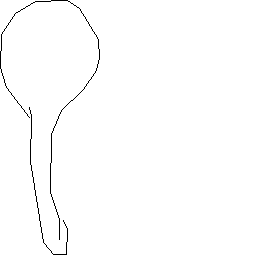} && \includegraphics[width=1.5cm, height=1.5cm]{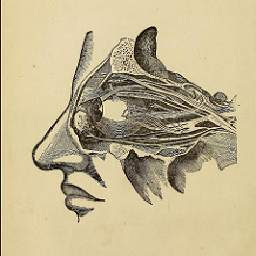} & \includegraphics[width=1.5cm, height=1.5cm]{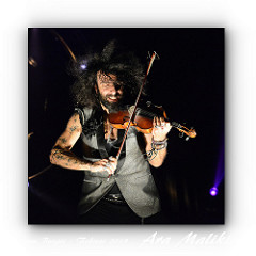} & \includegraphics[width=1.5cm, height=1.5cm]{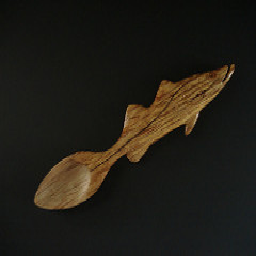} & \includegraphics[width=1.5cm, height=1.5cm]{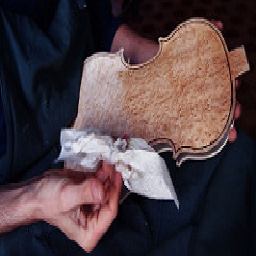} & \includegraphics[width=1.5cm, height=1.5cm]{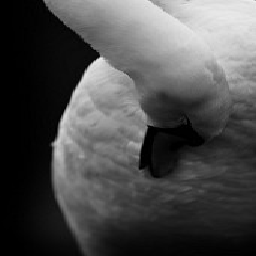} && \includegraphics[width=1.5cm, height=1.5cm]{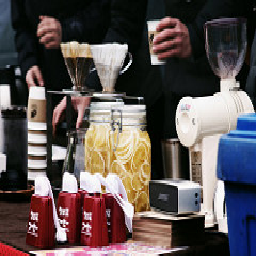} & \includegraphics[width=1.5cm, height=1.5cm]{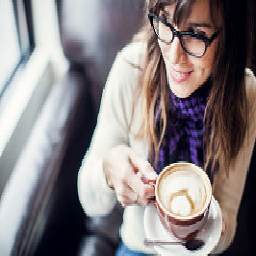} & \includegraphics[width=1.5cm, height=1.5cm]{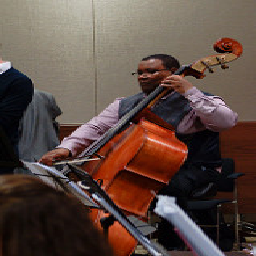} & \includegraphics[width=1.5cm, height=1.5cm]{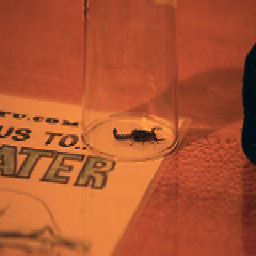} & \includegraphics[width=1.5cm, height=1.5cm]{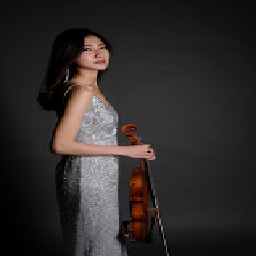} && \includegraphics[width=1.5cm, height=1.5cm]{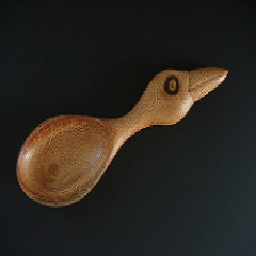} & \includegraphics[width=1.5cm, height=1.5cm]{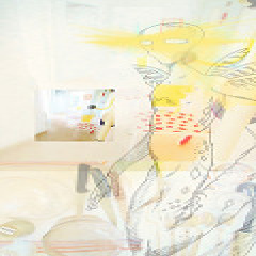} & \includegraphics[width=1.5cm, height=1.5cm]{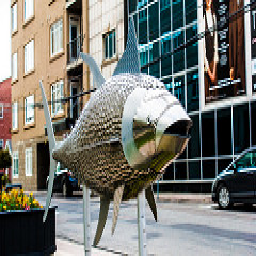} & \includegraphics[width=1.5cm, height=1.5cm]{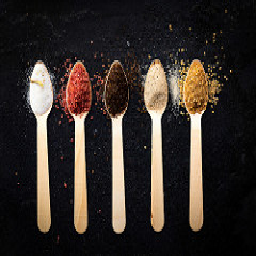} & \includegraphics[width=1.5cm, height=1.5cm]{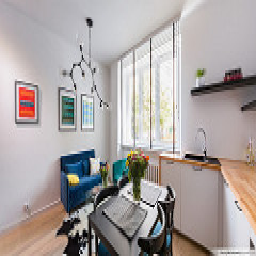} && \includegraphics[width=1.5cm, height=1.5cm]{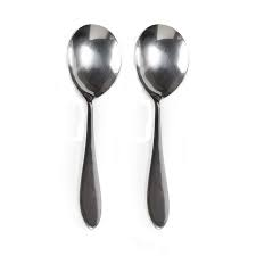} & \includegraphics[width=1.5cm, height=1.5cm]{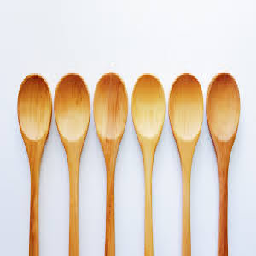} & \includegraphics[width=1.5cm, height=1.5cm]{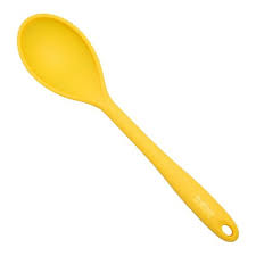} & \includegraphics[width=1.5cm, height=1.5cm]{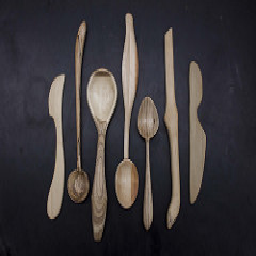} & \includegraphics[width=1.5cm, height=1.5cm]{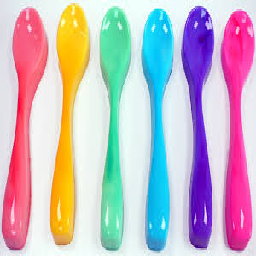} \\ 
&&& \xmark & \xmark & \cmark & \xmark & \xmark && \xmark & \xmark & \xmark & \xmark & \xmark && \cmark & \xmark & \xmark & \cmark & \xmark && \cmark & \cmark & \cmark & \cmark & \cmark \\ 
& \includegraphics[width=1.5cm, height=1.5cm]{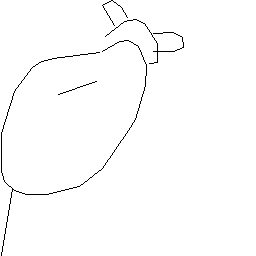} && \includegraphics[width=1.5cm, height=1.5cm]{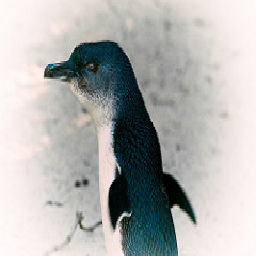} & \includegraphics[width=1.5cm, height=1.5cm]{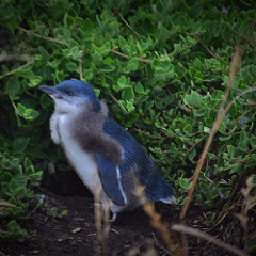} & \includegraphics[width=1.5cm, height=1.5cm]{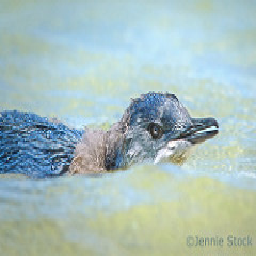} & \includegraphics[width=1.5cm, height=1.5cm]{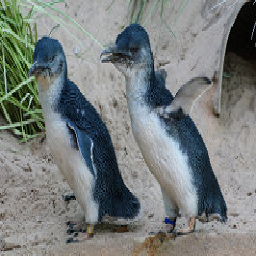} & \includegraphics[width=1.5cm, height=1.5cm]{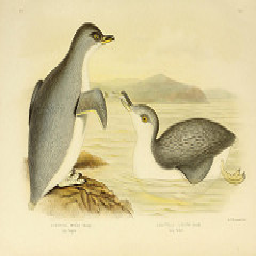} && \includegraphics[width=1.5cm, height=1.5cm]{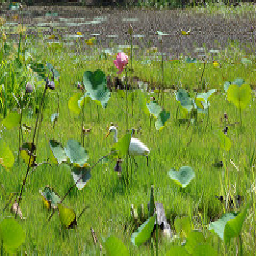} & \includegraphics[width=1.5cm, height=1.5cm]{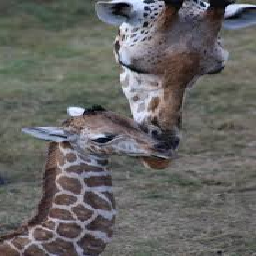} & \includegraphics[width=1.5cm, height=1.5cm]{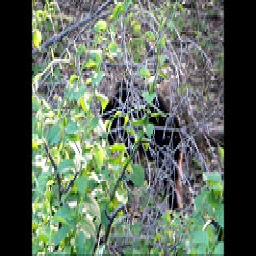} & \includegraphics[width=1.5cm, height=1.5cm]{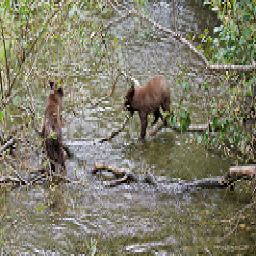} & \includegraphics[width=1.5cm, height=1.5cm]{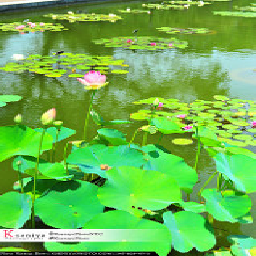} && \includegraphics[width=1.5cm, height=1.5cm]{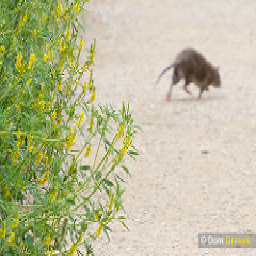} & \includegraphics[width=1.5cm, height=1.5cm]{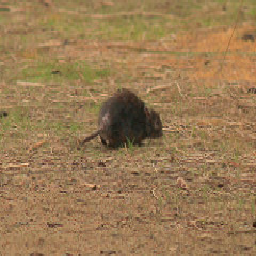} & \includegraphics[width=1.5cm, height=1.5cm]{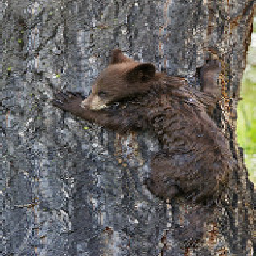} & \includegraphics[width=1.5cm, height=1.5cm]{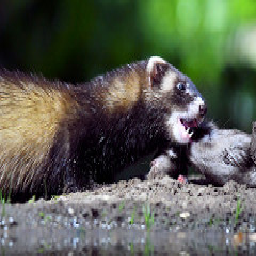} & \includegraphics[width=1.5cm, height=1.5cm]{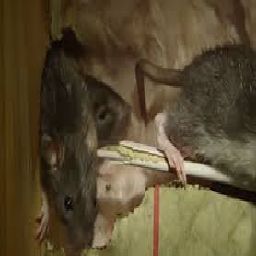} && \includegraphics[width=1.5cm, height=1.5cm]{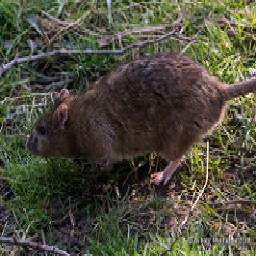} & \includegraphics[width=1.5cm, height=1.5cm]{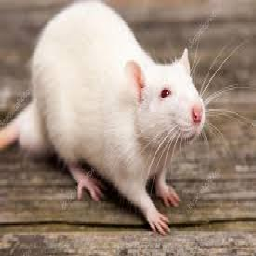} & \includegraphics[width=1.5cm, height=1.5cm]{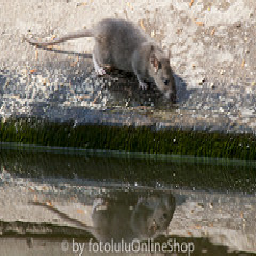} & \includegraphics[width=1.5cm, height=1.5cm]{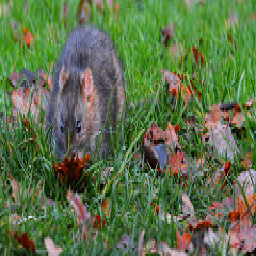} & \includegraphics[width=1.5cm, height=1.5cm]{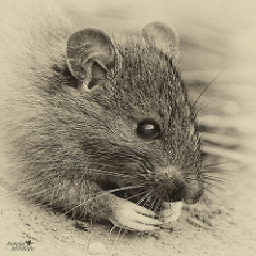} \\ 
&&& \xmark & \xmark & \xmark & \xmark & \xmark && \xmark & \xmark & \xmark & \xmark & \xmark && \cmark & \cmark & \xmark & \cmark & \cmark && \cmark & \cmark & \cmark & \cmark & \cmark \\ 
& \includegraphics[width=1.5cm, height=1.5cm]{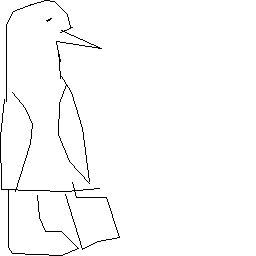} && \includegraphics[width=1.5cm, height=1.5cm]{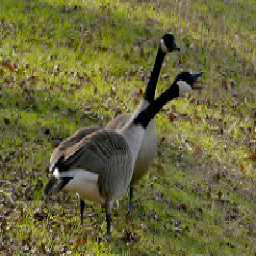} & \includegraphics[width=1.5cm, height=1.5cm]{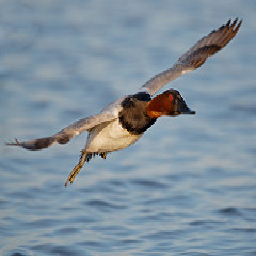} & \includegraphics[width=1.5cm, height=1.5cm]{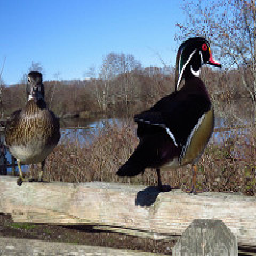} & \includegraphics[width=1.5cm, height=1.5cm]{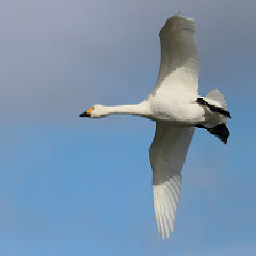} & \includegraphics[width=1.5cm, height=1.5cm]{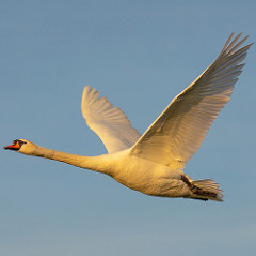} && \includegraphics[width=1.5cm, height=1.5cm]{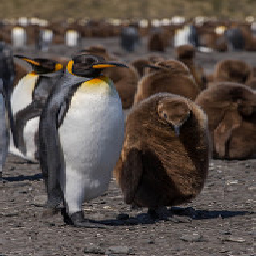} & \includegraphics[width=1.5cm, height=1.5cm]{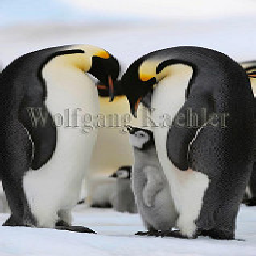} & \includegraphics[width=1.5cm, height=1.5cm]{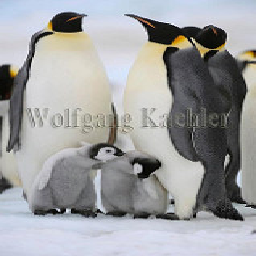} & \includegraphics[width=1.5cm, height=1.5cm]{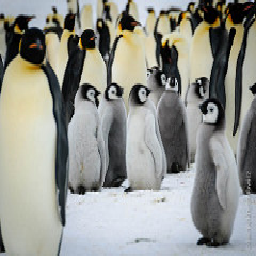} & \includegraphics[width=1.5cm, height=1.5cm]{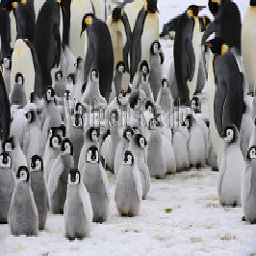} && \includegraphics[width=1.5cm, height=1.5cm]{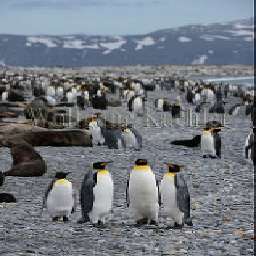} & \includegraphics[width=1.5cm, height=1.5cm]{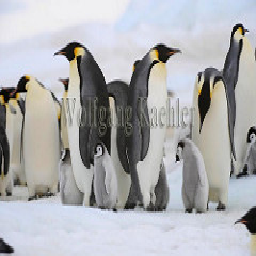} & \includegraphics[width=1.5cm, height=1.5cm]{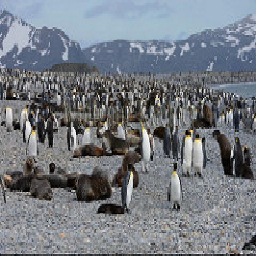} & \includegraphics[width=1.5cm, height=1.5cm]{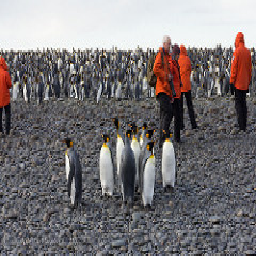} & \includegraphics[width=1.5cm, height=1.5cm]{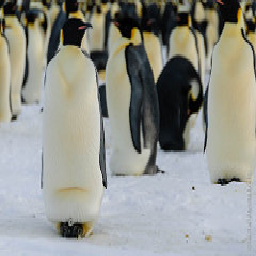} && \includegraphics[width=1.5cm, height=1.5cm]{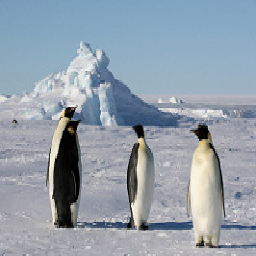} & \includegraphics[width=1.5cm, height=1.5cm]{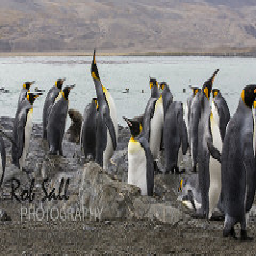} & \includegraphics[width=1.5cm, height=1.5cm]{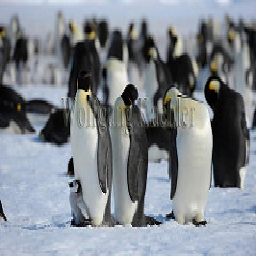} & \includegraphics[width=1.5cm, height=1.5cm]{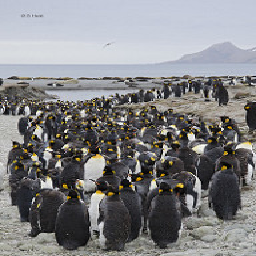} & \includegraphics[width=1.5cm, height=1.5cm]{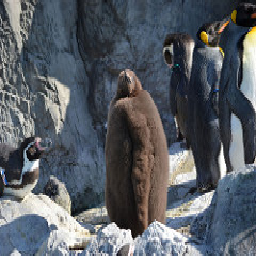} \\ 
&&& \xmark & \xmark & \xmark & \xmark & \xmark && \cmark & \cmark & \cmark & \cmark & \cmark && \cmark & \cmark & \cmark & \cmark & \cmark && \cmark & \cmark & \cmark & \cmark & \cmark
\end{tabular}}
\end{center}
\caption{Top-5 $k$-shot ($k=0, 1, 5, 10$) SBIR results obtained by our SEM-PCYC model on the QuickDraw (Extended) dataset are shown here according to the Euclidean distances, where the green ticks denote the correctly retrieved candidates, whereas the red crosses indicate the wrong retrievals. (best viewed in color)}
\label{fig:qual_results_fewshot_quickdraw}
\end{figure*}

\subsection{(Generalized) Few-Shot Sketch-based Image Retrieval}
For the few-shot scenario, we start with the pre-trained model trained in the zero-shot setting, and then fine tune it using a few example images, e.g. $k$-shot, from ``novel'' classes. For fine tuning the model in $k$-shot setting, we consider $k$ different sketch and image instances from each of the \emph{unseen} classes and cross-combine according to the coarse-grained and fine-grained settings to fine tune the model. The performance is evaluated on the rest of the instances from each class at test time. 

\myparagraph{Few-Shot Sketch-based Image Retrieval.} \fig{fig:few_shot}(a)-(c) present the few-shot SBIR performance of our SEM-PCYC model together with ZSIH~\cite{Shen2018ZSIH} and ZS-SBIR~\cite{Yelamarthi2018ZSBIR} respectively on the Sketchy, TU-Berlin and QuickDraw databases. All these plots show that the considered methods have performed consistently with the increment of $k$. However, this growth slowly gets saturated after $k=10$. In this case also our proposed SEM-PCYC model consistently outperforms the other prior works, which clearly points out the supremacy of our proposal.

\myparagraph{Generalized Few-Shot Sketch-based Image Retrieval.}
We also tested our few-shot model in generalized scenario, where during the test phase the search space includes both the \emph{seen} and \emph{novel} classes. Typically, this setting poses remarkably challenging scenario as the \emph{seen} classes may create significant confusion to the \emph{novel} queries. However, the generalized setting is more realistic as it allows to query the system with sketch from any classes. In this setting as well, we considered ZSIH~\cite{Shen2018ZSIH} and ZS-SBIR~\cite{Yelamarthi2018ZSBIR} as two benchmark methods and trained them with the same experimental settings as ours. In FS-SBIR the generalized setting results follow the non-generalized setting quite closely (see~\fig{fig:few_shot}(d)-(f)). This eventually indicates the convergence of the generalization ability of different models. In this setting as well, our proposed model steadily surpassed both the benchmark models, which indicates the advantage of our proposed model.

\myparagraph{Qualitative Results.} 
\fig{fig:qual_results_fewshot_sketchy}, \fig{fig:qual_results_fewshot_tu-berlin} and \fig{fig:qual_results_fewshot_quickdraw} present a selection of qualitative results obtained by our SEM-PCYC model respectively on the Sketchy, TU-Berlin and QuickDraw datasets in the scenario of increasing number of shots, which show an evolution of model performance with the increment of $k$ ($=0,1,5,10$) for the classes where 0-shot results are weak. From these results, we can see that sometimes a single unseen example is sufficient to correctly retrieve images (row $3$ of \fig{fig:qual_results_fewshot_sketchy}, row $5$ of \fig{fig:qual_results_fewshot_tu-berlin} and row $5$ of \fig{fig:qual_results_fewshot_quickdraw}), however, sometimes it needs more examples (row $2$ and $5$ of \fig{fig:qual_results_fewshot_sketchy}, row $2$, $3$, $4$ of \fig{fig:qual_results_fewshot_tu-berlin} and row $2$, $3$, $4$ of \fig{fig:qual_results_fewshot_quickdraw}) to remove the confusion from the other similar classes. This uncertainty may either come from visual or semantic similarity. As expected, increasing the number of examples also improves the performance.

{
\setlength{\tabcolsep}{3pt}
\renewcommand{\arraystretch}{1.2} 
\begin{table*}[!ht]
\centering
\begin{tabular}{lccc}
\hline
\textbf{Description} & \textbf{Sketchy (5-shot)} & \textbf{TU-Berlin (5-shot)} & \textbf{QuickDraw (5-shot)} \\
\hline
Only adversarial loss & $0.512$ & $0.489$ & $0.312$ \\
Adversarial + cycle consistency loss & $0.508$ & $0.499$ & $0.307$ \\
Adversarial + classification loss & $0.534$ & $0.483$ & $0.298$ \\
Adversarial (sketch + image) + cycle consistency + classification loss & $0.592$ & $0.559$ & $0.378$ \\
Without regularizer in~\eq{eqn:aenc_loss} & $0.602$ & $0.543$ & $0.365$ \\
\textbf{SEM-PCYC (full model)} & $\mathbf{0.607}$ & $\mathbf{0.566}$ & $\mathbf{0.412}$ \\
\hline
\end{tabular}
\caption{Ablation study with \emph{few shot} setting on our SEM-PCYC model ($64$d) on three datasets (measured with mAP@all).}
\label{tab:ablation_study_few_shot}
\end{table*}
}

\myparagraph{Model Ablations.} Similar to zero-shot setting, we perform an ablation study for few-shot scenario as well, where we consider the same model baselines as of \tab{tab:ablation_study_zero_shot}. The mAP@all values obtained by those baselines in $5$-shot scenario are shown in~\tab{tab:ablation_study_few_shot}. In this case, all the baselines have achieved much better performance than the corresponding zero-shot performance on that dataset, which is absolutely justified since the model is already trained to zero-shot setting and having few examples from novel classes provide some gain with any combination of losses. We observe that the first three configurations (first three rows of~\tab{tab:ablation_study_few_shot}) work quite closely across all the three datasets and we haven't found any prominent difference among these three baselines on the considered datasets. However, the baselines with more criterion or losses (bottom three rows of~\tab{tab:ablation_study_few_shot}) achieve much better performance from the previously mentioned three baselines. Among these baselines, we have not found much difference between the ones that do and do not use side information. This is due to the consideration of pre-trained zero-shot model which already has past knowledge of side information, and in this case training with side information could be slightly redundant.

\begin{figure*}[!t]
\centering
\begin{tabular}{cc}
\includegraphics[width=0.33\textwidth]{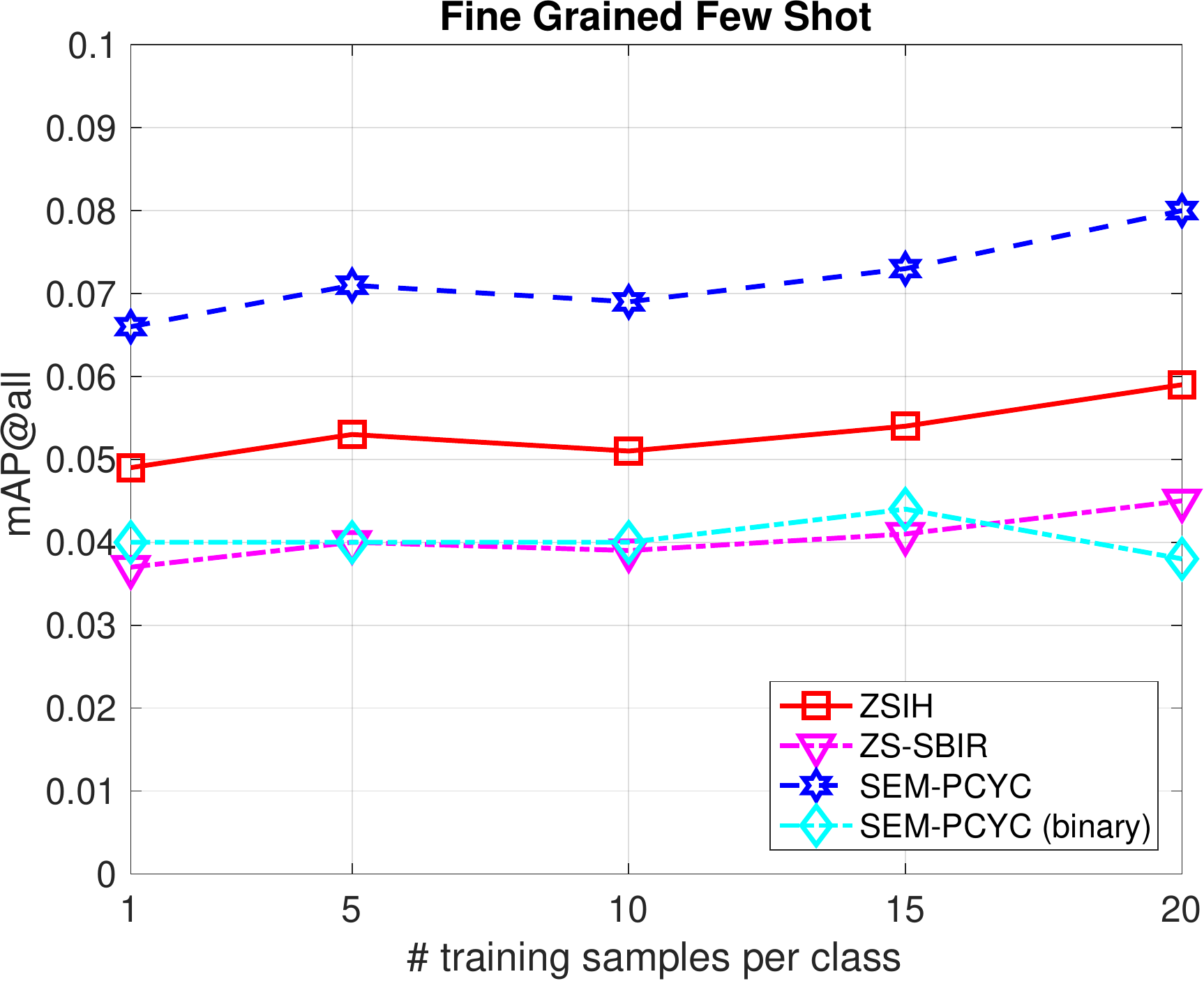} & \includegraphics[width=0.33\textwidth]{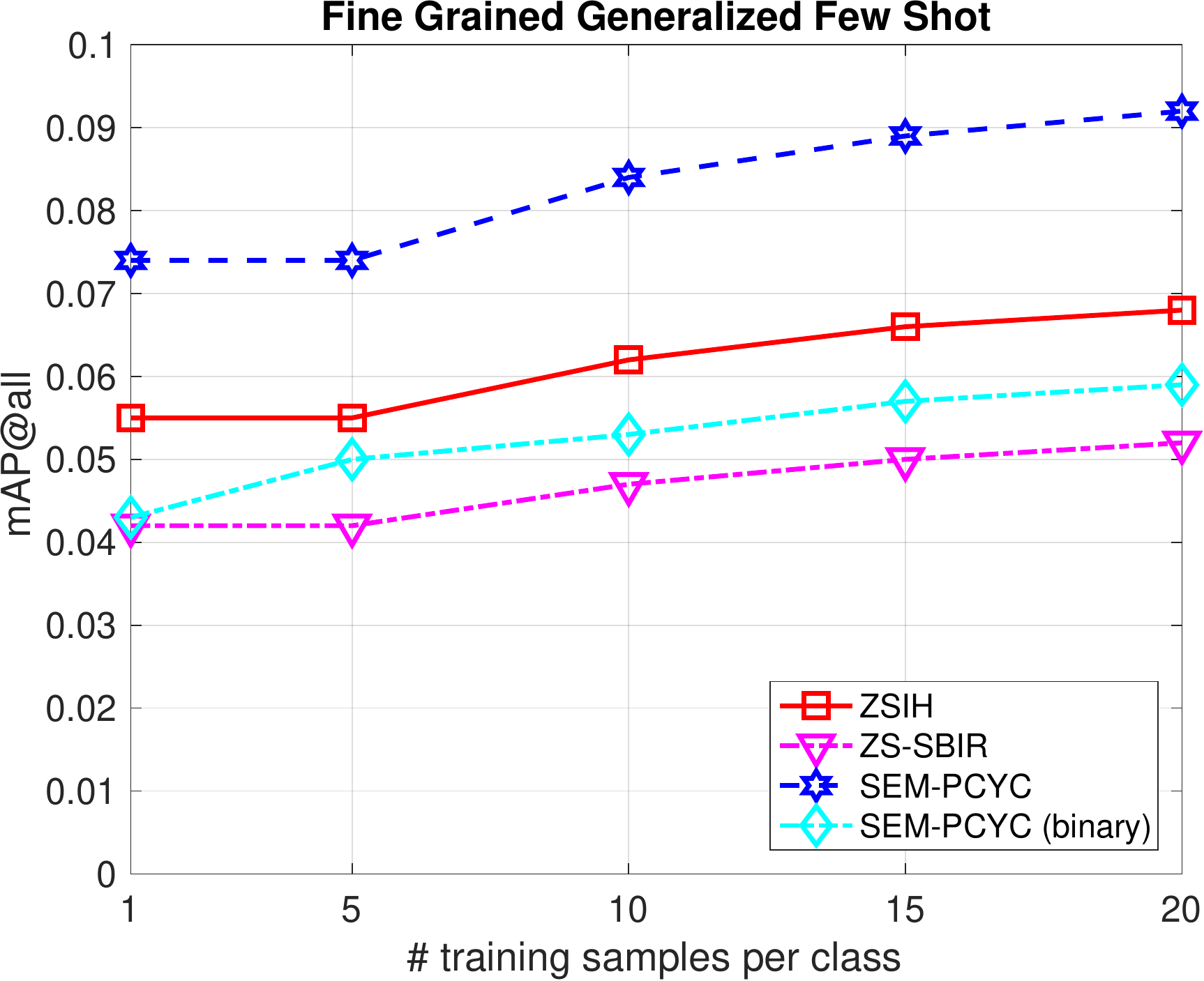} \\
(a) & (b)
\end{tabular}
\caption{Fine-grained (generalized) few-shot sketch-based image retrieval performance comparison.}
\label{fig:fine_grained_plots}
\end{figure*}

\myparagraph{Fine-Grained Settings.} We have further evaluated our model in fine-grained setting where the task is to find a specific object image of a drawn sketch, and we have combined it with the above mentioned variations of $k$-shot scenarios. For this experiment, we only considered the Sketchy dataset as only this corpus contains aligned sketch-image pairs, which are often used for fine-grained SBIR evaluation tasks. We have not considered other fine-grained datasets, such as \emph{shoe}, \emph{chair} etc~\cite{Song2017FineGrained} as they do not contain class information which we need for semantic space mapping. For this setting as well, we have considered ZSIH~\cite{Shen2018ZSIH} and ZS-SBIR~\cite{Yelamarthi2018ZSBIR} as the two benchmark methods and the same experimental protocol.

\fig{fig:fine_grained_plots}(a) and \fig{fig:fine_grained_plots}(b) show the performance of our model in fine-grained generalized few-shot together with ZSIH \cite{Shen2018ZSIH} and ZS-SBIR~\cite{Yelamarthi2018ZSBIR}. In fine-grained setting, all the methods have performed remarkably poor. We explain this fact as the drawback of semantic space mapping which intends to map visual information from sketch and image to the same neighborhood and ignores fine-grained information. Therefore the proposed solution to low-shot task and the notion of fine-grained problem contradicts, and as a consequence the performance of all the considered models deteriorates. In generalized setting, we have observed that all the models have performed slightly better. We conjecture that the considered models can memorize the fine-grained information of the training or \emph{seen} samples, which gives a slight rise (as they are very few in number) in performance in generalized scenario. However, we see that low-shot fine-grained paradigm is very important for SBIR. Nevertheless, we admit that it is an extremely challenging task, which needs substantial research work to be solved.



\section{Conclusion}
\label{sec:concl}
In this paper, we proposed the SEM-PCYC model for the any-shot SBIR task. Our SEM-PCYC model is a semantically aligned paired cycle consistent generative adversarial network whose each branch either maps a sketch or an image to a common semantic space via adversarial training with a shared discriminator. Thanks to cycle consistency on both the branches our model does not require aligned sketch-image pairs. Moreover, it acts as a regularizer in the adversarial training. The classification losses on the generators guarantee the features to be discriminative. We show that combining heterogeneous side information through an auto-encoder, which encodes a compact side information useful for adversarial training, is effective. In addition to the model, in this paper, we introduced (generalized) few-shot SBIR as a new task, which is combined with fine-grained setting. We considered three benchmark datasets with varying difficulties and challenges, and performed exhaustive evaluation with the above mentioned paradigms. Our assessment on these three datasets has shown that our model consistently outperforms the existing methods in (generalized) zero- and few-shot, and fine-grained settings. We encourage future work to evaluate sketch based image retrieval methods in these incrementally challenging and realistic settings.

\section*{Acknowledgments}
This work has received funding from the European Union under Marie Sk\l{}odowska-Curie grant agreement No. 665919, from the ERC under the Horizon 2020 program (grant agreement No. 853489), the Spanish Ministry project RTI2018-102285-A-I00 and DFG-EXC-Nummer 2064/1-Projektnummer 390727645. The TITAN Xp and TITAN V used for this research were donated by the NVIDIA Corporation.

\bibliographystyle{spmpsci}      
\bibliography{bib/bibliography}   

%
%

\end{document}